\pdfoutput=1
\documentclass{article}
\usepackage{iclr2027_conference,times}

\usepackage{amsmath,amsfonts,bm}

\def\eqref#1{equation~\ref{#1}}

\def\1{\bm{1}}

\DeclareMathAlphabet{\mathsfit}{\encodingdefault}{\sfdefault}{m}{sl}
\SetMathAlphabet{\mathsfit}{bold}{\encodingdefault}{\sfdefault}{bx}{n}

\usepackage[T1]{fontenc}
\usepackage[utf8]{inputenc}
\usepackage[section]{placeins}
\usepackage{microtype}
\usepackage{booktabs}
\usepackage{multirow}
\usepackage{graphicx}
\usepackage{amsmath,amssymb}
\usepackage{xcolor}
\usepackage{wrapfig}
\usepackage[shortlabels]{enumitem}
\usepackage{makecell}
\usepackage{soul}
\usepackage{hyperref}
\usepackage{url}
\usepackage{xurl}
\usepackage{pgfplots}
\pgfplotsset{compat=1.17}
\usepgfplotslibrary{fillbetween}
\usepackage{caption}
\usepackage{colortbl}
\usepackage{fontawesome5}
\usetikzlibrary{positioning,arrows.meta,fit,calc,shapes.geometric,backgrounds}

\newcommand{\sol}{GPT-5.6 Sol}
\newcommand{\luna}{GPT-5.6 Luna}
\newcommand{\opus}{Claude Opus 5}
\newcommand{\haiku}{Claude Haiku 4.5}
\newcommand{\kimi}{Kimi K3}
\newcommand{\gpro}{Gemini 3.1 Pro}
\newcommand{\gflash}{Gemini 3.8 Flash}
\newcommand{\reader}{\textsc{TeaReader}}
\newcommand{\bench}{\textsc{Tea Leaves}}

\newcommand{\ours}{\rowcolor{teal!10}}
\definecolor{cCurve}{HTML}{9DB7D5}
\definecolor{cFull}{HTML}{2A5D9F}
\definecolor{cAcc}{HTML}{D9822B}

\title{Reading the Tea Leaves, Again:\\ Can Large Language Models Predict Language Model Training Runs?}

\iclrfinalcopy
\author{Xinran Zhao$^{1}$\thanks{Correspondence to \texttt{xinranz3@andrew.cmu.edu}. Our code and data are at:\newline \url{https://github.com/colinzhaoust/read-tea-leaves}.}\,,\; Xiang Li$^{2}$,\; Tong Chen$^{3}$,\; Jinchuan Tian$^{1}$,\; Zichun Yu$^{1}$,\; Tongshuang Wu$^{1}$\\
{\small $^{1}$Carnegie Mellon University \quad $^{2}$University of Illinois Urbana-Champaign \quad $^{3}$University of Washington}}

\begin{document}
\maketitle

\begin{abstract}
LLM agents now propose research ideas and run the experiments that test them, but they train every run to its end, whereas a researcher reads a run's configuration and the beginning of its curve and stops the runs that show no promise.
Reading a run this way takes more than describing or implementing a training method: it takes knowing what a design choice does to a run.
We build \bench{} to measure this knowledge on 682 real training runs across pretraining, fine-tuning, and reinforcement learning (RL): a forecaster sees the first 20\% of a run with its configuration, description, and code, and predicts the rest.
Across 11 LLMs and a range of forecasting baselines, the strongest LLMs match or exceed the best curve fitter on smooth loss curves (skill 0.66 against 0.62) and lead on noisy RL reward curves (0.43 against 0.13).
What bounds every forecaster is what the log carries: on sparsely logged community runs all methods lose most of their skill, and none reliably flags a failing run in advance.
Beyond the curve, information helps and effort does not: related runs in the prompt cut the error on loss curves by 39\%, and knowledge of training dynamics lets the strongest LLMs order all 36 pairs of runs that differ by a described code change, which the early curves do not separate, whereas tools add little.
This knowledge can be learned from logged runs: a small open model trained on public runs matches the best LLM on loss curves.
Applied to the progress curves of research agents, the same reading ranks tasks by their remaining headroom, although for a single agent run it does not improve on the last value.
We release the benchmark, the trained readers, and their training data to test and train LLMs' understanding of training and to help research agents decide which runs deserve their compute.

\end{abstract}

\section{Introduction}
\label{sec:intro}

Large language model (LLM) agents now propose research ideas and pursue them on their own~\citep{lu2024aiscientist,yamada2025aiscientistv2,jansen2025codescientist,schmidgall2025agentlab}.
Because every idea is expensive to execute, prior work evaluates and trains LLMs to predict the outcome of research from its description: whether an idea will work, which of two proposals will score higher, or what a study will find~\citep{wen2025predicting,luo2024brainbench,anticipatory2025}.
Yet even a promising idea has to be executed, and in machine learning (ML) that means training models, run after run, until the idea is judged~\citep{si2025ideas,si2026ideationexecution,si2026executiongrounded}.
Today's agents, and the benchmarks that score them~\citep{huang2024mlagentbench,chan2025mlebench,nathani2025mlgym}, train every run to its end, whereas a researcher babysits a run~\citep{karpathy2016cs231n}: they read its configuration and code, watch the first stretch of its curve, and stop it early if it shows no promise, saving the compute for the next one.
Without this judgment, an agent pays for every run in full.

This judgment rests on an understanding of training that current evaluations do not isolate.
An LLM can explain what a learning-rate warmup or the Adam optimizer is, and a coding agent can implement either~\citep{huang2024mlagentbench,chan2025mlebench}; neither shows that the model knows what the component does to a run, that is, how the curve moves when the setting changes.
A logged run records this directly: a set of design choices and their measured consequence.
Forecasting a run from its settings therefore tests that understanding, and because public logs hold many such pairs, the same task supplies a training signal for it.

\begin{figure}[t]
\centering
\includegraphics[width=\linewidth]{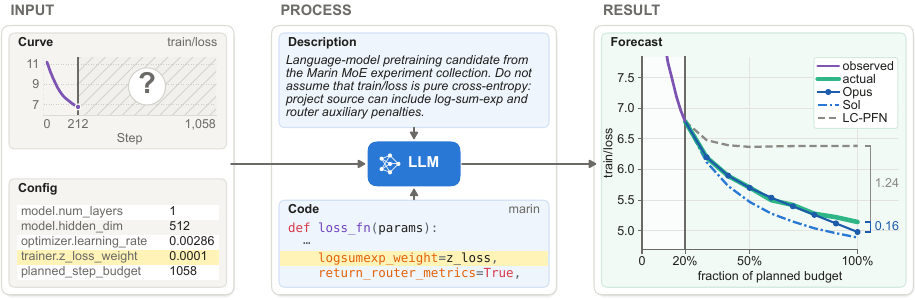}
\caption{The task on one test run (Marin MoE pretraining). Input: the first 20\% of the loss curve, the rest hidden, and the configuration. Process: \opus{} reads them with the description and code, which here flag auxiliary loss terms (highlighted). Result: its forecast against the actual curve (green), with those of \sol{} and LC-PFN; brackets mark the final-loss error.}
\label{fig:overview}
\vspace{-6pt}
\end{figure}

To test whether LLMs can read the same signs,\footnote{The title borrows from \citet{chang2009tealeaves}, who asked whether humans can read a topic model from the faint signs it leaves. We ask, again, whether a model can read a training run from its faint signs, a partial curve, a configuration, and code, with GPU hours riding on the answer.} we build \bench{} from 682 real public training runs, each queried at eight future points of its curve (5{,}456 forecast targets), and three choices in its design decide what it can measure (\S\ref{sec:benchmark}).
Figure~\ref{fig:overview} illustrates the task. First, a forecaster sees the first 20\% of a run together with its configuration and the \emph{semantics} of the experiment, its description and training code, and predicts the rest with uncertainty. Answering well takes both extrapolation and a working knowledge of training dynamics, how the learning rate, model size, data, and algorithm shape a curve~\citep{kaplan2020scaling,tissue2024annealing,luo2025multipower}; two companion tasks isolate each.
Second, the runs span pretraining (PT), fine-tuning (FT), and reinforcement learning (RL), whose curves range from smooth losses to noisy rewards, so we can see where extrapolation stops sufficing.
Third, the runs keep their provenance. Curated runs come from controlled sweeps with recorded lineage, so related runs can be retrieved and leakage controlled; community runs from the Hugging Face Hub pose the question in the wild, and failed runs test whether a failure is visible in advance.
A skill score $S$ normalizes each forecast by the error of repeating the last observed value, so runs can be compared across metrics and scales. 

We evaluate 11 LLMs against 8 baselines, spanning curve fitters, time-series foundation models, and predictors trained on past runs~\citep{adriaensen2023lcpfn,ansari2025chronos2,das2024timesfm,zhang2026ncpl}, and find that \textbf{the strongest LLMs are competitive on loss curves and lead on rewards, yet their skill falls as the runs get harder to read} (\S\ref{sec:main}).
Smooth loss curves are the easy case: the strongest LLMs match or exceed the best curve fitter, which knows nothing about the experiment (pooled over pretraining and fine-tuning, \sol{} reaches a skill of $0.66$ and \opus{} $0.65$, against $0.62$ for LC-PFN).
Noisy reward curves are harder: the best LLM reaches $0.43$, whereas the best curve fitter in our main comparison reaches $0.13$ and most methods barely improve on repeating the last value.
Sparsely logged community runs are harder still, and every method loses most of its skill there (\opus{} falls to $0.31$).
Hardest of all are runs that eventually fail: no method, LLM or otherwise, reliably tells them apart from healthy runs in advance (\S\ref{sec:extreme}).

We then analyze what separates the easy runs from the hard ones, and what a forecaster needs in order to read the hard ones.
The easy runs are those whose \emph{shape already carries most of the answer}: on smooth loss curves, the strongest LLMs reach most of their skill from a tenth of the run, and neither finer input windows (\S\ref{sec:extreme}) nor the semantics of the experiment (\S\ref{sec:main}) help consistently.
On the hard runs, a forecaster needs information that the curve does not carry, and we find two kinds that help.
The first is a \textbf{working knowledge of training dynamics}.
On reward curves, a third of the strongest LLM's advantage over curve fitters comes from reading the semantics (\S\ref{sec:main}).
Not every model has this knowledge: asked which of two anonymized settings produced which curve, three of nine models answer at chance, while the five strongest are right over 80\% of the time.
The knowledge can be learned from logged runs: a 9B model fine-tuned on public runs with their semantics matches the best LLM on loss curves and still trails on rewards (\S\ref{sec:training}).
The second is \textbf{experience from related runs}.
Related runs in the prompt cut the error on loss curves by 39\%, because they show how runs like this one went on (\S\ref{sec:retrieval}); tools alone, a sandbox and literature search, add little (\S\ref{sec:agent}).
For research agents, this suggests that runs are worth logging densely and storing with their configuration, code, and lineage, and that the expensive call is better spent on runs whose early curve does not already give the answer.

Finally, we take two steps from the forecast toward its use.
In a sweep we trained, where a code change acts only after the observed 20\%, \sol{} and \opus{} order all 36 pairs of runs that differ in the code change correctly from a description of it, against 46\% for keeping the current leader; on the benchmark's pretraining sweeps, continuing one run per sweep by \sol{}'s forecast ends within 1\% of the sweep's range from the best final loss, with 33.5\% and 42.6\% less training budget than keeping the current leaders needs (\S\ref{sec:decisions}).
The reading also carries over to the progress curves of autonomous research agents: the runs of a task, read together, order the tasks by the gain still to come (Spearman $0.72$; \S\ref{sec:rsi}), whereas a single agent run is no more forecastable than its last value.

In summary, we contribute (1) \bench{}, a benchmark of 682 real training runs at three stages, with its scorer, every model answer, the database of runs, and the trained readers; (2) evidence that whether a run can be read depends on what its log carries, and that beyond the log only knowledge of training dynamics and related runs help; (3) a measurement of that knowledge, in forecasting across 11 LLMs and in matching across nine, and a recipe that trains a small model to match the best LLM on loss curves; and (4) two pilot studies that connect the forecast to the choice of runs to continue, with an extension to the progress curves of research agents.
We hope \bench{} lets research agents be evaluated on, and trained for, the judgment a researcher applies before spending compute, and that logged runs come to serve as supervision for what a design choice does to training, beyond what it is and how it is implemented.

\section{Related Work}
\label{sec:related}

\textbf{Predicting training from numbers.}
Scaling laws predict the final loss from model and data size~\citep{kaplan2020scaling,hoffmann2022training}, or downstream task performance from a ladder of small models~\citep{bhagia2025ladders}, and require a sweep of completed runs; learning-curve extrapolation forecasts from a run's own prefix with parametric families~\citep{domhan2015speeding}, Bayesian models~\citep{swersky2014freezethaw}, or prior-fitted networks~\citep{adriaensen2023lcpfn,rakotoarison2024ftpfn}, and multi-fidelity optimization uses such forecasts to stop runs early~\citep{wistuba2022dyhpo}.
Recent predictors condition on more of the run: schedule-aware laws~\citep{tissue2024annealing,luo2025multipower}, NCPL, which maps a numeric configuration to the residual of a scaling-law fit~\citep{zhang2026ncpl}, NeuNeu, which forecasts downstream accuracy from training signals~\citep{neuneu2026}, and compute--performance curves for RL~\citep{khatri2025scalerl}.
Time-series foundation models~\citep{ansari2024chronos,ansari2025chronos2,das2024timesfm} and LLMs prompted with numbers~\citep{gruver2023llmtime,requeima2024llmp} forecast generic series, and whether text improves an LLM's numeric forecast is contested~\citep{tan2024llmts,williams2024cik,beyondnaive2025}.
The generic forecasters read only the series, and the run-specific predictors read numbers rather than text; we test what the configuration, description, and code add to the curve.

\textbf{Predicting the outcome of research.}
LLMs have been asked to anticipate results from descriptions: which of two ideas will score higher~\citep{wen2025predicting}, how a neuroscience or social-science study turns out~\citep{luo2024brainbench,ashokkumar2026socialscience}, what score a method will reach~\citep{anticipatory2025,scipredict2026}, and what a community will study next~\citep{li2026ideaforecast,wang2026futurealigned}; the same models forecast real-world events at near-human level~\citep{halawi2024forecasting}.
Theories distilled from the literature predict later results better than a model's parameters alone~\citep{jansen2026theories}, hypothesis generators learn from past outcomes which ideas will work~\citep{vasu2025harpa}, and LLM surrogates guide Bayesian optimization~\citep{liu2024llambo}.
Much of this work predicts research endpoints; our target is the logged training trajectory, published or not.

\textbf{Agents that run experiments.}
Benchmarks for research agents score what the agent ships: a model improved on a fixed task~\citep{huang2024mlagentbench,chan2025mlebench}, a result in a research-level challenge~\citep{nathani2025mlgym}, a replication graded against a paper~\citep{starace2025paperbench}, or engineering progress measured against human experts under a time budget~\citep{wijk2025rebench}; full systems propose, run, and write up experiments end to end~\citep{lu2024aiscientist,jansen2024discoveryworld,jansen2025codescientist}.
Deciding mid-run which experiments to keep is part of every such task but is never measured on its own; \bench{} isolates that step.
Our database of runs (\S\ref{sec:retrieval}) follows retrieval-augmented forecasting~\citep{tire2026raf}, which retrieves by series shape, whereas we retrieve by configuration; large public curve collections show how varied the shapes are~\citep{yan2025lcdb}.

\section{The \bench{} Benchmark}
\label{sec:benchmark}

\bench{} measures the forecast behind a researcher's decision to stop or continue a run: the forecaster sees the settings and the prefix and predicts the rest; scoring the stopping decision itself is left to future work, and \S\ref{sec:decisions} reports two pilot studies in that direction.
\S\ref{sec:tasks} defines the task and its metric, \S\ref{sec:data} the data, and \S\ref{sec:baselines} the forecasters.

\subsection{Task formulation}
\label{sec:tasks}

A run is a triple $r=(c,s,y)$: a numeric configuration $c$ (learning rate, batch size, model size, budget, and the other fields the project logs), the semantics $s$ (a description of the task and data, facts about the source project, and excerpts of the training code), and a metric curve $y:[0,1]\to\mathbb{R}$ indexed by the fraction $x$ of the run's \emph{planned} budget in its own axis (tokens, steps, or epochs).
Together, $c$ and $s$ describe the experiment: its settings, task, and implementation.
For the pretraining and fine-tuning windowed protocol, the observed curve is summarized as means over the first ten windows of width $\delta=0.02$,
\begin{equation}
\bar y_k=\operatorname{mean}\{\,y(x): x\in((k-1)\delta,\,k\delta]\,\},\qquad k=1,\dots,10,
\label{eq:prefix}
\end{equation}
so that the prefix $y_{\le0.2}$ is the ten values $\bar y_{1:10}$.
Given $(\bar y_{1:10}, c, s)$, a forecaster returns the 10\%, 50\%, and 90\% quantiles of $y$ at the query grid $\mathcal{X}=\{0.3,0.4,\dots,1.0\}$; the median $\hat y$ is the point forecast.
This is the \textsc{Full} condition, in which a researcher or an agent would use a forecaster. %
Answering well requires two abilities: extrapolating the observed curve, and a working knowledge of training dynamics, how the learning rate, model size, data, and algorithm shape a curve~\citep{kaplan2020scaling,tissue2024annealing,luo2025multipower}.
For example, a flat prefix during learning-rate warmup need not imply that training has converged. Two further tasks probe these abilities.
Forecasting from the curve alone, as curve fitters do~\citep{domhan2015speeding,adriaensen2023lcpfn}, tests extrapolation by itself; Appendix~\ref{app:conditions} removes $s$ and $c$ in turn.
A matching task tests the knowledge by itself: given two runs from the same sweep, the model sees both curves and their anonymized settings and says which setting produced which curve (Appendix~\ref{app:matching-task}).

\textbf{Metric.}
Curves differ in metric, unit, and scale (a loss in nats, a reward in $[0,1]$), so errors cannot be averaged across runs; we therefore score each run against its own persistence forecast $\hat y_0(x)\equiv\bar y_{10}$, which repeats the last observed value.
A forecast is scored by its mean absolute error over the grid in the run's own units, and by the ratio of that error to the persistence error:
\begin{equation}
\mathrm{MAE}(r)=\frac{1}{|\mathcal{X}|}\sum_{x\in\mathcal{X}}\big|\hat y(x)-y(x)\big|,\qquad
\rho(r)=\frac{\mathrm{MAE}(r)}{\mathrm{MAE}_0(r)}.
\label{eq:mae}
\end{equation}
The ratio $\rho$ is unit-free, so we aggregate it geometrically over the runs of a cell $g$ (a stage $\times$ source group) and then over cells,
\begin{equation}
R=\exp\Big(\frac{1}{|G|}\sum_{g\in G}\frac{1}{|g|}\sum_{r\in g}\ln\rho(r)\Big),\qquad S=1-R,
\label{eq:skill}
\end{equation}
and report the \emph{skill score} $S$, as in weather forecasting~\citep{murphy1988skill}.
$S$ is 0 for repeating the last value, 1 for a perfect forecast, and negative for a forecast worse than persistence; 95\% intervals resample lineage components, so that runs sharing a trajectory are never resampled independently (worked example and edge cases in Appendix~\ref{app:metrics}). The main comparison scores medians; Appendix~\ref{app:metrics} compares quantile losses, and Appendix~\ref{app:reader} examines the trained reader's interval coverage.

\subsection{Data}
\label{sec:data}

\textbf{Sources.}
We enumerate runs from three kinds of source: public experiment-tracker projects of open training efforts (19{,}650 run identities recorded across 236 collection-log files), training-state files on the Hugging Face Hub\footnote{\url{https://huggingface.co/docs/hub/index}} (47{,}929 listed models, 14{,}370 ungated, non-excluded repositories with a trainer state, 4{,}503 parsed), and a public speedrun log archive (1{,}516 logs)\footnote{\url{https://github.com/KellerJordan/modded-nanogpt}}.
The training pools are built from 17{,}805 collected records before identity and quality filtering, a count of records and not of independent experiments (sources and funnel in Appendix~\ref{app:data}).

\textbf{Filtering.}
A run enters the benchmark only if it passes every one of the following rules.
(1) \textbf{Defined target.} The run logs a training metric with a known axis and a known planned budget, so that $y$ and the fraction $x$ are defined.
(2) \textbf{Visible semantics.} The project exposes a configuration and either code or a description, so that $s$ and $c$ can be shown; we keep the configuration as logged and provide source-linked code excerpts.
(3) \textbf{Curve quality.} The log has enough points before and after the 20\% mark to define every window mean and every query point, no duplicated history, and at most two seed replicates of the same setting (counts in Appendix~\ref{app:data}).
(4) \textbf{Controlled variation (test sets only).} The run belongs to a sweep that varies a setting a researcher changes (learning rate, batch size, warmup, budget, model size, data, or initial checkpoint), and every stage draws on several sources.

\textbf{Splitting.}
Test sets are drawn first: 150 pretraining and 150 fine-tuning runs, drawn together with the matching pairs from 589 run identities, and 150 RL runs from more than twenty repositories.
Runs that resume from or branch off one another share part of a trajectory, so whole lineage components are assigned to one split. Every run also carries its sweep family and source project; training pools are built to remove every test sweep family (1{,}463 pool IDs remain; Appendix~\ref{app:reader} reports the links to test runs that remain in the reader's pools) or, where stated, the whole source (\S\ref{sec:training}).
The pretraining, fine-tuning, and RL test sets were frozen with content hashes before any model was evaluated; for the community set, the selection rule and the requests were fixed before any model was run on it.
Each run is queried at the eight points of the grid $\mathcal{X}$, so the 682 test runs pose 5{,}456 forecast targets, each answered with three quantiles.
Table~\ref{tab:data} summarizes the four test sets, and Figure~\ref{fig:stage-examples} in Appendix~\ref{app:data} shows one run of each as logged.
Appendix~\ref{app:census} lists the model scales, task types, and budgets of each set, and the training time that the logs record.
The community set, our test in the wild, holds 232 runs by 184 authors, drawn from 1{,}513 Hub candidates; none shares a repository with another set, and we removed 36 candidates whose author or token collided with the frozen sets.

\newcommand{\vary}[1]{\ul{#1}}
\newcommand{\exrun}[1]{%
\arrayrulecolor{black!30}\midrule
\multicolumn{5}{@{}l@{}}{\textcolor{gray}{\emph{Ex.}: #1}}}
\begin{table}[t]
\centering\footnotesize
\setlength{\tabcolsep}{4pt}
\caption{The \bench{} test sets; \emph{Ex.} gives one question from each set as the forecaster sees it, with the settings that vary within its sweep. Sources: Marin~\citep{marin2025}, whose fine-tuning runs follow the T\"ulu~3 recipe~\citep{lambert2024tulu3}, DCLM~\citep{li2024dclm}, OLMoE~\citep{muennighoff2024olmoe}, and Open-R1~\citep{openr1}.}
\label{tab:data}
\resizebox{\linewidth}{!}{%
\begin{tabular}{@{}l|l|l|r|l@{}}
\toprule
\textbf{Stage} & \textbf{Metric (axis)} & \textbf{Sources} & \textbf{Runs} & \textbf{What varies within a sweep} (\vary{underline}) \\
\midrule
Pretraining & loss (steps, tokens) & Marin, DCLM, OLMoE & 150 & learning rate, model size, batch, budget, data \\
\exrun{a Marin run of width \vary{1280} at learning rate \vary{1.6}, 20\% of its \vary{30.8B tokens} observed; \emph{where does train/loss end?}} \\
\arrayrulecolor{black!100}\midrule
Fine-tuning & loss (steps, epochs) & Marin (T\"ulu~3 recipe), Open-R1 & 150 & learning rate, epochs, model size, checkpoint \\
\exrun{Open-R1 fine-tuning of a \vary{7B checkpoint} for \vary{three epochs}, the first epoch partly observed; \emph{where does the loss settle?}} \\
\arrayrulecolor{black!100}\midrule
RL & reward, loss (steps, epochs) & 20+ lab and paper repositories & 150 & reward, budget, checkpoint, algorithm \\
\exrun{\vary{GRPO} on \vary{math prompts} from a \vary{1.5B checkpoint}, 86 of 430 steps observed; \emph{will the mean reward keep rising?}} \\
\arrayrulecolor{black!100}\midrule
Community (in the wild) & mixed & Hugging Face Hub, 184 authors & 232 & uncontrolled \\
\exrun{a community LoRA run whose trainer state logs 62 points, twelve of them in the prefix; the repository is the only description.} \\
\arrayrulecolor{black!100}\bottomrule
\end{tabular}%
}
\end{table}

\textbf{Setup.}
Following NeuNeu~\citep{neuneu2026}, the forecaster observes the first 20\% of the planned budget and predicts the rest (\S\ref{sec:extreme} varies both); an LLM receives the ten prefix means (on RL, the released raw prefix, whose density varies by run; Appendix~\ref{app:baselines}), the configuration, the description, and the code excerpts in one prompt and answers once (decoding and exact model identifiers in Appendix~\ref{app:models}).
We call one run posed in this setup a \emph{question}, and the input of the \textsc{Full} condition the full context. Unless stated otherwise, every experiment in \S\ref{sec:experiments} uses this setup; one complete question and its answer are shown in Appendix~\ref{app:example-full}.

\subsection{Baselines and models}
\label{sec:baselines}

We evaluate five families of forecaster, each standing for a way a researcher or a lab could obtain a forecast today.
(1) \textbf{Curve fitting} extrapolates the prefix alone, as a researcher would first: the last value, a power-law fit, an AIC-weighted ensemble of parametric fits, and the prior-fitted networks LC-PFN~\citep{adriaensen2023lcpfn} and FT-PFN~\citep{rakotoarison2024ftpfn}.
(2) \textbf{Time-series foundation models} forecast any numeric series zero-shot: Chronos-2~\citep{ansari2025chronos2} and TimesFM-2.5~\citep{das2024timesfm,timesfm25}.
(3) \textbf{Predictors trained on past runs} represent a lab that learns from its own history: NeuNeu-NoLoss~\citep{neuneu2026}, trained on an external source's runs, and NCPL~\citep{zhang2026ncpl}, a predictor built for neural scaling laws, retrained with the authors' code on our pool with every test sweep held out (implementation of every baseline in Appendix~\ref{app:baselines}); two shape regressors trained on the pool are reported in Appendix~\ref{app:regimes-main}.
(4) \textbf{Closed-weight LLMs} read the curve and the semantics together: \sol{} and \luna{}~\citep{openai2026gpt56}, \opus{} and Claude Haiku 4.5~\citep{anthropic2025haiku45,anthropic2026opus5}, and \gpro{} and \gflash{}~\citep{gemini2026flash38,gemini2026pro31}.
(5) \textbf{Open-weight LLMs}, which a lab can run and train itself: Kimi K3~\citep{kimi2026k3}, Qwen3-Next-80B~\citep{qwen2025qwen3next}, DeepSeek-V3.2~\citep{deepseek2025v32}, Muse-Glimmer-30B~\citep{meta2026museglimmer}, and Qwen3.5 at 0.8B, 2B, 4B, and 9B~\citep{qwen2026qwen35}.
All LLMs are zero-shot with the same prompt unless otherwise specified.

\begin{figure}[t]
\centering
\includegraphics[width=\linewidth]{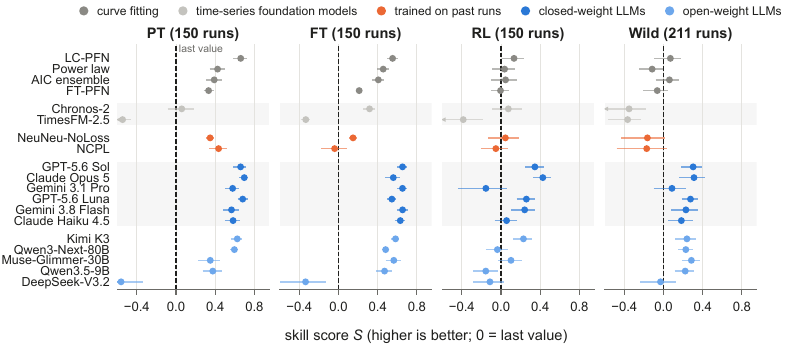}
\caption{Forecasting skill $S$ on PT, FT, RL (150 runs each), and the 211 community runs (Wild) that every model evaluated first on this set answered validly (Appendix~\ref{app:wild-details}); each model is scored on its own valid answers, and RL on the 133 runs whose logged series moves after the prefix (Appendix~\ref{app:metrics}). Bars: 95\% intervals; dashed: last value. Pool-trained models hold out every test sweep family. }
\label{fig:main}
\end{figure}

\section{Experiments and Analysis}
\label{sec:experiments}

The experiments follow three questions.
\emph{Can LLMs read a run?} \S\ref{sec:main} reports the benchmark, and \S\ref{sec:extreme} varies how much of a run is shown.
\emph{What does a reader need beyond the curve?} We test knowledge learned from logged runs (\S\ref{sec:training}), related runs retrieved at query time (\S\ref{sec:retrieval}), and tools (\S\ref{sec:agent}).
\emph{What is the reading good for?} We test whether forecasts improve the choice of runs to continue (\S\ref{sec:decisions}) and whether the reading carries over to research agents (\S\ref{sec:rsi}).
In the section titles, a \emph{reader} is any forecaster that is shown a run.
Figure~\ref{fig:main} compares the 11 LLMs with the forecasting baselines; three smaller Qwen3.5 models appear only in Appendix~\ref{app:main-table}. Unless stated otherwise, every forecaster sees the first 20\% of a run, LLMs see the full context, and scores are the skill score $S$ of \S\ref{sec:tasks} with 95\% intervals over lineage components; paired contrasts are collected in Appendix~\ref{app:significance}.

\subsection{Reading the tea leaves}
\label{sec:main}

\paragraph{The strongest LLMs are competitive on loss curves and lead on rewards.}
Figure~\ref{fig:main} plots forecasting skill on the four test sets, one column per stage; several patterns stand out.
On pretraining runs \opus{} reaches $S=0.70$, \luna{} $0.68$, and \sol{} $0.66$, against $0.66$ for LC-PFN, the best curve fitter, and at most $0.46$ for every other zero-shot baseline (Appendix~\ref{app:main-table}).
On fine-tuning runs the best LLMs reach $0.63$--$0.65$ against $0.55$ for LC-PFN.
Predictors trained on past runs with every test sweep family held out do not close this gap (on pretraining and fine-tuning, NCPL $0.43$ and $-0.04$, NeuNeu-NoLoss $0.35$ and $0.15$); a predictor that has seen the test run's own sweep interpolates within it (retrained NCPL falls from $0.79$ to $0.28$ when that sweep is held out; Appendix~\ref{app:regimes-main}), and the LLMs' gain over LC-PFN concentrates on the runs where the fitter errs (Appendix~\ref{app:routing}).
Within the LLMs, the six closed-weight models fall within $0.13$ of the best on loss curves, and the open-weight models vary widely (Muse-Glimmer-30B $0.35$ and $0.56$).

\paragraph{Reward curves are harder, and skill on losses does not transfer.}
Across the columns, the spread between stages exceeds the spread between most models within a stage.
Reward curves are noisy enough that the last value is a strong forecast, so on RL runs the best curve fitter in Figure~\ref{fig:main} reaches $0.13$, \opus{} $0.43$, \sol{} $0.35$, and no other method in the figure exceeds $0.26$.
Skill on RL depends on the model: the closed models spread from $-0.15$ to $0.43$, and \gpro{}, level with \sol{} on fine-tuning ($0.65$), scores $-0.15$ (Appendix~\ref{app:rl}; per-project errors in the supplementary document, \S S3).

\paragraph{In the wild, every method loses most of its skill with sparse, noisy logs.}
In the last column, on the 211 community runs, \opus{} falls to $0.31$, \sol{} to $0.30$, LC-PFN to $0.07$, and NCPL, retrained on our curated runs, to $-0.17$; the drop concentrates in RL runs and in runs by individual authors. No baseline without an LLM in Figure~\ref{fig:main} exceeds $0.07$ here, so what skill remains in the wild comes from the LLMs.
Descriptions do not account for the drop: removing every identifying field, or adding everything the model cards recover, leaves the error ratio within $0.96$--$1.02$ of the original.
The logs explain much of it: thinning curated curves to the community density (7 raw points per window against 59) with comparable noise reproduces, in the error floor of a hindsight-best fit, 95\% of the gap (Appendix~\ref{app:wild-details}).

\paragraph{Semantics help where the prefix misleads.}
Adding semantics raises \opus's RL skill from $0.33$ to $0.43$, a third of its lead over the best curve fitter (Table~\ref{tab:main}), and cuts its error by about a third on warmup-dominated prefixes; on loss curves the effect on the three strongest models is mixed, from $-0.19$ for \opus{} on fine-tuning to $+0.11$ for \luna{} on pretraining (Appendices~\ref{app:significance}, \ref{app:conditions}, and~\ref{app:context-hurts}).
The matching task shows the knowledge and its limit: the strongest models match settings to completed curves with 82--86\% accuracy, yet assignments derived from their forecasts are near chance (Appendix~\ref{app:matching-task}).
With the curve and the numeric configuration fixed, neither the description nor the code gives a consistent gain on 200 benchmark runs, where the text is rarely specific to a run (paired error ratios 0.94--1.07; Appendix~\ref{app:factorial}); a description that states a run-specific change does move the forecast in the right direction (\S\ref{sec:decisions}).

\subsection{How much of a run must a reader see?}
\label{sec:extreme}

\begin{figure}[t]
\centering
\includegraphics[width=\linewidth]{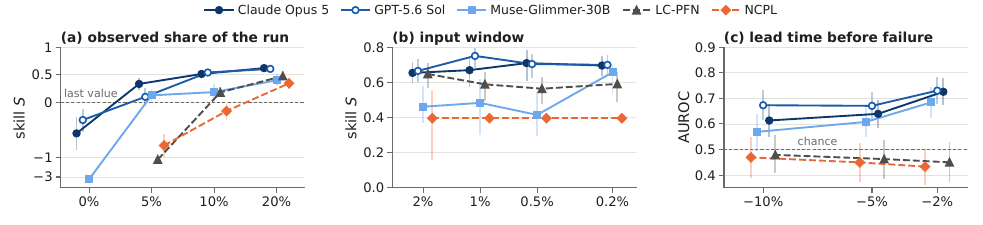}
\caption{(a) Skill $S$ against the observed share of the run, relative to the 20\% last value throughout (at 0\% only the configuration is given; $y$ axis compressed below $-1$); (b) skill against the input averaging window; (c) failure-detection AUROC against lead time. Orange: retrained NCPL, whose ten-window score is repeated in (b); bars: 95\% intervals.}
\label{fig:extremes}
\end{figure}

The main experiments observe 20\% of each run, average the logs in 2\% windows, and score a run only if it finishes.
We vary the first two on fixed subsets (100 and 86 questions); for the third, we ask whether 100 failed runs can be told from 200 matched healthy runs before they fail; protocols and full tables are in Appendix~\ref{app:extremes-details}.

\textbf{For the strongest LLMs, a tenth of a run carries most of the signal.}
Figure~\ref{fig:extremes}a shows skill as the observed share grows. \opus{} and \sol{} reach $0.51$--$0.54$ from 10\% of the budget and $0.61$--$0.62$ at 20\%. LC-PFN and the retrained NCPL are negative at 5\% ($-1.13$, $-0.79$), still at $0.18$ and $-0.16$ at 10\%, and need the full 20\% to reach $0.48$ and $0.35$.

\textbf{Finer windows help Muse-Glimmer-30B, while the strongest readers remain stable.}
Narrowing the input window from 2\% to 0.2\% leaves \opus{} and \sol{} at $0.65$--$0.75$ but raises Muse-Glimmer-30B from $0.46$ to $0.66$ (Figure~\ref{fig:extremes}b).
Finer windows expose more of an existing trace; sparse community logs lack the measurements themselves (\S\ref{sec:main}).

\textbf{No monitor reliably catches a failing run in advance.}
Figure~\ref{fig:extremes}c turns to failure.
Two percent of the budget before a failure, the LLMs separate failing from healthy runs with AUROC $0.73$ (\opus{} and \sol{}: three times in four they rank a failing run above a healthy one; $0.68$ for Muse-Glimmer-30B), which fades to $0.61$--$0.67$ ten percent ahead; LC-PFN and NCPL used as monitors stay at chance ($0.43$--$0.48$).
Run online at checkpoints with one false alarm allowed per five healthy runs, no monitor catches half of the failures (the best $0.41$, interval up to $0.51$, at the cost of the healthy runs it stops), apart from a KL-divergence threshold that reads the criterion labeling 38 of the 100 failures, and LLMs with the full context do no better than boosted trees on curve statistics ($0.36$ against $0.41$; both protocols in Appendix~\ref{app:extremes-details}).
The scalar log warns late.

\subsection{Can we train a reader?}
\label{sec:training}

Open-weight models vary widely on loss curves (\S\ref{sec:main}), and a lab may want an in-house reader for its own runs. We ask how much of the gap training on public runs closes: we fine-tune Qwen3.5-2B and 9B as text-to-text regressors~\citep{song2024omnipred,akhauri2025rlm} that read the prefix and the configuration, either alone or with the semantics of \S\ref{sec:tasks} condensed into one paragraph, and predict each query point as a ratio to the last observed value; we call the result \reader{}.
Training pools grow from the 918-run reference pool to 2{,}641, 3{,}002, and 8{,}727 runs by adding public pretraining sweeps, Hugging Face RL runs, and further public pretraining sweeps, admitted under the rules of \S\ref{sec:data}.
Two links to the test runs remain: 22 records in the two middle pools are linked to test runs, and under a broader graph of family and alias links, 462 of the 918 reference identities connect to an evaluation run (Appendix~\ref{app:reader}).
Each reader trains for 1{,}000 updates with three seeds, with a recipe fixed on the reference pool.

\begin{figure}[t]
\centering
\includegraphics[width=\linewidth]{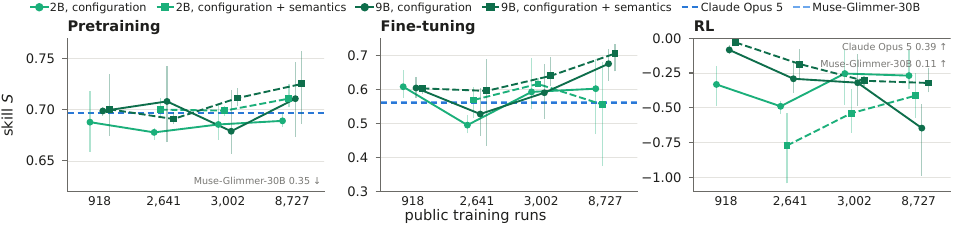}
\caption{Reader skill $S$ by stage, backbone, and training-pool size, with the configuration alone or with the semantics; dashed: \opus{} and Muse-Glimmer-30B on the same planned panels, each method using its own valid answers; faint bars: full min--max range over three training seeds, not confidence intervals.}
\label{fig:reader}
\end{figure}

\textbf{Training closes the gap on loss curves and not on rewards.}
Figure~\ref{fig:reader} plots skill against training runs, one panel per stage.
Before training, the 2B backbone scores $0.58$ (pretraining) and $0.18$ (fine-tuning) with the curve and configuration, and the 9B backbone $0.38$ and $0.47$ with the full context.
Trained, the 2B reader reaches $0.69$ and $0.61$ on pretraining and fine-tuning, level with \opus{} on the same runs, and the 9B reader with the semantics on the largest pool reaches $0.73$ and $0.71$, level with the best LLM.
On 129 later runs, created by new owners after the reader was trained, the 2B reader, with adapters and prompts unchanged, reaches $S=0.72$, against $0.74$ for \sol{} and \opus{}, $0.76$ for an ExtraTrees regressor trained on the benchmark's training split, and $0.64$ for LC-PFN (Appendix~\ref{app:temporal}); retraining the 2B readers after removing the 22 linked records changes their skill on the pretraining and fine-tuning test sets by at most $0.022$ (Appendix~\ref{app:clean}).
The semantics raise the 9B reader's average over the three stages on every pool (by $0.02$ to $0.12$), mostly on fine-tuning and RL, and hurt the 2B reader on RL.
On the RL test set none of these readers beats the last value, and more public runs make it worse: the 9B reader falls from $-0.08$ on the reference pool to $-0.64$ on the largest, and the semantics only slow the fall ($-0.03$ to $-0.32$), so the best three-stage average ($0.42$) stays far below \opus{} ($0.55$).
A reader trained later on 8{,}691 runs is the strongest we obtained: on a held-out development panel of 350 runs, not the test sets, it reaches $S=0.53$, against $0.31$ for the 2B recipe retrained for that panel and $0.37$ for LC-PFN, with positive skill on the panel's RL runs ($0.22$; Appendix~\ref{app:reader-v2}).

\subsection{Can past runs give a reader a memory?}
\label{sec:retrieval}

Every reader so far sees only the run in front of it, yet a lab keeps its past runs and the community keeps many more in public.
We therefore let the reader retrieve related runs from a database of past runs stored with their configuration, code, and lineage.
A shared community collection could extend this memory beyond one organization, grown by researcher submissions of curves and their context.

\begin{figure}[t]
\centering
\includegraphics[width=\linewidth]{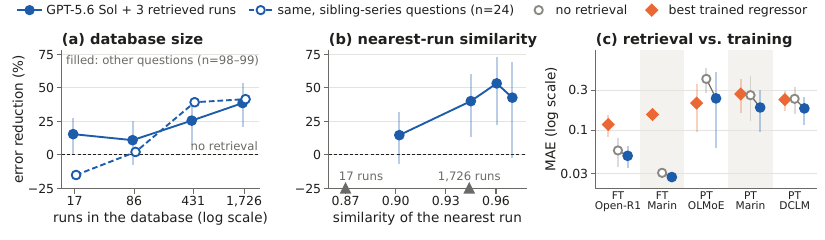}
\caption{Retrieval for \sol{} with three retrieved runs. (a) Error reduction against database size; hollow: questions for which a run from the same experiment series, never the same sweep family, was retrieved. (b) Error reduction by quartile of the similarity of the nearest retrieved run; triangles mark the mean similarity in the smallest and the largest database. (c) MAE per source with and without retrieval, against the best regressor trained on the same runs with the question's sweep family held out. Bars: 95\% intervals.}
\label{fig:retrieval}
\end{figure}

On 125 questions (123 with valid answers) we embed the configuration, fetch the three nearest runs outside the question's sweep family, and add their complete curves to \sol{}'s prompt; databases of 17 to 1{,}726 runs are subsamples of the same public pool (Appendix~\ref{app:det-retrieval}).

\textbf{A bigger database helps because it holds a closer run.}
Over all questions the smallest database barely helps (10\%, interval including zero) and the largest cuts \sol{}'s error by 39\% (Figure~\ref{fig:retrieval}a), as the nearest run becomes more similar (Figure~\ref{fig:retrieval}b); questions for which a run from a sibling series was retrieved (a project can group several sweep families into one experiment series) gain about as much as the rest (42\% against 39\%), so runs from the same series do not explain the gain. Retrieval also matches the benefit of training: with the sweep family excluded, \sol{} with retrieval beats the best regressor trained on the same runs in four of five sources (Figure~\ref{fig:retrieval}c); with the question's own sweep family allowed, it beats the regressor trained on it in all four sources that have one (Marin pretraining MAE $0.108$ against $0.209$); and it picks the best change in a real sweep 61\% of the time with retrieved sweeps against 46\% without (Appendix~\ref{app:det-retrieval}).

Given the same three neighbors and an explicit configuration encoding, FT-PFN reaches $S=0.82$ against $0.80$ for \sol{} with retrieval (\sol{} is better on pretraining, FT-PFN on fine-tuning), and a numerical transfer of the neighbors' curves reaches $0.60$ (Appendix~\ref{app:matched}).

\subsection{Does effort substitute for information?}
\label{sec:agent}

\begin{figure}[t]
\centering
\includegraphics[width=\linewidth]{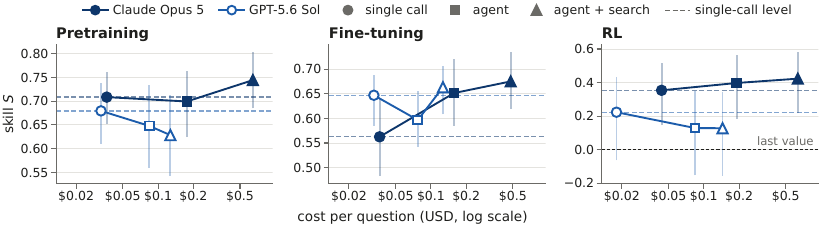}
\caption{An agent costs approximately 2--18 times a single call and changes skill by at most $0.11$. Skill $S$ against cost per question for \opus{} (filled) and \sol{} (hollow): single call (circle), agent (square), agent with search (triangle); dashed: each model's single-call skill; bars: 95\% intervals.}
\label{fig:agent}
\end{figure}

A run comes with a repository, a log, and related literature; does a forecaster that can open and use them do better than one given the prompt alone?
We let \opus{} and \sol{} work as agents on 100 pretraining, 100 fine-tuning, and 50 RL questions: the same material as a folder, with file access and Python in a sandbox for up to twenty turns; a second variant adds literature and code search with the run's own project filtered out (6 of the 1{,}000 planned attempts were not~run;~Appendix~\ref{app:det-agent}).

\textbf{Agents work like researchers and forecast like single calls.}
Figure~\ref{fig:agent} plots skill against cost.
The agents opened the code, fitted the prefix, and searched (logs in Appendix~\ref{app:det-agent}), yet \opus{} gained $0.09$ $[0.04,0.14]$ from the sandbox and $0.11$ $[0.05,0.17]$ with search added on fine-tuning only and stayed within $0.07$ of the single call elsewhere; \sol{} gained nothing: it lost $0.06$ on fine-tuning with the sandbox, and its RL skill falls from $0.22$ to $0.13$ with overlapping intervals.
An agent costs 2--5 times a single call, and 4--18 times with search (approximate cost reference; Appendix~\ref{app:det-agent}).
In a pilot on 50 questions (Appendix~\ref{app:det-agent}), an agent given the curves of related runs was level with a single call given the same curves, which alone cut the error by 24\% (\opus{}) and 44\% (\sol{}).
What a reader lacks is not effort but the runs that went~before (\S\ref{sec:retrieval},~\S\ref{sec:discussion}).

\subsection{From forecasts to decisions}
\label{sec:decisions}

\begin{figure}[t]
\centering
\includegraphics[width=\linewidth]{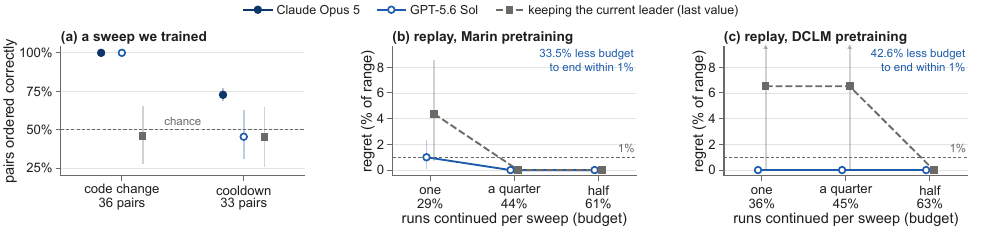}
\caption{From forecasts to decisions. (a) Share of pairs from the sweep we trained whose lower-ending run is named from the first 20\% (36 code-change and 33 cooldown pairs; intervals over the six seeds). (b, c) Replay of the 6 Marin and 4 DCLM pretraining sweeps: mean regret, as a share of the sweep's range, when one run, a quarter, or half of the runs of a sweep are continued, with the training budget each choice spends; intervals over sweeps, cut at the frame in (c), where they reach 19.6\%. Panels (b) and (c) show \sol{}, which ends within 1\% with one run per sweep.}
\label{fig:decisions}
\end{figure}

The capability that \bench{} measures is what a researcher uses to decide which runs deserve their compute. We ran two pilot studies of that step (Appendix~\ref{app:decisions}).

\textbf{A sweep we trained.} We trained 72 runs of a 124M-parameter model on FineWeb: three learning rates, two cooldown lengths, with or without a code change that freezes the embeddings from step 150, in six seeds. The cooldown and the code change act only after the 20\% cutoff, so the observed curves do not show them. Asked which of two runs will end lower, \sol{} and \opus{}, given the configuration and a description of each run, are right on all 36 code-change pairs (46\% for keeping the current leader; Figure~\ref{fig:decisions}a), \opus{} is right on 73\% of the 33 cooldown pairs (45\%), and both beat the current leader in all six seeds ($p=0.031$); \sol{} predicts the cost of the code change as $+0.098$ nats against a measured $+0.116$.

\textbf{A replay of the benchmark's sweeps.} In each sweep of at least four comparable runs, every run is observed for 20\% of its budget and the $m$ runs forecast to end best are continued, at budget $0.2+0.8\,m/n$ of training all $n$ runs. On the Marin and DCLM pretraining sweeps (6 and 4 sweeps, 85 runs), \sol{} continues one run per sweep and ends, on average, within 1\% of the sweep's range from the best final loss, where keeping the current leaders needs a quarter (Marin) or half (DCLM) of the runs (Figure~\ref{fig:decisions}b,\,c): 33.5\% and 42.6\% less training budget, and about two thirds less than training every run. On the fine-tuning and RL sweeps, the forecasts do not improve on keeping the current leaders. Across the LLM forecasters, skill and saving are positively related but not identical (pooled Spearman 0.23 $[0.08, 0.49]$): a simulation with forecasters of controlled skill (Appendix~\ref{app:skill-saving}) shows that an error shared by a whole sweep lowers the skill without changing the order of its runs.

\subsection{Does the reading carry over to research agents?}
\label{sec:rsi}

\begin{figure}[t]
\centering
\includegraphics[width=\linewidth]{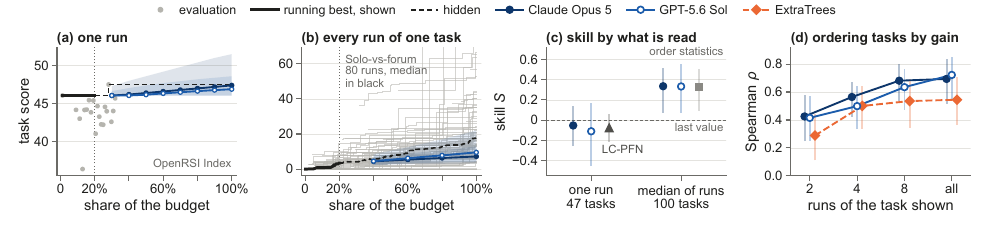}
\caption{Research-progress curves of research agents, read at 20\% of the budget. (a) One run of the OpenRSI Index: its scored evaluations, the running best shown to the forecaster and hidden from it, and the forecasts (median and 10--90\% band). (b) Every run of one task, their median, and the forecasts of that median. (c) Skill $S$ on a single run and on the median of a task's runs (\opus{} on 46 and 98 tasks). (d) Spearman correlation between the forecast and the realized remaining gain of a task, against the number of its runs shown. Bars: 95\% intervals over tasks. Panels (a) and (b) draw the view of their source whose error relative to the last value is the median for \sol{}; three failed evaluations in (a) and the seven best-scoring runs in (b)~fall~outside~the~frame.}
\label{fig:rsi-curves}
\end{figure}

An autonomous research agent produces a curve of its own, the best score it has reached so far over the hours or iterations of its budget.
We read these research-progress curves (Figure~\ref{fig:rsi-curves}a,\,b) under the protocol of \bench{}, on agent runs from 13 public sources, among them the OpenRSI Index~\citep{openrsi2026index}, with the task as the independent unit (Appendix~\ref{app:rsi}).

\textbf{One run says little about its own future; the runs of a task together say where it has headroom.}
Read one run at a time from 5--20\% of its budget, no forecaster improves on the last value across the eight sources that publish every evaluation (47 test tasks, 46 for \opus{}): with the curve alone, $S=-0.11$ $[-0.45, 0.17]$ for \sol{} and $-0.05$ $[-0.26, 0.14]$ for \opus{}, against $-0.08$ $[-0.21, 0.06]$ for LC-PFN (Figure~\ref{fig:rsi-curves}c).
The task description and the configuration do not help: the full context raises \sol{}'s error by 27\% $[18, 41]$, and the configuration of a run on another task serves as well as the true one.
The runs of a task under one setting, read together as a fleet, carry more.
From the first 20\% of every run on a task, \sol{} and \opus{} forecast the median final score of the task's runs with $S=0.33$ $[0.07, 0.55]$ and $0.34$ $[0.08, 0.52]$ over the 100 test tasks of the six sources with at least three runs per task (98 for \opus{}), level with a baseline built on order statistics ($0.33$), and they order the tasks by the gain still to come with a Spearman correlation of $0.72$ $[0.55, 0.85]$ and $0.69$ $[0.52, 0.84]$, where the highest baseline, ExtraTrees trained on the development tasks of the other sources, reaches $0.54$ $[0.36, 0.71]$.
The ordering sharpens with the number of runs shown (for \sol{}, from $0.42$ with two runs to $0.63$ with eight; Figure~\ref{fig:rsi-curves}d), and it rests on the curves: from the description of the task alone the error on the same fleets is four to five times as large.
A research-progress curve is a staircase with long flat stretches, on which the models forecast gains that do not occur, so what carries over from training runs is the ordering of tasks by their headroom, which is the decision of where to spend the next agent runs.

\section{Discussion}
\label{sec:discussion}

\subsection{What it means if an LLM can read the tea leaves}
\label{sec:disc-meaning}

\paragraph{Three levels of knowing a component.}
A researcher who can predict a run's curve from its settings has compressed the runs they have watched into an expectation; an LLM that does the same has compressed public training reports, logs, and code into its weights, the record of the process that produced it.
Knowing a component of training has three levels: describing it, implementing it, and anticipating what it does to a run.
The first two are what a language model is usually asked for, an explanation of Adam or a correct implementation of it.
The third is what a researcher uses to decide whether a change deserves a run, and what a system that improves its own training would need, because proposing a change to an optimizer or a schedule is cheap and learning its effect by running it is not.
The curve is the supervision for this third level: every logged run pairs a set of design choices with their measured consequence, and such pairs accumulate with every experiment the field runs.

\paragraph{Where current models stand.}
Our results place current models on the path from the first level to the third.
For components that the public record covers well, the knowledge is present in qualitative form: the strongest models match settings to completed curves with 82--86\% accuracy and \sol{} orders a source's runs from the configuration alone (Appendix~\ref{app:extremes-details}), yet assignments derived from their forecasts are near chance (\S\ref{sec:main}).
The knowledge extends to a change that the curve does not yet show: from the description of a code change that acts after the observed prefix, \sol{} and \opus{} order all 36 pairs of runs, and \sol{} predicts the cost of the change as $+0.098$ nats against a measured $+0.116$ (\S\ref{sec:decisions}).
It can also be trained: under a thousand logged runs bring a 2B model level with \opus{} on loss curves (\S\ref{sec:training}).

\paragraph{From known components to untried designs.}
The next steps on this path are combinations of components that the record covers thinly, and then designs that no one has run, where a forecast cannot come from recall and has to come from the mechanism.
A model that forecasts an untried design can rank its own proposals before training them, which is the step of a self-improving research loop that costs the most compute today.
How far a model stands from that point depends on how much of a run is decided by its setup and how much of that the field has written down: on reward curves and on failures the models read far less, and these are also the outcomes least often written up.
\bench{} thus measures how much of the practitioner's tacit knowledge has reached the models, a boundary that will move as agents publish their own runs.

\subsection{Toward autonomous experiment management}
\label{sec:disc-management}

\paragraph{From forecasting a run to managing a sweep.}
Our results show that a forecast can already stand in for part of a researcher's judgment. Told what changed, the strongest LLMs order runs that the early curves do not separate, and on the benchmark's pretraining sweeps, continuing one run per sweep by \sol{}'s forecast ends, on average, within 1\% of the sweep's range from the best final loss, with 33.5\% and 42.6\% less training budget than keeping the current leaders needs (\S\ref{sec:decisions}).
However, each of these is a single decision taken once, at 20\% of the budget, and we believe it only scratches the surface of what managing experiments involves.
A researcher decides when to look at a run, stops some runs and extends others, relaunches a run with a changed setting, and spends the compute that a stopped run frees; each decision changes which runs exist to be read next.
Methods for hyperparameter optimization take the stopping decision from the curve alone~\citep{domhan2015speeding,swersky2014freezethaw}, and our results indicate both what an LLM adds to them and where it does not help yet: it anticipates a change that the curve does not show, but on the fine-tuning and RL sweeps its forecasts do not improve on keeping the current leaders.

\paragraph{A benchmark of the whole loop.}
We propose that the next benchmark score this loop and not the single forecast.
Its environment would be a sweep under a fixed compute budget, either replayed from logged runs, which is cheap and reproducible but offers only the runs that someone trained, or trained live, which lets the agent launch runs of its own at the cost of real compute, as in our 72-run sweep.
Its actions would be to launch, stop, continue, or modify a run, and its tools the agent harness of \S\ref{sec:agent} and the run database of \S\ref{sec:retrieval}.
It would report the best final result reached under the budget, the compute saved at equal quality, and the promising runs stopped by mistake, because stopping the eventual winner turns a saving into a loss.
Varying what the agent sees (the curve, the configuration, the code) and how early it must decide would show which of the abilities measured here carries the decision.
Our replay is the simplest instance of this design, with one decision at one checkpoint, and \bench{} measures the capability such an agent needs first, anticipating the outcome of~a~run.

\paragraph{What a lab can do today.}
Two of our results can be used before such a benchmark exists.
A lab can train an in-house reader, since under a thousand logged runs and about a GPU-hour give a 2B model that matches \opus{} on loss curves (\S\ref{sec:training}), or keep a database of its runs for a frontier model to read at query time, which cuts the error by 39\% with the 1{,}726-run database and no training (\S\ref{sec:retrieval}).
Both depend on how runs are kept: logged densely, since sparse logs cost every method most of its skill (\S\ref{sec:main}), and stored with their configuration, code, and lineage, as ours are.
The forecast is then best spent on the runs whose early curve does not already give the answer, and where it cannot separate the runs, the last value or a longer observation remains the fallback (Appendix~\ref{app:decisions}).

\subsection{Where reading runs out, and what a reader would need next}
\label{sec:disc-limits}

\paragraph{Rewards and failures.}
Reward curves during RL are hard to forecast (Figure~\ref{fig:main}: $0.43$ for the best LLM against $0.13$ for the best curve fitter), hard to learn from public runs (Figure~\ref{fig:reader}: every trained reader trails the last value on the RL test set), and a failing run is not reliably caught in advance (Figure~\ref{fig:extremes}c).
By loose analogy with the halting problem, we call the last of these the \emph{neural halting problem}.
Prior work characterizes instabilities~\citep{wortsman2024proxies,takase2025spike,molybog2023adam,chowdhery2023palm} or warns of divergence from internal activations and reference runs~\citep{huang2026preemptive,bergsma2026collapse}; our results suggest the scalar log alone carries little warning, and a better one may lie in the data, outputs, gradients, and checkpoints a researcher inspects next.
Mounting them in the agent harness of \S\ref{sec:agent}, for the model to check as a researcher would, would show whether the limit is the log or the reader.
Two reference points frame what can be forecast: the final losses of runs that differ only in the random seed have a standard deviation of 0.006 nats in our first controlled sweep, and the best smooth curve fitted in hindsight reaches $S=0.90$ on the 129 later runs, where the best forecaster, an ExtraTrees regressor trained on the benchmark's training split, reaches $0.76$ (Appendix~\ref{app:temporal}).

\paragraph{Toward an agentic data ladder.}
Scaling laws reach the same forecast by another route: they fit a functional form to a ladder of small runs designed in advance and extrapolate to the large one~\citep{kaplan2020scaling,hoffmann2022training,bhagia2025ladders}.
Recent work extends the ladder in two ways. Hinted scaling laws give small models privileged information, so that a benchmark on which they would score near zero still shows a trend to extrapolate~\citep{vanadrichem2026hinted}. Active scaling law discovery hands the ladder to an agent, which chooses the experiments to run, pays each one's recorded cost, and revises its law as outcomes arrive; the laws that current agents find this way extrapolate worse than those they fit when every result is given upfront~\citep{meng2026sldbench}.
Choosing the rungs is thus the open part, and a reader of runs could take it up: it forecasts from whatever runs exist, with their configurations, code, and curves, and its forecasts indicate which small run would be most informative to train next.
We see the reading of \bench{} growing in this direction, toward an \emph{agentic data ladder}, whose rungs (which data and which small runs to train) the agent chooses as the logs come in, and an \emph{agentic scaling law}, a forecast of the large run conditioned on everything the small ones recorded.
The replay of \S\ref{sec:decisions}, where continuing one run per sweep suffices on pretraining, is a first step.

\paragraph{Scope and limitations.}
Every run in \bench{} trains a language model and logs a scalar; the task and the scorer apply unchanged to any domain that logs runs, such as vision~\citep{alabdulmohsin2022revisiting}, robotics~\citep{lin2025robotscaling}, and chess~\citep{zhang2024chess}.
The curated runs come mostly from large, well-instrumented projects, so the benchmark likely overstates what forecasters achieve on typical runs; the community set shows the direction of that bias.
The runs also do not record why their authors ran them, and a prospective test inside a research loop comes next.

\section{Conclusion}
\label{sec:conclusion}
We built \bench{}, 682 real public runs, to test whether LLMs can read a training run from its first 20\% and the semantics of the experiment.
They can: the strongest LLMs match the best curve fitter on smooth losses and lead on noisy rewards.
What bounds the reading is the log, and what extends it is information, not effort: every method loses most of its skill on sparse community logs and no monitor reliably catches a failing run in advance, whereas related runs in the prompt cut the error on loss curves by 39\% and tools add little.
The knowledge behind the reading can be learned and already informs decisions: a small open model trained on public runs matches the best LLM on loss curves; in a sweep we trained, the strongest LLMs, told what changed, order runs that the early curves do not yet separate; and from research agents' runs on each task they rank the tasks by their remaining headroom.
The judgment a researcher applies before spending compute is thus measurable, partly learnable, and bounded by the log.
The benchmark, the scorer, the trained readers, and their training data are released; the model answers and the database of runs will~follow.

\subsubsection*{Acknowledgments}
We thank Yiming Zhang, Hongming Zhang, Lijun Yu, Fengyu Cai, Tianyu Chen, Shoubin Yu, Yuxiao Qu, Haoran Li, Melanie Sclar, and Timo Schick for helpful discussions.
We also thank Jieun Kim, Vijay Viswanathan, Christina Ma, Chenyang Yang, Eunsu Kim, Jessie Mindel, and Zheyuan Zhang of the WInE Lab; our mentors at Google DeepMind, Azade Nova, Hanie Sedghi, and Doina Precup; and fellow interns at Google, Weiwei Sun, Yu Feng, Romy Luo, Xunjian Yin, Wenhao Chai, Ang Cao, Shengcao Cao, and Sanxing Chen, for their feedback.
Xinran Zhao is supported by the Amazon AI PhD Fellowship.

\bibliography{references,additions_20260925}
\bibliographystyle{iclr2027_conference}

\clearpage
\appendix
\subsection*{AI use statement}
The project was conceived by the human authors. The data collection was designed by the authors and implemented with coding agents; the experiments were designed by the authors, with code partly written and jobs monitored by coding agents. The text was written by the authors and edited with AI assistance; every number and claim was verified by the authors.

\subsection*{Ethics statement}
All runs were collected from public sources, namely public experiment-tracker projects, public repositories on the Hugging Face Hub, and a public speedrun log archive, under the terms on which they were made public. Provenance (source and run identifier) is recorded per run; licenses were not audited run by run, and a license is recorded only for the speedrun archive (Appendix~\ref{app:data}). In the RL set, runs whose owners could not be tied to an institution with an associated paper or official repository were treated as requiring the authors' consent to redistribution and were excluded before any model was evaluated; the other sets were not screened for consent. Community runs are analyzed in aggregate and are not attributed to individual authors in the paper. The models' outputs are forecasts of training metrics and carry no personal or sensitive content.

\subsection*{Reproducibility statement}
Every number on the benchmark can be regenerated from the artifacts of the complete release; a supplementary document, an ancillary file of this preprint that is also part of the release, holds the tables, figure, examples, and input schemas that the appendix summarizes (its sections are numbered S1 to S9); for the research-agent curves of Appendix~\ref{app:rsi}, the requests derived from sources that state no license are not released.
The pretraining, fine-tuning, and RL test sets (150 runs each) were frozen with content hashes before any model was evaluated, and the selection rule and the requests of the community set (232 runs) were fixed before any model was run on it; Appendix~\ref{app:data} gives the sources, the filtering funnel, and the lineage isolation and its scope; Appendix~\ref{app:reader} reports the 22 related records in two reader pools and the check against a broader relation graph; Appendix~\ref{app:example-full} shows one complete question as the forecaster sees it, with its answer, Appendix~\ref{app:prompt-text} the exact prompts, and the supplementary document (\S S8--S9) further examples, the input schemas, and the agents' instructions.
Appendix~\ref{app:models} lists the identifier sent to each provider, the pinned commit of every locally served model and the provider identifier of every API-served model, including the three open-weight models served through a provider, the decoding settings, and the compute used; Appendix~\ref{app:baselines} gives the implementation, checkpoint, and input contract of every baseline, and Appendix~\ref{app:validity} the rule that decides whether an answer is scored.
The skill score is defined in \S\ref{sec:tasks} and worked through on an example in Appendix~\ref{app:metrics}; the paired contrasts behind the semantics, agent, reader-training, and database claims are tabulated with their resampling units in Appendix~\ref{app:significance}, and the observed-share, input-window, failure-monitoring, matching, and community-set results are tabulated per method in Appendices~\ref{app:extremes-details}, \ref{app:matching-task}, and~\ref{app:wild-details}; the reader is compared with the LLMs by unpaired per-stage scores (Appendix~\ref{app:reader-macro}).
The reader's training recipe, pools, untrained baselines, three-stage averages, and a summary of the ablations are in Appendix~\ref{app:reader} (the ablation tables in the supplementary document, \S S1), and the protocols of the agent, observed-share and failure-monitoring, and database analyses in Appendices~\ref{app:det-agent}, \ref{app:extremes-details}, and~\ref{app:det-retrieval}.
The benchmark, the scorer, and the trained reader adapters are released at the address on the first page; the request and answer records of every API and open-weight model, packaged with the hashes~above, will~follow.

\section{Implementation details}
\label{app:details}
This appendix gives what is needed to reproduce every number in the paper: the model identifiers, decoding settings, and compute, the implementation of each baseline of \S\ref{sec:baselines}, and the rules that decide whether an answer is valid.
Every pretraining and fine-tuning question is asked on two query grids, \textsc{Grid10} ($x=0.3,0.4,\dots,1.0$, the grid of \S\ref{sec:tasks} and the default) and \textsc{Grid20} ($x=0.4,0.6,0.8,1.0$); RL scores pool both grids.
Scores are aggregated over cells, one per stage and source: Marin, MLF, and AI2 pretraining, and Marin and HF fine-tuning (MLF = DCLM from MLFoundations, AI2 = OLMoE, HF = Open-R1; Table~\ref{tab:data}).

\subsection{Models and inference settings}
\label{app:models}

\begin{table}[t]
\centering\footnotesize
\caption{The 14 LLMs of the paper (Table~\ref{tab:main}), with the identifier actually sent to the provider. Every model received the same frozen requests (the \textsc{Curve} and \textsc{Full} requests were frozen on 16 September 2026, the \textsc{Curve+Config} requests on 18 September); \kimi{}, Qwen3.5-0.8B, Qwen3.5-4B, and Qwen3.5-9B were not run in the \textsc{Curve+Config} condition. Locally served models are pinned to a Hugging Face commit, whose first eight characters are given; the models served through Bedrock or the Gemini API, including the open-weight \kimi{}, Qwen3-Next-80B, and DeepSeek-V3.2, are identified by the provider's model identifier only.}
\label{tab:models}
\setlength{\tabcolsep}{3.5pt}
\scriptsize
\begin{tabular}{llll}
\toprule
Short name & Identifier & Provider & Decoding \\
\midrule
\sol & \texttt{us.openai.gpt-5.6-sol} & Bedrock & provider default \\
\luna & \texttt{us.openai.gpt-5.6-luna} & Bedrock & provider default \\
\opus & \texttt{global.anthropic.claude-opus-5} & Bedrock & provider default \\
Claude Haiku 4.5 & \texttt{global.anthropic.claude-haiku-4-5-20251001-v1:0} & Bedrock & provider default \\
\kimi & \texttt{global.moonshotai.kimi-k3} & Bedrock & provider default \\
Qwen3-Next-80B & \texttt{qwen.qwen3-next-80b-a3b} & Bedrock & provider default \\
DeepSeek-V3.2 & \texttt{deepseek.v3.2} & Bedrock & provider default \\
\gpro & \texttt{gemini-3.1-pro-preview} & Gemini API & thinking low \\
\gflash & \texttt{gemini-3.8-flash} & Gemini API & thinking low \\
Muse-Glimmer-30B & \texttt{meta-models/Muse-Glimmer-30B} (\texttt{a4e59da5}) & local, HF & greedy, reasoning low \\
Qwen3.5-0.8B & \texttt{Qwen/Qwen3.5-0.8B} (\texttt{2fc06364}) & local, vLLM & greedy, no thinking \\
Qwen3.5-2B & \texttt{Qwen/Qwen3.5-2B} (\texttt{15852e8c}) & local, vLLM & greedy, no thinking \\
Qwen3.5-4B & \texttt{Qwen/Qwen3.5-4B} (\texttt{851bf6e8}) & local, HF & greedy, no thinking \\
Qwen3.5-9B & \texttt{Qwen/Qwen3.5-9B} (\texttt{c2022362}) & local, HF & greedy, no thinking \\
\bottomrule
\end{tabular}
\end{table}

Table~\ref{tab:models} lists the 14 LLMs with the identifier sent to each provider.
All API models are called through the Bedrock Converse interface or the Gemini generative-language endpoint with a single sample per request and an output limit of 4{,}096 tokens.
They use provider-default sampling (no temperature, nucleus, stop, or seed parameter), except that the two Gemini models run with thinking set to low; intervals are conditional on the one answer per request.
The three open-weight models served through Bedrock (\kimi{}, Qwen3-Next-80B, and DeepSeek-V3.2) are called in the same way.
The five locally served models are decoded greedily with bf16 weights and a fixed seed; thinking is disabled for the four Qwen3.5 models, and Muse-Glimmer-30B runs with reasoning set to low.
The 0.8B and 2B models were served by vLLM 0.29.0 and the 4B, 9B, and 30B models by Hugging Face Transformers 5.17.0.
Run on the same 240 requests under both backends, Qwen3.5-9B gives identical parsed answers on 203 of the 227 that both parse, and its median forecasts differ by 0.021 on average (55 of 65 forecasting items identical).

\paragraph{Compute and release.}
Every local job in the paper runs on a single 80\,GB GPU (A100 or H100) on academic clusters; no run uses more than one GPU, and the reader recipe trains in about 1.2 GPU-hours (Appendix~\ref{app:recipe}).
The frozen requests, every raw answer, the scorer, and the trained reader adapters (LoRA weights on top of the pinned Qwen3.5 revisions) are packaged with their hashes; the requests, the scorer, and the adapters are released with the benchmark, and the raw answers will follow.

\subsection{Baselines and prior methods: implementation}
\label{app:baselines}

Every baseline sees the same prefix as the LLMs: on pretraining and fine-tuning runs, the ten window means $\bar y_{1:10}$ at 2\%, 4\%, \ldots, 20\% of the planned budget; on RL runs, the dense prefix as released, from which the parametric fits take the ten 2\% window means. The RL prefix comes in two forms: for 96 of the 150 runs it gives trailing 2\% window means every 0.25\% of the budget, and for the other 54 the raw logged points. The persistence forecast (the last value) repeats the last point of the prefix, which is $\bar y_{10}$ on pretraining and fine-tuning runs and on the 96 RL runs, and the raw last logged point on the other 54.
Methods that output quantiles are scored on their median; the others give a point forecast only.
Unless stated otherwise the method runs on CPU with a fixed seed, and every checkpoint is pinned by its hash.
Table~\ref{tab:baseline-impl} summarizes what each of the 12 methods without an LLM in Table~\ref{tab:main-b} sees.

\begin{table}[t]
\centering\scriptsize
\caption{What each non-LLM method sees and where it comes from. ``Prefix'' is the ten observed points; ``pool'' is our training runs, with the test run's sweep held out unless stated otherwise (Appendix~\ref{app:regimes-main}).}
\label{tab:baseline-impl}
\setlength{\tabcolsep}{3pt}
\begin{tabular}{@{}lllll@{}}
\toprule
Method & Code and weights & Input & Output & Trained on \\
\midrule
Last value & ours & last prefix point & point & -- \\
Power-law fit & ours (SciPy) & prefix & point & -- \\
AIC ensemble & ours (SciPy) & prefix & point & -- \\
LC-PFN & authors' code and checkpoint & prefix & quantiles & synthetic prior (authors) \\
FT-PFN (ifBO) & authors' code and checkpoint & prefix & quantiles & synthetic prior (authors) \\
Chronos-2 & released model & prefix, upsampled & quantiles & authors' pretraining \\
TimesFM-2.5 & released model & prefix & quantiles & authors' pretraining \\
NeuNeu-NoLoss & authors' architecture, our training & prefix & quantiles & 89 external runs \\
NCPL (released) & authors' checkpoint & eight config fields & point & authors' data \\
NCPL (retrained) & authors' code & config + prefix + budget & point & our pool \\
kNN on prefix shape, mean training shape & ours & prefix & point & our pool \\
\bottomrule
\end{tabular}
\end{table}

\paragraph{Parametric fits.}
The power law is $a+b\,x^{-c}$ with $b\ge0$ and $c\in[0.01,5]$, fitted by bounded least squares from four starting exponents, keeping the lowest squared error.
The AIC ensemble fits a power law, an exponential $a+b\,e^{-cx}$, and a log-linear law $a+b\ln x$, and averages their forecasts with weights $\exp(-\tfrac12\Delta\mathrm{AIC})$.
When every fit fails the forecast falls back to the last value.
On RL runs the fits are applied to the negated reward, so that all curves are fitted as decreasing.

\paragraph{LC-PFN.}
We use the authors' code (automl/lcpfn at commit \texttt{ba892f6}) and their released six-layer checkpoint (\texttt{pfn\_EPOCH1000\_EMSIZE512\_NLAYERS6\_NBUCKETS1000}).
The network was trained on synthetic learning curves indexed by epoch from 1 to 100, so we pass the fraction of the budget multiplied by 100 as the epoch.
Values are mapped into the prior's range with the authors' \texttt{pfn\_normalize}, using the minimum and maximum of the observed prefix as soft bounds and treating the curve as one to be minimized; RL rewards are negated first and the quantiles flipped back.
Two small changes were needed to load the checkpoint in a current PyTorch release: the encoder loop passes the explicit attention mask as the original code did, and the GELU layers use the exact form they were trained with.
The prior is saturating curves over epochs rather than LLM loss over tokens, and the soft bounds come from the prefix.

\paragraph{FT-PFN.}
We run the released ifBO checkpoint (\texttt{bopfn\_broken\_unisep\_\allowbreak 1000curves\_\allowbreak 10params\_\allowbreak 2M}) with a single curve in context, the run's own prefix, with time as the raw budget fraction.
Values are mapped to $(0,1)$ by a sigmoid centered on the prefix median and scaled by a robust spread, and mapped back through the logit.
The model also takes ten hyperparameter slots; we fill the ones that match a fixed list of names and set the rest to the neutral value 0.5, so in practice it forecasts from the curve.
Its 80\% intervals are too narrow on our runs.

\paragraph{Time-series foundation models.}
Chronos-2 (\texttt{amazon/chronos-2}) reads its input in patches of 16 points, so ten points would fill less than one patch; we therefore interpolate the prefix linearly onto a grid of 0.2\% of the budget (91 points) and forecast 400 steps to the end of the run.
TimesFM-2.5 (\texttt{google/timesfm-2.5-200m}) receives the ten raw points and forecasts 40 steps of 2\%; its context therefore sits inside a single input patch.
Both models are used zero-shot, and their 10th, 50th, and 90th percentile outputs are taken as the forecast.
On RL runs both receive the dense prefix interpolated onto 100 points.

\paragraph{NeuNeu-NoLoss.}
No trained checkpoint of this variant is released, so we trained the authors' architecture (six layers, width 512, about 19M parameters, five quantile heads) ourselves: 1{,}000 updates on 89 pretraining runs of an external source (billzid/predictable-scale), never on our runs.
The model reads only the history of the loss; it was trained with a standardization fitted on the external source and tested with each prefix standardized by its own mean and standard deviation.
The original paper predicts downstream accuracy, whereas we train it to predict the loss.

\paragraph{NCPL.}
The released checkpoint (\texttt{OptimizerStudy/NCPL-intermediate}) maps eight numeric configuration fields (model size, data size, learning rate, warmup, weight decay, batch size, number of steps, and the query fraction) to a residual on top of a Chinchilla-style loss offset; it has no way to mark a field as missing, so it can score only the 16 test runs that report all eight (supplementary document, \S S2).
The retrained NCPL uses the authors' code and a Qwen3-1.7B backbone, and is adapted to our task: its input is the signed-log numeric configuration, the ten prefix points standardized by their own mean and standard deviation, and the budget fraction of the query, and a scalar head is trained with squared error on the standardized target.
It trains for 100 updates of the head alone and 400 updates of the full model (AdamW, learning rates $5\times10^{-5}$ and $10^{-5}$), with sampling weights that balance sources, then runs, then query points, and no validation-based checkpoint selection.
It is retrained once per regime; in the sweep-held-out regime every test sweep is removed from the pool at once, and in the source-held-out regime the test cell's whole source is removed, one model per cell.
The community-run row and the ablations of \S\ref{sec:extreme} use one sweep-held-out model trained under this recipe, pool, and seed; on the ablations it is used as a fixed forecaster (the observed prefix is resampled to its ten input slots for shorter shares, its input is the 2\% window by construction, and as a failure monitor it is scored by the predeclared adverse-change rule used for LC-PFN), never~retrained~per~arm.

\paragraph{Pool-trained regressors.}
The two shape regressors predict each future value divided by the last observed value, as our reader does.
The nearest-neighbor regressor (kNN on prefix shape) averages the future ratios of the five training runs whose standardized prefixes are closest, and the mean training shape applies the average future ratio of the pool.
On pretraining and fine-tuning runs these two use the training split of the frozen release, with the test run's sweep removed per test run; on RL runs they use our training pool, with the target written as a change from the last value scaled by the prefix range, and the held-out variant removes the test run's whole project.

\paragraph{ExtraTrees.}
The additional analyses of Appendices~\ref{app:wild-details}, \ref{app:additional}, and~\ref{app:rsi} use one more regressor trained on past runs, extremely randomized trees~\citep{geurts2006extratrees}.
On the benchmark's loss and reward curves it maps the ten prefix values and their ratios to the last value, with the numeric configuration where stated, to the ratio of every future value to the last observed value; on the RL runs of the supplementary document (\S S7) the target is the change from the last value scaled by the range of the prefix, as for the shape regressors above.
It has 500 trees (300 for the fleets of Appendix~\ref{app:rsi}), a minimum leaf size of 2 (5 for the fleets), and one fixed seed, without tuning.
Its training runs are stated where it is used and never include the evaluated runs: the training split of the benchmark with all stages and sources pooled, for the community runs and the later runs (Appendices~\ref{app:wild-details} and~\ref{app:temporal}); the database with the question's sweep family held out (Appendix~\ref{app:matched}); the development projects, for the auxiliary curves of RL runs (supplementary document, \S S7); and, for the fleets of agent runs, the development tasks of the other sources, on scale-free statistics of the fleet's prefixes (Appendix~\ref{app:rsi}).
It is not part of the main comparison of Figure~\ref{fig:main}.

\subsection{Validity}
\label{app:validity}

\begin{table}[t]
\centering\small
\caption{Answers accepted per LLM, out of 1{,}200 forecasting requests (300 runs, two grids, \textsc{Curve} and \textsc{Full}) and 2{,}400 matching requests (the 300 pairs of the frozen pretraining and fine-tuning release, two orders, two conditions, two grids). \textsc{Curve+Config} answers are not included (their coverage is in Appendix~\ref{app:main-table}). Strict requires the exact output schema; lenient requires only that the numbers parse and that crossed quantiles can be sorted; answers that pass lenient parsing but are absurd (any median outside the observed prefix's range widened by five ranges on each side; see text) are removed afterwards, before scoring, and are counted in the column ``absurd'' (a subset of lenient).}
\label{tab:validity}
\setlength{\tabcolsep}{5pt}
\begin{tabular}{lccccc}
\toprule
 & \multicolumn{3}{c}{Forecasting} & \multicolumn{2}{c}{Matching} \\
\cmidrule(lr){2-4}\cmidrule(lr){5-6}
Model & strict & lenient & absurd & strict & lenient \\
\midrule
\gpro & 1198 & 1198 & 1 & 2392 & 2393 \\
\luna & 1196 & 1196 & 0 & 2390 & 2390 \\
\sol & 1192 & 1192 & 1 & 2398 & 2398 \\
Muse-Glimmer-30B & 1183 & 1198 & 4 & 2254 & 2399 \\
Claude Haiku 4.5 & 1150 & 1198 & 8 & 2395 & 2395 \\
\kimi & 1139 & 1145 & 1 & 2386 & 2396 \\
\gflash & 982 & 984 & 5 & 2372 & 2391 \\
Qwen3-Next-80B & 880 & 1200 & 0 & 2398 & 2400 \\
DeepSeek-V3.2 & 852 & 1163 & 19 & 2390 & 2392 \\
\opus & 822 & 1181 & 0 & 1993 & 2373 \\
Qwen3.5-0.8B & 341 & 610 & 0 & 1561 & 1930 \\
Qwen3.5-2B & 203 & 572 & 0 & 1077 & 1576 \\
Qwen3.5-9B & 166 & 1199 & 1 & 2145 & 2208 \\
Qwen3.5-4B & 89 & 242 & 0 & 2040 & 2118 \\
\bottomrule
\end{tabular}
\end{table}

Every error ratio in the paper is computed over the views a model answered validly (lenient); invalid answers are dropped rather than penalized.
Per-cell counts are the superscripts of Table~\ref{tab:main}; Qwen3.5-4B has no \textsc{Grid10} \textsc{Full} score because it answers no full-context view in the AI2~pretraining~cell.

\paragraph{Absurd outputs.}
After lenient parsing, a pretraining or fine-tuning forecast is removed as absurd when the median at any query point lies outside $[m-Kr,\,M+Kr]$ with $K=5$, where $m$ and $M$ are the smallest and largest of the observed prefix values and $r=M-m$; a constant prefix ($r=0$) is exempt, and none of the 600 frozen views has one.
The rule reads only the prefix and the forecast, never the target, and removes the whole answer.
It applies to the answers of every LLM, in every condition and grid, on the frozen pretraining and fine-tuning release (Tables~\ref{tab:main} and~\ref{tab:common-valid}), and to the reader, whose scorer uses the same rule; the methods without an LLM, and the RL, community-run, and failure-monitoring sets, are scored without it.
It removes 40 of the 14{,}278 lenient forecasts in Table~\ref{tab:validity}, 19 of them DeepSeek-V3.2's (1.6\% of its forecasts), 8 Claude Haiku 4.5's, and at most 5 of any other model's (none for \opus{} and \luna{}); in the \textsc{Curve+Config} condition it removes 3 answers of Claude Haiku 4.5, 8 of DeepSeek-V3.2, 1 of Muse-Glimmer-30B, and none of \opus{}, \sol{}, \luna{}, \gpro{}, \gflash{}, or Qwen3-Next-80B.
The robustness check of Appendix~\ref{app:common-valid}, which counts every invalid or missing answer as the last value ($R=1$), treats a removed absurd forecast the same way, so the scores reported there include these answers (for every model but \kimi{}, which has no paired contrast and one removed answer), at the error of the last value; this is lenient to the model, because an absurd forecast's own error is usually larger.
For the two models whose unfiltered scores were also computed, scoring the removed answers at their own error instead lowers $S$ by at most 0.007 (Claude Haiku 4.5, \textsc{Grid10} \textsc{Curve}; 0.002 for \sol{}, \textsc{Grid20} \textsc{Curve}).

Table~\ref{tab:validity} motivates lenient validity: strict validity measures obedience to an output schema and not forecasting.
Of the answers that the strict schema rejected and the lenient one accepted, 735 of \opus{}'s differ from a valid answer only in a missing or malformed rationale field, and 318 of Qwen3-Next-80B's are quantiles written in the wrong order.
Scoring only the questions that every model answers under the strict schema makes the denominator both small and unrepresentative: on \textsc{Grid20} it leaves 2 of the 25 AI2 pretraining questions, 10 of 25 on MLFoundations, and 946 of 1{,}800 information-condition pairs overall.
Under lenient validity the untrained Qwen3.5-0.8B, 2B, and 4B still give no parsable forecast on 49\%, 52\%, and 80\% of the forecasting requests; the tables include these models with those rates.
A valid answer can still be poor: of DeepSeek-V3.2's 287 curve-only pretraining answers, 81 invert the quantiles and 60 extrapolate the median far below every value in the prefix (0 and 7 for \sol{}), and its mean MAE there is 0.759 against~0.272~for~\sol{}.

\FloatBarrier

\section{Data collection and provenance}
\label{app:data}
Appendix~\ref{app:census} adds the composition of the four test sets by source, model scale, task type, and budget.

This appendix describes where the runs of \S\ref{sec:data} come from and how the pools were built: the sources, the funnel from collected runs to the training pool, and the scope of the lineage isolation.

\paragraph{Where the runs come from.}
We enumerated training runs from three kinds of source.
The tracker index records 19{,}650 identities across 236 collection-log files, including metadata-only and failed lookups; a log file can contain many attempts.
On the Hugging Face Hub, a capped listing returned 47{,}929 distinct models; 14{,}370 ungated repositories carried a trainer-state file after quantized, converted, and bulk-uploaded repositories were removed, and 4{,}503 were fetched and parsed.
The speedrun archive contributed 1{,}516 parsed log records with 1{,}476 distinct recorded run IDs.
Separately, the training pools start from 17{,}805 collected records (16{,}650 distinct pool IDs) before filtering. These inventories have different scopes and are not a deduplicated count of independent experiments.

\begin{figure}[t]
\centering
\includegraphics[width=\linewidth]{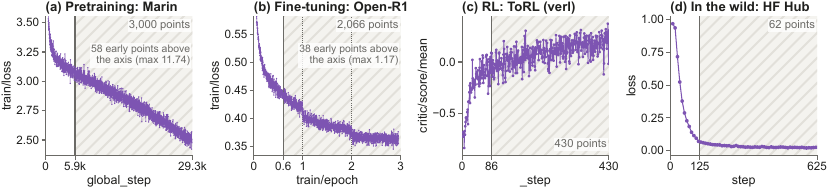}
\caption{One test run per set, as logged. Every point inside the planned budget is drawn; the vertical line marks 20\% of the budget, which the forecaster sees, and the shaded 80\% is what it predicts. (a) Marin pretraining, 29{,}329 steps (30.8B tokens); our collector's sampled-history request returned 3{,}000 of its per-step records. (b) Open-R1 fine-tuning of a 7B checkpoint for three epochs; the loss drops at each epoch boundary (dotted). (c) GRPO on math from a 1.5B checkpoint; the mean verifier score is noisy and rising. (d) A community LoRA fine-tuning run whose trainer state logs 62 points. In (a) and (b) the $y$ axis starts after the first 2\% of the budget, and the panel gives the number of earlier points above it.}
\label{fig:stage-examples}
\end{figure}

\paragraph{What a run of each stage looks like.}
Figure~\ref{fig:stage-examples} shows one test run per set as it was logged, with the 20\% prefix that a forecaster sees and the 80\% it must predict.
The four stages differ in what the prefix reveals.
A pretraining loss (a) falls steeply in the first few percent of the budget and then declines slowly and almost linearly for the rest; the prefix fixes the level, and the forecast turns on how fast the slow decline continues, which the learning-rate schedule and the token budget decide.
A fine-tuning loss (b) steps down at every epoch boundary as the model sees the data again; the prefix ends inside the first epoch, so the size of those steps has to be read from the number of epochs and the size of the checkpoint rather than from the curve.
An RL reward (c) rises through noise whose amplitude is as large as the trend over the prefix; the forecast is a statement about where the reward saturates, which the algorithm, the reward, and the budget govern.
A community run (d) logs sparsely, here 62 points over 625 steps, and most of the curve is flat after the prefix, so repeating the last value is already a strong forecast and the skill of every method is low (\S\ref{sec:main}).
The logging density behind these panels is typical of each set: the median pretraining and fine-tuning run reaches the 3{,}000-row cap of our collector's sampled-history request (83 of the 150 pretraining runs end within two rows of the cap, and 122 of the 150 fine-tuning runs at it), the median RL run logs 518 points (interquartile range 200--1{,}953), and the median community run 308 (133--1{,}003).

\paragraph{From runs to a training pool.}
Table~\ref{tab:funnel} gives the filters of the training pool. Its first six rows count records, whereas the final two count distinct pool IDs. Of the 6{,}780 eligible records, 5{,}707 mark the planned budget as known, so an eligible record is not always ready to forecast.
A \emph{sweep family} groups runs that differ only in the values a sweep varies (learning rate, seed, width, budget); its construction is given in Appendix~\ref{app:regimes-main}.
Removing every run that shares a sweep family with any test run costs 1{,}654 runs, 82\% of the 2{,}017 runs in the NCPL training and validation split that an NCPL-style predictor would otherwise train on; this is the step that matters for Appendix~\ref{app:regimes-main}.

\begin{table}[t]
\centering\small
\caption{Training-pool funnel, with the unit of each count stated. Of 3{,}024 clean records, 2{,}055 are marked forecast-eligible and 969 are retained for matching/raw use. The history/replicate filter leaves 1{,}835 records but 1{,}793 distinct pool IDs. This relaxed pool disables hard family/project caps and uses diversity sampling weights; extending family isolation yields 1{,}463 strict-pool IDs (1{,}105 forecast-eligible). The reference reader also uses a 340-record HF supplement; its training selection and isolation limitations are described in Appendix~\ref{app:reader}.}
\label{tab:funnel}
\begin{tabular}{llr}
\toprule
Step & Rule & count \\
\midrule
Universe & collected records before identity/quality filtering & 17{,}805 \\
Eligible & the source's protocol can be applied & 6{,}780 \\
Test isolation & drop test run keys and test lineage components & 6{,}552 \\
Family isolation (one source) & drop the sweep family of any test run & 4{,}898 \\
Clean & curve-quality thresholds & 3{,}024 \\
History/replicate filter & history hash; replicate cap (census exempt) & 1{,}835 \\
Relaxed pool (IDs) & distinct pool IDs; diversity weights, no hard caps & 1{,}793 \\
Strict pool (IDs) & extended family isolation across sources & 1{,}463 \\
\bottomrule
\end{tabular}
\end{table}

\paragraph{Lineage and exposure.}
Splits isolate whole recorded \emph{lineage components}, where a component joins runs connected by continuation, restart, or branch relations; intervals on pretraining and fine-tuning runs resample these components.
Author identity alone does not form a component, so two unrelated runs by the same person may fall on opposite sides of the split.
Validation targets 20\% of the non-test pool, rounded to whole components.
Selection uses identities and configurations, never forecasting performance, and no model was called before the sets were frozen.
The frozen pretraining and fine-tuning test release contains 300 forecasting runs and 300 matching pairs (150 per stage) drawn from 589 distinct identities, whose lineage closure reserves 609; there are no cross-split violations. Its matching pairs are the ones counted in the validity table (Table~\ref{tab:validity}, 2{,}400 matching requests per model).
Its exclusion sets, used to build the training pool, are 1{,}089 test run keys and 649 test lineage components.
Matching results (Appendix~\ref{app:matching-task}) use a separate 330-pair matching release (150 pretraining and 70 fine-tuning main pairs, and 110 pairs in a speedrun cell and two extension sets) in which no curve endpoint appears in two pairs of one set.
The wild set of 232 runs was drawn from 1{,}513 candidates after removing 32 with an author collision and 4 with a token collision against~the~other~sets.

\paragraph{Licensing and consent.}
Every run in the evaluation sets is recorded with its provenance, the source (experiment-tracker project, Hugging Face repository, or log archive) and the run identifier, and the released files are pinned by content hashes.
A license is recorded per source only for the speedrun log archive (MIT); experiment-tracker and Hugging Face runs carry no license field, so they are used under the terms on which their owners made them public, and their licenses were not audited run by run.
Consent is screened only in the RL set. Before any model result was seen, every run and matching pair was assigned an institution tier: tier~1 for institutional accounts and for the official logs of a published paper or framework, tier~2 for named researchers with a verifiable affiliation; runs from individual or unattributable accounts, and tier-2 runs without an associated paper or official repository, were flagged as requiring the authors' consent to redistribution and excluded.
The screen removed 5 run identities and 26 matching pairs from an earlier candidate set, and the freed slots were refilled by the same seeded draw; the 150 RL runs (138 tier~1, 12 tier~2) and the 120 matching pairs in the RL release (78 and 42) carry no consent flag. The matching set uses 240 distinct trajectories, of which 40 also occur in the 96-run forecasting set.

\paragraph{Scope of the exposure check.}
The check for exposure covers the files declared as used in development and the recorded relations: for the pretraining and fine-tuning prediction release it scanned 963 development-input files against the release's graph of 3{,}729 identities by exact key and found no match, so every record is marked ``no exact match in the declared scope''. The matching, RL, and community releases were not scanned.
The test set is restricted to runs logged densely enough to support a ten-point prefix, which biases it toward projects that log often.

\FloatBarrier

\section{Metric details and statistical significance}
\label{app:metrics}
This appendix shows how the skill score of \S\ref{sec:tasks} is computed on an example, how its edge cases are handled, and the paired significance tables behind the claims of \S\ref{sec:experiments}.

\subsection{Metric study}

\paragraph{How the skill score is computed, on an example.}
Take one cell with two runs.
On run A the forecast has an MAE of 0.05 over the query points and repeating the last value has 0.20, a ratio of 0.25; on run B the two MAEs are 0.30 and 0.20, a ratio of 1.5.
The cell's error ratio is the geometric mean of the run ratios, $\sqrt{0.25\times1.5}=0.61$, and the overall $R$ is the geometric mean of the five cells' ratios, so each cell counts equally however many runs it holds; the skill score is $S=1-R$, here $0.39$ for this cell.
$S$ is not the share of the summed error removed: the same two runs have a summed-error ratio of $(0.05+0.30)/(0.20+0.20)=0.875$, which would read as 12.5\% of the error removed.
We use the geometric mean so that one run with a large loss scale cannot dominate a cell, and so that halving the error on one run and doubling it on another cancels out.
The 95\% intervals resample lineage components with replacement (2{,}000 resamples, fixed seed) and recompute the whole aggregation, and an interval $[L,U]$ on $R$ is reported as $[1-U,1-L]$ on $S$.

\paragraph{Zero and invalid denominators.}
The ratio is undefined when repeating the last value is exact, which happens only when the future of a run is constant.
No pretraining or fine-tuning test run is affected.
In 17 of the 150 RL runs the logged series never leaves its last observed value; these runs (reward in 14, training loss in 3) are excluded from the RL score for every method alike, never replaced by a small constant, so RL scores are over the remaining 133 runs; Table~\ref{tab:rl-constant} lists which forecasts were exactly constant on them.
Invalid answers are handled as in Appendix~\ref{app:validity}; comparisons on common valid answers are in Appendix~\ref{app:common-valid}.

\begin{table}[t]
\centering\small
\caption{Agreement between the ranking of 26 methods under MAE and their ranking under five other metrics (12--13 quantile-emitting methods for weighted quantile loss), as Kendall's $\tau$ over the ten combinations of~grid~and~cell.}
\label{tab:metrics}
\begin{tabular}{lccl}
\toprule
Metric against MAE & mean $\tau$ & min $\tau$ & cell at the minimum \\
\midrule
RMSE & 0.968 & 0.945 & \textsc{Grid10} FT/HF \\
Error ratio against the last value & 0.898 & 0.834 & \textsc{Grid10} FT/HF \\
Final-point absolute error & 0.891 & 0.822 & \textsc{Grid10} PT/AI2 \\
Final-point relative error & 0.770 & 0.237 & \textsc{Grid20} FT/HF \\
Weighted quantile loss & 0.678 & 0.256 & \textsc{Grid20} FT/HF \\
\bottomrule
\end{tabular}
\end{table}

Which metric we report could change who wins, so we ranked 26 methods under six metrics in each of the ten cells (Table~\ref{tab:metrics}): LLM entries under \textsc{Curve} and \textsc{Full}, retrained NCPL, LC-PFN, NeuNeu-NoLoss, Chronos-2, and simple baselines (the last value, linear extrapolations, prefix statistics, the parametric fits and their AIC ensemble, and the two shape regressors).
RMSE, error ratio, and absolute final-point error rank methods essentially as MAE does, with mean $\tau$ between 0.89 and 0.97 and no cell below 0.82, although the top three methods are the same under all four metrics in only three of the ten cells.
Relative final-point error and weighted quantile loss reorder methods on the Hugging Face fine-tuning cell ($\tau=0.237$ and $0.256$): several of its runs end near zero, which makes the relative error unstable, and weighted quantile loss is defined only on the 12 or 13 methods that~emit~quantiles.

\subsection{Statistical significance}
\label{app:significance}
Tables~\ref{tab:significance}--\ref{tab:significance-c} report paired differences in $S$ between two forecasters scored on the same questions, restricted to questions both answered validly; a positive $\Delta S$ favors the first.
The 95\% intervals are percentile bootstrap intervals from 2{,}000 resamples of lineage components (pretraining and fine-tuning), runs (RL), or owners (wild), and from 10{,}000 resamples for retrieval; they are not adjusted for multiplicity.
Untrained-reader scores rest on the answers the untrained model produced validly.
We report intervals, not $p$-values.

\begin{table}[t]
\centering\scriptsize
\setlength{\tabcolsep}{3pt}
\caption{Paired contrasts for the benchmark (\S\ref{sec:main}): \textsc{Grid10} for PT and FT, both grids for RL. $\Delta S$ is the first forecaster's skill minus the second's; $n$ is the number of paired questions. Intervals are unadjusted 95\%~bootstrap~intervals.}
\label{tab:significance}
\begin{tabular}{llrlr}
\toprule
Contrast & Stage & $\Delta S$ & 95\% CI & $n$ \\
\midrule
\multicolumn{5}{l}{\textit{\S\ref{sec:main}: strongest LLMs (\textup{\textsc{Full}}) minus LC-PFN}} \\
\opus{} & PT & $0.03$ & $[-0.03, 0.11]$ & 147 \\
\opus{} & FT & $0.008$ & $[-0.099, 0.106]$ & 147 \\
\opus{} & RL & $0.29$ & $[0.18, 0.41]$ & 251 \\
\sol{} & PT & $0.000$ & $[-0.066, 0.064]$ & 149 \\
\sol{} & FT & $0.10$ & $[0.05, 0.16]$ & 148 \\
\sol{} & RL & $0.21$ & $[0.09, 0.34]$ & 263 \\
\midrule
\multicolumn{5}{l}{\textit{\S\ref{sec:main}: \opus{} minus \sol{} (\textup{\textsc{Full}})}} \\
 & PT & $0.03$ & $[-0.02, 0.09]$ & 146 \\
 & FT & $-0.09$ & $[-0.20, 0.00]$ & 145 \\
 & RL & $0.08$ & $[0.01, 0.15]$ & 250 \\
\midrule
\multicolumn{5}{l}{\textit{\S\ref{sec:main}: semantics, \textup{\textsc{Full}} minus \textup{\textsc{Curve}}}} \\
\opus{} & PT & $0.01$ & $[-0.04, 0.07]$ & 145 \\
\opus{} & FT & $-0.19$ & $[-0.29, -0.10]$ & 143 \\
\opus{} & RL & $0.08$ & $[0.02, 0.14]$ & 241 \\
\sol{} & PT & $0.03$ & $[-0.02, 0.08]$ & 149 \\
\sol{} & FT & $0.06$ & $[0.02, 0.11]$ & 148 \\
\sol{} & RL & $0.009$ & $[-0.048, 0.066]$ & 263 \\
\luna{} & PT & $0.11$ & $[0.05, 0.18]$ & 150 \\
\luna{} & FT & $-0.04$ & $[-0.08, 0.00]$ & 148 \\
\gpro{} & PT & $0.07$ & $[0.00, 0.12]$ & 149 \\
\gpro{} & FT & $-0.07$ & $[-0.12, -0.02]$ & 149 \\
\gflash{} & PT & $0.05$ & $[0.00, 0.11]$ & 110 \\
\gflash{} & FT & $-0.04$ & $[-0.11, 0.02]$ & 81 \\
Claude Haiku 4.5 & PT & $0.17$ & $[0.09, 0.26]$ & 150 \\
Claude Haiku 4.5 & FT & $0.12$ & $[0.07, 0.17]$ & 147 \\
Muse-Glimmer-30B & PT & $-0.13$ & $[-0.24, -0.03]$ & 149 \\
Muse-Glimmer-30B & FT & $-0.12$ & $[-0.19, -0.06]$ & 148 \\
\bottomrule
\end{tabular}
\end{table}

\begin{table}[t]
\centering\scriptsize
\setlength{\tabcolsep}{3pt}
\caption{Further paired contrasts, continuing Table~\ref{tab:significance}. Untrained readers receive the same reader prompt; PT+FT pools the two stages. Intervals are unadjusted 95\% bootstrap intervals, except in the trained-minus-untrained block, which gives the mean $\Delta S$ over three training seeds with the range over seeds in brackets; every per-seed 95\% interval in that block lies above zero (lowest lower bound 0.19), and $n$ is the per-seed number of~paired~questions.}
\label{tab:significance-b}
\begin{tabular}{llrlr}
\toprule
Contrast & Stage & $\Delta S$ & 95\% CI & $n$ \\
\midrule
\multicolumn{5}{l}{\textit{\S\ref{sec:main}: semantics, \textup{\textsc{Full}} minus \textup{\textsc{Curve}} (other open-weight models)}} \\
DeepSeek-V3.2 & PT & $-0.28$ & $[-0.56, 0.03]$ & 138 \\
DeepSeek-V3.2 & FT & $-0.62$ & $[-0.88, -0.38]$ & 141 \\
Qwen3-Next-80B & PT & $0.03$ & $[-0.03, 0.10]$ & 150 \\
Qwen3-Next-80B & FT & $-0.02$ & $[-0.06, 0.01]$ & 150 \\
Kimi K3 & PT & $-0.04$ & $[-0.11, 0.03]$ & 136 \\
Kimi K3 & FT & $-0.07$ & $[-0.11, -0.02]$ & 131 \\
\midrule
\multicolumn{5}{l}{\textit{\S\ref{sec:agent}: agent minus single call}} \\
\opus{}, local tools & PT & $-0.01$ & $[-0.06, 0.03]$ & 95 \\
\opus{}, local tools & FT & $0.09$ & $[0.04, 0.14]$ & 99 \\
\opus{}, local tools & RL & $0.04$ & $[-0.11, 0.19]$ & 43 \\
\opus{}, local tools + search & PT & $0.03$ & $[-0.02, 0.08]$ & 96 \\
\opus{}, local tools + search & FT & $0.11$ & $[0.05, 0.17]$ & 98 \\
\opus{}, local tools + search & RL & $0.07$ & $[-0.06, 0.20]$ & 43 \\
\sol{}, local tools & PT & $-0.03$ & $[-0.09, 0.03]$ & 80 \\
\sol{}, local tools & FT & $-0.06$ & $[-0.07, -0.02]$ & 83 \\
\sol{}, local tools & RL & $-0.09$ & $[-0.28, 0.10]$ & 42 \\
\sol{}, local tools + search & PT & $-0.04$ & $[-0.10, 0.02]$ & 81 \\
\sol{}, local tools + search & FT & $-0.004$ & $[-0.055, 0.060]$ & 86 \\
\sol{}, local tools + search & RL & $-0.09$ & $[-0.26, 0.08]$ & 43 \\
\midrule
\multicolumn{5}{l}{\textit{\S\ref{sec:training}: trained reader (8{,}727 runs) minus untrained; mean and range over three seeds}} \\
Qwen3.5-2B, config & PT & $12.81$ & $[12.80, 12.82]$ & 131 \\
Qwen3.5-2B, config & FT & $1.78$ & $[1.65, 1.86]$ & 148--149 \\
Qwen3.5-2B, config & RL & $3.20$ & $[3.05, 3.41]$ & 74 \\
Qwen3.5-2B, config + semantics & PT & $6.42$ & $[6.41, 6.44]$ & 44 \\
Qwen3.5-2B, config + semantics & FT & $0.53$ & $[0.35, 0.66]$ & 108--111 \\
Qwen3.5-2B, config + semantics & RL & $3.29$ & $[3.20, 3.41]$ & 77 \\
Qwen3.5-9B, config & PT & $20.24$ & $[20.20, 20.27]$ & 122 \\
Qwen3.5-9B, config & FT & $1.82$ & $[1.77, 1.86]$ & 150 \\
Qwen3.5-9B, config & RL & $2.46$ & $[2.12, 2.64]$ & 164 \\
Qwen3.5-9B, config + semantics & PT & $16.10$ & $[16.06, 16.14]$ & 110 \\
Qwen3.5-9B, config + semantics & FT & $1.23$ & $[1.20, 1.25]$ & 130--131 \\
Qwen3.5-9B, config + semantics & RL & $2.74$ & $[2.68, 2.84]$ & 162 \\
\midrule
\multicolumn{5}{l}{\textit{\S\ref{sec:training}: trained reader (8{,}727 runs, mean of 3 seeds), config + semantics minus config}} \\
Qwen3.5-2B reader & PT+FT & $0.002$ & $[-0.014, 0.017]$ & 296 \\
Qwen3.5-2B reader & RL & $-0.14$ & $[-0.28, -0.03]$ & 164 \\
Qwen3.5-2B reader & Wild & $-0.03$ & $[-0.07, 0.01]$ & 464 \\
Qwen3.5-9B reader & PT+FT & $0.02$ & $[0.01, 0.03]$ & 299 \\
Qwen3.5-9B reader & RL & $0.32$ & $[0.17, 0.50]$ & 164 \\
Qwen3.5-9B reader & Wild & $0.07$ & $[0.03, 0.11]$ & 464 \\
\bottomrule
\end{tabular}
\end{table}

\begin{table}[t]
\centering\scriptsize
\setlength{\tabcolsep}{3pt}
\caption{Further paired contrasts, continuing Table~\ref{tab:significance-b}. Retrieval rows report the geometric mean of the ratio of errors with and without retrieval instead of $\Delta S$. Intervals are unadjusted 95\% bootstrap intervals.}
\label{tab:significance-c}
\begin{tabular}{llrlr}
\toprule
Contrast & Stage & $\Delta S$ or ratio & 95\% CI & $n$ \\
\midrule
\multicolumn{5}{l}{\textit{\S\ref{sec:retrieval}: retrieval over none, geometric-mean error ratio ($<1$ favors retrieval)}} \\
\opus{}, 1{,}726 runs & PT+FT & $0.58$ & $[0.47, 0.69]$ & 117 \\
\sol{}, 17 runs (1\%) & PT+FT & $0.90$ & $[0.78, 1.04]$ & 122 \\
\sol{}, 5\% of library & PT+FT & $0.91$ & $[0.77, 1.07]$ & 123 \\
\sol{}, 25\% of library & PT+FT & $0.71$ & $[0.57, 0.88]$ & 123 \\
\sol{}, 1{,}726 runs & PT+FT & $0.61$ & $[0.47, 0.76]$ & 123 \\
\sol{}, 1{,}726 runs, $k{=}1$ & PT+FT & $0.75$ & $[0.61, 0.93]$ & 123 \\
\sol{}, 1{,}726 runs, $k{=}5$ & PT+FT & $0.63$ & $[0.49, 0.80]$ & 123 \\
\sol{}, BM25 & PT+FT & $0.58$ & $[0.44, 0.74]$ & 123 \\
\sol{}, prefix-shape kNN & PT+FT & $0.62$ & $[0.48, 0.78]$ & 123 \\
\sol{}, random runs & PT+FT & $0.97$ & $[0.84, 1.13]$ & 123 \\
\sol{}, BM25, own family allowed & PT+FT & $0.19$ & $[0.13, 0.28]$ & 99 \\
\luna{}, 1{,}726 runs & PT+FT & $0.60$ & $[0.48, 0.75]$ & 124 \\
\luna{}, BM25 & PT+FT & $0.54$ & $[0.41, 0.70]$ & 124 \\
\luna{}, prefix-shape kNN & PT+FT & $0.64$ & $[0.49, 0.82]$ & 123 \\
\luna{}, random runs & PT+FT & $0.77$ & $[0.64, 0.92]$ & 124 \\
\luna{}, BM25, own family allowed & PT+FT & $0.28$ & $[0.21, 0.39]$ & 123 \\
\midrule
\multicolumn{5}{l}{\textit{Wild: LLM (\textup{\textsc{Full}}) minus LC-PFN, core-211 cohort (Appendix~\ref{app:det-wild})}} \\
\opus{} & Wild & $0.24$ & $[0.07, 0.42]$ & 211 \\
\sol{} & Wild & $0.23$ & $[0.10, 0.39]$ & 211 \\
\luna{} & Wild & $0.21$ & $[0.09, 0.36]$ & 211 \\
\bottomrule
\end{tabular}
\end{table}

\FloatBarrier

\section{Main results: full tables}
\label{app:full}
This appendix gives the tables behind Figure~\ref{fig:main} and \S\ref{sec:main}: per-cell results, the comparison on common valid answers, the reinforcement-learning and community-run sets, and the trained predictors under the three held-out regimes.

\subsection{Main table}
\label{app:main-table}
Tables~\ref{tab:main} and~\ref{tab:main-b} give the per-cell MAE and the skill score $S$ of every method on the \textsc{Grid10} query points, $x=0.3,0.4,\dots,1.0$; Table~\ref{tab:main} holds the zero-shot LLMs under the three information conditions, and Table~\ref{tab:main-b} the methods without an LLM and the predictors trained on past runs.
MAE is in the loss units of each source and is read down a column; $S$ is comparable across rows.
Every LLM and reader entry of Tables~\ref{tab:main}, \ref{tab:main-b}, and~\ref{tab:common-valid} is computed after the filter for absurd outputs (Appendix~\ref{app:validity}), so the tables agree on every shared entry.
The \reader{} rows of Tables~\ref{tab:main-b} and~\ref{tab:common-valid} are one training run of the reference recipe (Appendix~\ref{app:reader}), which scores $S=0.675$ on pretraining and fine-tuning and $-0.311$ on RL; Table~\ref{tab:reader-macro} gives the mean over three training seeds on the same pool ($0.659$ and $-0.332$).

\begin{table}[t]
\centering\footnotesize
\providecommand{\na}{\textsc{n/a}}
\setlength{\tabcolsep}{3pt}\renewcommand{\arraystretch}{0.94}
\caption{Curve prediction on the frozen test release, \textsc{Grid10}: zero-shot LLMs (methods without an LLM in Table~\ref{tab:main-b}). MAE per cell in the loss units of each source (superscript: scored views when below the cell size 100/25/25/100/50) and skill score $S$ with 95\% intervals over lineage components. Sources: MLF = DCLM (MLFoundations), AI2 = OLMoE, HF = Open-R1 (Table~\ref{tab:data}); PT+FT pools the five pretraining and fine-tuning cells. Every entry excludes absurd outputs (Appendix~\ref{app:validity}). Bold: best zero-shot entry in a column; underline: best across both halves. \na{}: not run, or no valid answer in some cell. $^{\ddagger}$\textsc{Curve+Config} requests answered: 598 of 600 (DeepSeek-V3.2) and 599 of 600 (Qwen3-Next-80B).}
\label{tab:main}
\begin{tabular}{@{}l ccc cc cc@{}}
\toprule
 & \multicolumn{3}{c}{Pretraining MAE} & \multicolumn{2}{c}{Fine-tuning MAE} & & \\
\cmidrule(lr){2-4}\cmidrule(lr){5-6}
Model, condition & Marin & MLF & AI2 & Marin & HF & PT+FT $S\uparrow$ & RL $S\uparrow$ \\
\midrule
\multicolumn{8}{@{}l}{\opus{}}\\
\quad \textsc{Curve} & 0.170$^{98}$ & 0.191 & 0.404 & 0.018$^{98}$ & \textbf{0.057$^{47}$} & \textbf{0.706} {\tiny [0.67,0.74]} & 0.326 {\tiny [0.19,0.44]} \\
\quad \textsc{Curve+Config} & 0.158$^{96}$ & 0.190 & 0.290$^{24}$ & 0.020$^{98}$ & 0.071$^{48}$ & 0.704 {\tiny [0.66,0.74]} & \na \\
\quad \textsc{Full} & \textbf{0.150$^{99}$} & 0.213$^{24}$ & \textbf{0.216$^{24}$} & 0.022$^{97}$ & 0.119 & 0.648 {\tiny [0.61,0.68]} & \underline{\textbf{0.427}} {\tiny [0.33,0.51]} \\
\multicolumn{8}{@{}l}{\sol{}}\\
\quad \textsc{Curve} & 0.234 & 0.208 & 0.428 & 0.031 & 0.078 & 0.612 {\tiny [0.57,0.66]} & 0.335 {\tiny [0.23,0.43]} \\
\quad \textsc{Curve+Config} & 0.222 & 0.190 & 0.447 & 0.029 & 0.074 & 0.623 {\tiny [0.58,0.66]} & \na \\
\quad \textsc{Full} & 0.193$^{99}$ & 0.228 & 0.411 & 0.028$^{98}$ & 0.069 & 0.657 {\tiny [0.61,0.70]} & 0.346 {\tiny [0.25,0.44]} \\
\multicolumn{8}{@{}l}{\luna{}}\\
\quad \textsc{Curve} & 0.264 & 0.202 & 0.457 & 0.032 & 0.083 & 0.574 {\tiny [0.52,0.62]} & 0.180 {\tiny [0.07,0.28]} \\
\quad \textsc{Curve+Config} & 0.225 & 0.227 & 0.379 & 0.036 & 0.074 & 0.589 {\tiny [0.54,0.63]} & \na \\
\quad \textsc{Full} & 0.212 & \textbf{0.159} & 0.320 & 0.041$^{99}$ & 0.076$^{49}$ & 0.632 {\tiny [0.59,0.67]} & 0.259 {\tiny [0.16,0.35]} \\
\multicolumn{8}{@{}l}{\gpro{}}\\
\quad \textsc{Curve} & 0.249$^{99}$ & 0.316 & 0.441 & 0.020 & 0.074$^{49}$ & 0.606 {\tiny [0.56,0.65]} & 0.012 {\tiny [-0.17,0.16]} \\
\quad \textsc{Curve+Config} & 0.219 & 0.324 & 0.430 & 0.017 & 0.068 & 0.614 {\tiny [0.57,0.66]} & \na \\
\quad \textsc{Full} & 0.198 & 0.278 & 0.350 & 0.032 & 0.062 & 0.610 {\tiny [0.56,0.65]} & -0.154 {\tiny [-0.44,0.06]} \\
\multicolumn{8}{@{}l}{\gflash{}}\\
\quad \textsc{Curve} & 0.216$^{82}$ & 0.336$^{17}$ & 0.480$^{18}$ & 0.023$^{65}$ & 0.075$^{35}$ & 0.598 {\tiny [0.54,0.65]} & 0.293 {\tiny [0.18,0.39]} \\
\quad \textsc{Curve+Config} & 0.175$^{76}$ & 0.290$^{13}$ & 0.471$^{17}$ & \textbf{0.015$^{47}$} & 0.060$^{26}$ & 0.650 {\tiny [0.60,0.69]} & \na \\
\quad \textsc{Full} & 0.176$^{92}$ & 0.315$^{23}$ & 0.413$^{23}$ & 0.030$^{84}$ & 0.063$^{43}$ & 0.604 {\tiny [0.55,0.65]} & 0.242 {\tiny [0.11,0.35]} \\
\multicolumn{8}{@{}l}{Claude Haiku 4.5}\\
\quad \textsc{Curve} & 0.254 & 0.408 & 0.572 & 0.049 & 0.094$^{47}$ & 0.463 {\tiny [0.40,0.52]} & -0.045 {\tiny [-0.19,0.09]} \\
\quad \textsc{Curve+Config} & 0.185 & 0.382 & 0.525 & 0.041 & 0.073$^{49}$ & 0.550 {\tiny [0.49,0.60]} & \na \\
\quad \textsc{Full} & 0.229 & 0.248 & 0.428 & 0.032 & 0.078$^{47}$ & 0.608 {\tiny [0.55,0.66]} & 0.057 {\tiny [-0.06,0.17]} \\
\multicolumn{8}{@{}l}{DeepSeek-V3.2}\\
\quad \textsc{Curve} & 0.693$^{97}$ & 0.874$^{21}$ & 1.005$^{23}$ & 0.127$^{98}$ & 0.094$^{47}$ & -0.035 {\tiny [-0.19,0.10]} & -0.007 {\tiny [-0.15,0.11]} \\
\quad \textsc{Curve+Config}$^{\ddagger}$ & 0.806$^{98}$ & 0.980$^{23}$ & 0.943$^{24}$ & 0.126$^{98}$ & 0.105$^{48}$ & -0.191 {\tiny [-0.33,-0.05]} & \na \\
\quad \textsc{Full} & 0.701$^{99}$ & 0.897$^{24}$ & 0.973$^{23}$ & 0.212$^{98}$ & 0.132$^{46}$ & -0.467 {\tiny [-0.64,-0.30]} & -0.113 {\tiny [-0.28,0.03]} \\
\multicolumn{8}{@{}l}{Qwen3-Next-80B}\\
\quad \textsc{Curve} & 0.283 & 0.230 & 0.366 & 0.047 & 0.075 & 0.540 {\tiny [0.50,0.58]} & -0.013 {\tiny [-0.12,0.10]} \\
\quad \textsc{Curve+Config}$^{\ddagger}$ & 0.259 & 0.254 & 0.288 & 0.049 & 0.078$^{49}$ & 0.545 {\tiny [0.51,0.58]} & \na \\
\quad \textsc{Full} & 0.245 & 0.202 & 0.328 & 0.050 & 0.074 & 0.553 {\tiny [0.52,0.58]} & -0.037 {\tiny [-0.15,0.08]} \\
\multicolumn{8}{@{}l}{Muse-Glimmer-30B}\\
\quad \textsc{Curve} & 0.221 & 0.431 & 0.407$^{24}$ & 0.027 & 0.064$^{49}$ & 0.580 {\tiny [0.53,0.63]} & 0.125 {\tiny [-0.02,0.25]} \\
\quad \textsc{Curve+Config} & 0.246 & 0.439 & 0.489 & 0.032$^{99}$ & 0.064$^{49}$ & 0.539 {\tiny [0.49,0.59]} & \na \\
\quad \textsc{Full} & 0.264 & 0.447 & 0.478 & 0.050$^{99}$ & 0.073$^{49}$ & 0.446 {\tiny [0.37,0.51]} & 0.102 {\tiny [-0.03,0.21]} \\
\multicolumn{8}{@{}l}{Qwen3.5-9B}\\
\quad \textsc{Curve} & 0.368 & 0.460 & 0.566 & 0.062 & 0.070 & 0.379 {\tiny [0.30,0.45]} & -0.239 {\tiny [-0.41,-0.09]} \\
\quad \textsc{Full} & 0.278$^{99}$ & 0.421 & 0.459 & 0.069 & 0.078 & 0.415 {\tiny [0.34,0.48]} & -0.155 {\tiny [-0.28,-0.04]} \\
\multicolumn{8}{@{}l}{Qwen3.5-2B}\\
\quad \textsc{Curve} & 0.306$^{87}$ & 0.199$^{24}$ & 0.315 & 0.055$^{95}$ & 0.129$^{42}$ & 0.480 {\tiny [0.43,0.53]} & \na \\
\quad \textsc{Curve+Config} & 0.277$^{98}$ & 0.207 & 0.340$^{20}$ & 0.062$^{96}$ & 0.136 & 0.454 {\tiny [0.39,0.51]} & \na \\
\quad \textsc{Full} & 0.665$^{19}$ & 0.388$^{5}$ & 0.247$^{5}$ & 0.048$^{26}$ & 0.281$^{4}$ & 0.306 {\tiny [0.13,0.42]} & \na \\
\multicolumn{8}{@{}l}{Qwen3.5-4B}\\
\quad \textsc{Curve} & 0.346$^{49}$ & 0.516$^{10}$ & 1.273$^{2}$ & 0.079$^{16}$ & 0.058$^{21}$ & 0.136 {\tiny [0.01,0.25]} & \na \\
\quad \textsc{Full} & 0.281$^{17}$ & 0.638$^{4}$ & \na & 0.036$^{6}$ & 0.074$^{8}$ & \na & \na \\
\multicolumn{8}{@{}l}{Qwen3.5-0.8B}\\
\quad \textsc{Curve} & 0.797 & 0.840 & 0.838$^{23}$ & 0.104 & 0.223$^{41}$ & -0.371 {\tiny [-0.49,-0.27]} & \na \\
\quad \textsc{Full} & \na & 4.042$^{15}$ & \na & \na & 0.431$^{1}$ & \na & \na \\
\multicolumn{8}{@{}l}{\kimi{}}\\
\quad \textsc{Curve} & 0.249$^{98}$ & 0.202$^{24}$ & 0.303$^{24}$ & 0.029$^{99}$ & 0.075$^{47}$ & 0.658 {\tiny [0.62,0.70]} & 0.264 {\tiny [0.17,0.35]} \\
\quad \textsc{Full} & 0.200$^{93}$ & 0.267$^{22}$ & 0.323$^{24}$ & 0.037$^{91}$ & 0.069$^{44}$ & 0.608 {\tiny [0.56,0.65]} & 0.229 {\tiny [0.13,0.32]} \\
\bottomrule
\end{tabular}

\end{table}

\begin{table}[t]
\centering\footnotesize
\providecommand{\na}{\textsc{n/a}}
\setlength{\tabcolsep}{2pt}\renewcommand{\arraystretch}{1.05}
\caption{Curve prediction on the frozen test release, \textsc{Grid10} ($x=0.3,\dots,1.0$): methods without an LLM and predictors trained on past runs; columns as in Table~\ref{tab:main}. The last block has seen the sweep of the test run in training, which amounts to within-sweep interpolation (Appendix~\ref{app:regimes-main}), and is not part of the main comparison.}
\label{tab:main-b}
\begin{tabular}{@{}l ccc cc cc@{}}
\toprule
 & \multicolumn{3}{c}{Pretraining MAE} & \multicolumn{2}{c}{Fine-tuning MAE} & & \\
\cmidrule(lr){2-4}\cmidrule(lr){5-6}
Method & Marin & MLF & AI2 & Marin & HF & PT+FT $S\uparrow$ & RL $S\uparrow$ \\
\midrule
\multicolumn{8}{@{}l}{\emph{Zero-shot, curve only}}\\
Last value & 0.533 & 0.495 & 0.575 & 0.097 & 0.108 & 0.000 {\tiny [0.00,0.00]} & 0.000 {\tiny [0.00,0.00]} \\
Power-law fit & 0.349 & 0.258 & 0.407 & 0.042 & 0.084 & 0.437 {\tiny [0.38,0.49]} & 0.037 {\tiny [-0.09,0.14]} \\
AIC ensemble of fits & 0.377 & 0.258 & 0.475 & 0.042 & 0.095 & 0.397 {\tiny [0.34,0.45]} & 0.047 {\tiny [-0.10,0.17]} \\
LC-PFN & 0.252 & 0.185 & 0.480 & 0.037 & 0.074 & 0.621 {\tiny [0.56,0.67]} & 0.133 {\tiny [0.01,0.24]} \\
FT-PFN (ifBO) & 0.368 & 0.416 & 0.483 & 0.071 & 0.100 & 0.286 {\tiny [0.25,0.32]} & -0.005 {\tiny [-0.10,0.08]} \\
Chronos-2 & 0.595 & 0.624 & 0.844 & 0.050 & 0.111 & 0.172 {\tiny [0.09,0.25]} & 0.075 {\tiny [-0.09,0.21]} \\
TimesFM-2.5 & 0.839 & 0.682 & 1.120 & 0.138 & 0.136 & -0.457 {\tiny [-0.51,-0.41]} & -0.385 {\tiny [-0.62,-0.19]} \\
\midrule
\multicolumn{8}{@{}l}{\emph{Trained on past runs, test sweep held out}}\\
\reader{} (2B, ours) & 0.123 & 0.247 & 0.275 & 0.031 & \underline{0.056} & 0.675 {\tiny [0.64,0.70]} & -0.311 {\tiny [-0.53,-0.13]}$^{\dagger}$ \\
NCPL (retrained) & 0.216 & 0.384 & 0.513 & 0.061 & 0.191 & 0.279 {\tiny [0.20,0.35]} & -0.053 {\tiny [-0.21,0.07]} \\
NCPL (retrained, source held out) & 0.303 & 0.328 & 0.435 & 0.032 & 0.094 & 0.460 {\tiny [0.41,0.51]} & \na \\
NeuNeu-NoLoss (our training) & 0.393 & 0.330 & 0.510 & 0.090 & 0.094 & 0.275 {\tiny [0.24,0.31]} & \na \\
NCPL (released checkpoint)$^{\S}$ & \na & 0.219$^{16}$ & \na & \na & \na & \na & \na \\
kNN on prefix shape & 0.165 & 0.205 & 0.195 & 0.214 & 0.108 & 0.570 {\tiny [0.51,0.63]} & -0.062 {\tiny [-0.26,0.10]} \\
Mean training shape & 0.183 & 0.237 & 0.197 & 0.214 & 0.110 & 0.498 {\tiny [0.42,0.56]} & 0.064 {\tiny [-0.12,0.22]} \\
\midrule
\multicolumn{8}{@{}l}{\emph{Trained on past runs, test sweep seen}}\\
NCPL (retrained) & 0.192 & \underline{0.152} & 0.234 & 0.005 & 0.082 & 0.792 {\tiny [0.77,0.82]} & \na \\
kNN on prefix shape & \underline{0.105} & 0.172 & \underline{0.193} & \underline{0.004} & 0.110 & \underline{0.847} {\tiny [0.82,0.87]} & -0.062 {\tiny [-0.26,0.10]} \\
Mean training shape & 0.182 & 0.231 & 0.195 & 0.009 & 0.113 & 0.734 {\tiny [0.69,0.77]} & 0.062 {\tiny [-0.12,0.21]} \\
\bottomrule
\end{tabular}

\vspace{2pt}
{\scriptsize\raggedright $^{\dagger}$The 96-run part of the RL set only; other RL entries use the whole set. NeuNeu-NoLoss is trained on 89 runs of an external source and never on ours (Appendix~\ref{app:baselines}). \na{}: not evaluated. $^{\S}$The released NCPL checkpoint sees the configuration only and needs eight numeric fields that 16 of the 300 test runs supply (all MLFoundations pretraining); its one MAE is over those 16 runs and is not comparable with its column (same-run comparison in the supplementary document, \S S2).\par}
\end{table}

\subsection{Main table, \textsc{Grid20}}
\label{app:main-grid20}
On the coarser \textsc{Grid20}, over the same runs and with the same scorer, the block order and the \textsc{Curve} and \textsc{Full} comparisons of \S\ref{sec:main} are unchanged, and the frontier models move by a few hundredths; \gflash{} ranks higher (error ratio 0.305 against 0.396) because its errors concentrate in the near half of the curve, which \textsc{Grid20} does not query.
The two \textsc{Grid20} tables will be part of the release; RL entries do not depend on the grid.

\subsection{Common valid answers}
\label{app:common-valid}
Each model in Table~\ref{tab:main} is scored on its own valid answers.
Table~\ref{tab:common-valid} repeats the comparison on the 282 views that every model of the core group answered validly and gives, for every model, the paired difference to LC-PFN on the views both answered; no set of views is answered validly by every model, because the small open-weight models answer too few in the \textsc{Full} condition.
On the common views the ordering of Table~\ref{tab:main} is unchanged, and only \opus{} with the curve alone is ahead of LC-PFN with an interval that excludes zero ($\Delta S=+0.08$).
Counting every invalid, missing, or absurd answer (Appendix~\ref{app:validity}) as the last value ($R=1$ for that view) changes the scores of \opus, \sol, \luna, \gpro, and Claude Haiku 4.5 by at most 0.01 and lowers the models with many invalid answers, such as \gflash{} (\textsc{Curve} from 0.60 to 0.48, \textsc{Full} from 0.60 to 0.56).

\begin{table}[h]
\centering\scriptsize
\caption{Skill score on common valid answers, pretraining and fine-tuning, \textsc{Grid10}. Top: the 282 views that every model in the core group (the rows of the top block) answered validly. Bottom: each model on its own valid views, as in Table~\ref{tab:main}. $\Delta S$ is the paired difference to LC-PFN on the same views (positive: better than LC-PFN), with a 95\% interval from a lineage-cluster bootstrap that resamples both methods together. Every entry excludes absurd outputs, as in Table~\ref{tab:main}. Kimi K3 \textsc{Curve} (292 views, $S=0.658$) and the \textsc{Curve+Config} rows of other models have no paired contrast in the final readout and appear only in Table~\ref{tab:main}.}
\label{tab:common-valid}
\setlength{\tabcolsep}{4pt}
\begin{tabular}{llrrr}
\toprule
Views & Method & $n$ & $S$ & $\Delta S$ vs.\ LC-PFN \\
\midrule
\multicolumn{5}{l}{\emph{Core group, common views}}\\
& \opus $\cdot$ \textsc{Curve} & 282 & 0.709 & +0.083 [+0.04, +0.13]$^{*}$ \\
& \reader{} (ours) & 282 & 0.677 & +0.051 [-0.00, +0.11] \\
& \sol $\cdot$ \textsc{Full} & 282 & 0.661 & +0.036 [-0.01, +0.08] \\
& \opus $\cdot$ \textsc{Full} & 282 & 0.646 & +0.021 [-0.04, +0.08] \\
& \luna $\cdot$ \textsc{Full} & 282 & 0.638 & +0.012 [-0.03, +0.07] \\
& \sol $\cdot$ \textsc{Curve+Config} & 282 & 0.626 & +0.001 [-0.04, +0.04] \\
& LC-PFN & 282 & 0.626 & -- \\
& \sol $\cdot$ \textsc{Curve} & 282 & 0.611 & -0.015 [-0.06, +0.03] \\
& \luna $\cdot$ \textsc{Curve+Config} & 282 & 0.595 & -0.031 [-0.09, +0.02] \\
& \luna $\cdot$ \textsc{Curve} & 282 & 0.573 & -0.053 [-0.11, +0.00] \\
\midrule
\multicolumn{5}{l}{\emph{Each model, own valid views}}\\
& kNN on prefix shape (sweep seen) & 300 & 0.847 & +0.226 [+0.17, +0.29]$^{*}$ \\
& NCPL, retrained (sweep seen) & 300 & 0.792 & +0.172 [+0.12, +0.22]$^{*}$ \\
& Mean training shape (sweep seen) & 300 & 0.734 & +0.113 [+0.05, +0.18]$^{*}$ \\
& \opus $\cdot$ \textsc{Curve} & 293 & 0.706 & +0.083 [+0.04, +0.13]$^{*}$ \\
& \reader{} (ours) & 299 & 0.675 & +0.052 [+0.00, +0.11]$^{*}$ \\
& \sol $\cdot$ \textsc{Full} & 297 & 0.657 & +0.037 [-0.01, +0.08] \\
& \opus $\cdot$ \textsc{Full} & 294 & 0.648 & +0.026 [-0.04, +0.09] \\
& \luna $\cdot$ \textsc{Full} & 298 & 0.632 & +0.011 [-0.03, +0.06] \\
& \sol $\cdot$ \textsc{Curve+Config} & 300 & 0.623 & +0.003 [-0.04, +0.05] \\
& LC-PFN & 300 & 0.621 & -- \\
& \sol $\cdot$ \textsc{Curve} & 300 & 0.612 & -0.008 [-0.06, +0.04] \\
& \gpro $\cdot$ \textsc{Full} & 300 & 0.610 & -0.011 [-0.07, +0.05] \\
& \kimi{} $\cdot$ \textsc{Full} & 274 & 0.608 & -0.015 [-0.07, +0.04] \\
& Claude Haiku 4.5 $\cdot$ \textsc{Full} & 297 & 0.608 & -0.014 [-0.08, +0.04] \\
& \gpro $\cdot$ \textsc{Curve} & 298 & 0.606 & -0.015 [-0.07, +0.04] \\
& \gflash $\cdot$ \textsc{Full} & 265 & 0.604 & -0.004 [-0.07, +0.06] \\
& \gflash $\cdot$ \textsc{Curve} & 217 & 0.598 & -0.045 [-0.10, +0.01] \\
& \luna $\cdot$ \textsc{Curve+Config} & 300 & 0.589 & -0.031 [-0.08, +0.02] \\
& Muse-Glimmer-30B $\cdot$ \textsc{Curve} & 298 & 0.580 & -0.045 [-0.11, +0.02] \\
& \luna $\cdot$ \textsc{Curve} & 300 & 0.574 & -0.047 [-0.10, +0.01] \\
& kNN on prefix shape (sweep held out) & 300 & 0.570 & -0.051 [-0.14, +0.03] \\
& Qwen3-Next-80B $\cdot$ \textsc{Full} & 300 & 0.553 & -0.067 [-0.12, -0.01]$^{*}$ \\
& Qwen3-Next-80B $\cdot$ \textsc{Curve} & 300 & 0.540 & -0.080 [-0.13, -0.03]$^{*}$ \\
& Mean training shape (sweep held out) & 300 & 0.498 & -0.122 [-0.22, -0.03]$^{*}$ \\
& Qwen3.5-2B $\cdot$ \textsc{Curve} & 273 & 0.480 & -0.141 [-0.21, -0.06]$^{*}$ \\
& Claude Haiku 4.5 $\cdot$ \textsc{Curve} & 297 & 0.463 & -0.159 [-0.23, -0.09]$^{*}$ \\
& NCPL, retrained (source held out) & 300 & 0.460 & -0.160 [-0.23, -0.10]$^{*}$ \\
& Qwen3.5-2B $\cdot$ \textsc{Curve+Config} & 289 & 0.454 & -0.172 [-0.26, -0.08]$^{*}$ \\
& Muse-Glimmer-30B $\cdot$ \textsc{Full} & 298 & 0.446 & -0.175 [-0.26, -0.09]$^{*}$ \\
& Power-law fit & 300 & 0.437 & -0.183 [-0.24, -0.11]$^{*}$ \\
& Qwen3.5-9B $\cdot$ \textsc{Full} & 299 & 0.415 & -0.205 [-0.28, -0.13]$^{*}$ \\
& AIC ensemble of fits & 300 & 0.397 & -0.224 [-0.29, -0.15]$^{*}$ \\
& Qwen3.5-9B $\cdot$ \textsc{Curve} & 300 & 0.379 & -0.241 [-0.33, -0.15]$^{*}$ \\
& Qwen3.5-2B $\cdot$ \textsc{Full} & 59 & 0.306 & -0.200 [-0.42, -0.05]$^{*}$ \\
& FT-PFN & 300 & 0.286 & -0.334 [-0.37, -0.29]$^{*}$ \\
& NCPL, retrained (sweep held out) & 300 & 0.279 & -0.342 [-0.44, -0.25]$^{*}$ \\
& NeuNeu-NoLoss & 300 & 0.275 & -0.346 [-0.39, -0.29]$^{*}$ \\
& Chronos-2 & 300 & 0.172 & -0.449 [-0.52, -0.38]$^{*}$ \\
& Qwen3.5-4B $\cdot$ \textsc{Curve} & 98 & 0.136 & -0.410 [-0.54, -0.28]$^{*}$ \\
& DeepSeek-V3.2 $\cdot$ \textsc{Curve} & 286 & -0.035 & -0.667 [-0.83, -0.53]$^{*}$ \\
& Qwen3.5-0.8B $\cdot$ \textsc{Curve} & 289 & -0.371 & -0.991 [-1.11, -0.88]$^{*}$ \\
& TimesFM-2.5 & 300 & -0.457 & -1.078 [-1.15, -1.00]$^{*}$ \\
& DeepSeek-V3.2 $\cdot$ \textsc{Full} & 290 & -0.467 & -1.098 [-1.28, -0.92]$^{*}$ \\
\bottomrule
\end{tabular}

\vspace{2pt}{\scriptsize $^{*}$ interval excludes zero.}
\end{table}

\subsection{Where the LLM beats the curve fitter}
\label{app:routing}
Table~\ref{tab:routing} asks on which runs the strongest LLMs gain over LC-PFN.
The 294--297 \textsc{Grid10} pretraining and fine-tuning views that both answered are split into quartiles of LC-PFN's error ratio $R$ (its error over that of the last value), and within each quartile the table gives the geometric-mean $R$ of both methods and the share of views on which the LLM is closer (per-view error ratios from the frozen scorer; geometric means over views, not over source cells as in Table~\ref{tab:main}).
The gain is concentrated where the fitter fails: on the quartile where LC-PFN already cuts the last-value error by 70--98\%, the LLMs are worse (1.35--1.71 times its error, closer on 38--43\% of views), and on the other three quartiles they cut its error by 24--47\% and are closer on 66--84\%.
Routing each view to whichever method errs less would raise the run-weighted skill from $0.69$--$0.71$ (LLM alone) to $0.74$--$0.79$, but the fitter's error is not observable before the answer, so a routing rule needs a signal computable from the prefix, which we leave to future work.

\begin{table}[t]
\centering\small
\caption{LLM against LC-PFN by quartile of LC-PFN's error ratio $R$ (\textsc{Grid10}, \textsc{Full} prompt, pretraining and fine-tuning views both methods answered). Geometric means over views; ``closer'' is the share of views on which the LLM's error is smaller.}
\label{tab:routing}
\setlength{\tabcolsep}{4pt}
\begin{tabular}{lrccc@{\hspace{14pt}}ccc}
\toprule
& & \multicolumn{3}{c}{\opus{} ($n{=}294$)} & \multicolumn{3}{c}{\sol{} ($n{=}297$)} \\
\cmidrule(lr){3-5}\cmidrule(lr){6-8}
LC-PFN $R$ quartile & LC-PFN $R$ & LLM $R$ & LLM / LC-PFN & closer & LLM $R$ & LLM / LC-PFN & closer \\
\midrule
Q1 (0.02--0.30) & 0.15 & 0.26 & 1.71 & 38\% & 0.21 & 1.35 & 43\% \\
Q2 (0.30--0.39) & 0.34 & 0.22 & 0.64 & 75\% & 0.26 & 0.76 & 66\% \\
Q3 (0.39--0.59) & 0.47 & 0.25 & 0.53 & 78\% & 0.32 & 0.68 & 84\% \\
Q4 (0.59--2.07) & 0.84 & 0.50 & 0.59 & 68\% & 0.56 & 0.66 & 81\% \\
\bottomrule
\end{tabular}
\end{table}

\subsection{The released NCPL checkpoint}
\label{app:ncpl-released}
The released NCPL checkpoint needs eight numeric fields without a missing-value mask, which only 16 of the 300 test runs supply, all MLFoundations pretraining runs; the supplementary document (\S S2) therefore scores every method on those 16 runs.
The checkpoint sees the configuration only; it beats the last value, but its error ratio is about twice that of LC-PFN (0.48 against 0.23).

\subsection{Reinforcement learning}
\label{app:rl}
The RL set holds 150 runs from more than 20 public projects, selected by a blind seeded rule (completion of at least 98\% of the recovered budget, at most 25\% of the set per project, at most two runs per group of identical configurations); its two parts of 96 and 54 runs share no run key and no project and differ in the form of the released prefix (Appendix~\ref{app:baselines}).
RL scores exclude the 17 runs with a constant future (Appendix~\ref{app:metrics}; Table~\ref{tab:rl-constant}); the reader is scored on the 96-run part, 82 of whose runs remain.
The supplementary document (\S S7) tests the other scalar curves that RL runs log as forecast inputs.
For 17 RL runs planned in epochs (not the same set as the 17 runs with a constant future), the planned step budget, which also sets the $x$ axis, is the number of epochs times the steps per epoch of the complete log (a ratio that is constant over the whole log in 9 of them) and is shown to the forecaster as configuration.
Reward units differ by project, so the supplementary document (\S S3) gives the error per project in its own units beside the spread of the reward inside the observed prefix; its rows should not be pooled.

\begin{table}[h]
\centering\scriptsize
\caption{The 17 RL runs whose logged series (reward in 14, training loss in 3 open-r1 runs) never leaves its last observed value, where the error ratio does not exist. For four LLMs and the reader, per condition: how many of these runs it answered, and on how many it forecast the constant exactly (zero error). Per-source errors in the series' own units are in the analysis files of the release. $^{*}$The reader is scored on the 96-run part of the RL set, which holds 14 of the 17.}
\label{tab:rl-constant}
\begin{tabular}{llrr}
\toprule
Model & Condition & Runs answered & Exactly constant \\
\midrule
Claude Haiku 4.5 & \textsc{Curve} & 17 & 12 \\
Claude Haiku 4.5 & \textsc{Full} & 17 & 6 \\
\luna & \textsc{Curve} & 17 & 15 \\
\luna & \textsc{Full} & 17 & 9 \\
\opus & \textsc{Curve} & 17 & 4 \\
\opus & \textsc{Full} & 17 & 10 \\
\reader{} (ours) & \textsc{Curve+Config} & 12$^{*}$ & 9 \\
\sol & \textsc{Curve} & 17 & 16 \\
\sol & \textsc{Full} & 17 & 16 \\
\bottomrule
\end{tabular}
\end{table}

The two parts disagree on the methods without an LLM: on the 96 runs the AIC ensemble reaches an error ratio of 0.802 $[0.669,0.971]$ and LC-PFN 0.806 $[0.684,0.953]$, while the pool-trained shape regressors do not beat the last value (mean training shape 1.093 $[0.859,1.405]$, nearest-neighbor regressor 1.247 $[1.021,1.549]$, source held out); on the other 54 the shape regressors reach 0.74 and 0.80 and the AIC ensemble 1.262.
Over all 133 scored runs, an exponential fit, which is not among the methods of Table~\ref{tab:main-b}, reaches $S=0.14$ $[0.05,0.23]$, against $0.13$ $[0.01,0.24]$ for LC-PFN.
\gpro{}'s low RL score comes from forecasts biased toward continued improvement: its median normalized error grows from $+0.06$ $[0.02,0.11]$ at 30\% of the budget to $+0.64$ $[0.34,0.84]$ at the end, against at most $+0.03$ for \opus{} and \sol{}; its median path leaves the range the run ever occupied in 49\% $[43,55]$ of views (\sol{} 22\%), and its 80\% intervals cover 0.69 $[0.64,0.74]$ of targets (\sol{} 0.85).
This optimism helps only where the reward keeps rising: on the 30 PPO runs of the 54-run part, where the last value falls short by $-1.04$ $[-1.39,-0.87]$ at the end, \gpro{} is the only model not significantly short ($+0.25$ $[-0.09,0.64]$).
For curve fitters a longer prefix helps on RL (LC-PFN from an error ratio of 0.83 at 20\% to 0.59 at 50\% on 90 questions), and discarding the earliest segment, as \citet{khatri2025scalerl} recommend, helps the parametric ensemble (1.37 to 0.80); neither helps the LLMs (\sol{} 0.68, 0.90, 0.67, 0.69 at 20--50\%, and 0.73 without the early segment).

\subsection{Community runs in the wild}
\label{app:wild-details}
Appendix~\ref{app:community-diag} examines candidate causes of the drop on community runs, and Appendix~\ref{app:axis} breaks the forecasting error down by data axis.

\label{app:det-wild}
The in-the-wild set holds 232 community runs by 184 authors, sampled from training-state files on the Hugging Face Hub (\S\ref{sec:data}).
Figure~\ref{fig:main} scores the \emph{core-211 cohort}: the 211 runs on which every model first evaluated on this set (\sol{}, \luna{}, and \opus{}, each with the curve alone and with the full context, and the 2B \reader{}) returned a valid \textsc{Grid10} forecast; the other 21 runs lack a valid answer from at least one of them (\opus{} answered 222 of the 232 runs with the full context).
The cohort was fixed before the other methods were run, and each method is scored on its own valid answers within it (199--211 runs per method).
Scoring all 232 runs instead changes skill by less than 0.022 for every scored method (the largest change is $+0.0216$, for the ExtraTrees regressor of Appendix~\ref{app:baselines}); on all 232 runs \opus{} and \sol{} score $0.31$ and LC-PFN $0.09$.
By stage, the drop in skill concentrates in the 92 RL runs: over all 232 runs and both grids, the full-context error ratio of \sol{}, \luna{}, and \gflash{} is 0.96--1.04 on RL runs against 0.55--0.64 on the 140 fine-tuning and preference-tuning runs, and the hindsight-best smooth fit reaches an error ratio of 0.31 on the RL runs against 0.18 on the others.
To test whether documentation explains the loss of skill, we edited the context of each run and forecast it again, paired within the run and query grid (both grids pooled for \sol{} and \luna{}, \textsc{Grid10} for \opus{}).
Deleting every identifying field (base model, datasets, reward functions, algorithm, framework) or adding everything recoverable offline from the model cards moves the error ratio by at most 4\% for \sol{}, \luna{}, and \opus{} (every interval covers 1); removing the context altogether gives point ratios of 1.16 and 1.14 for \sol{} and \luna{} (owner-cluster intervals $[0.98,1.51]$ and $[1.00,1.42]$, respectively; both include one before rounding), and 0.99 for \opus{}.
On \textsc{Grid10} alone the first two ratios are 1.26 $[0.97,1.98]$ and 1.27 $[1.01,1.86]$; the metadata-deletion and enrichment intervals still include one.
Runs by individuals are forecast worse than runs by institutions (error ratios 0.81--1.02 against 0.44--0.64 for \sol{}, \luna{}, and \gflash{} in both conditions), and the gap is already present with the curve alone (1.6--1.8$\times$); dividing each forecast by the error of the best smooth fit in hindsight for that run removes it (0.87--1.15, every interval covering one).

The logging explains the rest.
The hindsight-best fit reaches an error ratio of 0.23 on community runs against 0.09 on the frozen sets.
A prefix window rests on a median of 7 raw points in the wild against 59 in the frozen sets; thinning each frozen curve to the wild density raises its hindsight floor from 0.090 to 0.132, and adding wild-level residual noise raises it to 0.214 against the 0.225 observed, which reproduces 95\% of the gap on a log scale (92\% on a linear one).
Regressing the floor on these curve properties shrinks the author term from $+0.49$ ($t{=}2.9$) to $+0.14$ ($t{=}1.1$).
Repository creation date has no detectable association with error within stage ($|t|\le1.9$; the test detects changes of 27--52\% per year). Creation date approximates training time and the knowledge cutoffs of \sol{} and \luna{} are not documented, so the analysis measures recency, not exposure~before~a~cutoff.

The peak learning rate, schedule shape, and warmup fraction shown for the 232 community runs were computed from the complete learning-rate log; truncated at 20\%, the log gives a different schedule label for 118 runs and different values of the other two fields for 2.
The community-run column of Figure~\ref{fig:main} and the analyses of this section use these fields; pretraining and fine-tuning configurations are recorded values, and every shown prefix ends at 20\% of the budget.

\subsection{Trained predictors under three regimes}
\label{app:regimes-main}
\label{app:regimes}
A predictor trained on our pool is evaluated in one of three \emph{held-out regimes}: with other runs of the test run's own sweep in the pool (\emph{sweep seen}), with the test run's sweep family removed (\emph{sweep held out}, the default), or with its whole source removed (\emph{source held out}).
For sources that name their runs, the sweep family is the run name with the trailing hash stripped and every numeric token replaced by a placeholder, so that seed, learning rate, width, budget, and date collapse into one family (\texttt{muonh-may-recipe-lr-v1-d768-R120-lr0p4} becomes \texttt{muonh-may-recipe-lr-v\#-d\#-R\#-lr\#p\#}); for sources that expose only tracker identifiers, it is a signature of the configuration without the keys a sweep varies (learning rate, seed, weight decay, warmup, budget) and without infrastructure keys.
A source is one tracker entity, so pretraining and fine-tuning runs of one organization are different sources.
When a regime empties the pool, the method falls back to the last value and is scored normally.
Table~\ref{tab:regimes} gives the per-cell MAE.

\begin{table}[t]
\centering\small
\caption{Pool-trained baselines under the three regimes, \textsc{Grid10} MAE per cell (lower is better). The zero-shot references at the bottom never use the pool and are identical across regimes. Bold: better of the two shape regressors in a column within a regime.}
\label{tab:regimes}
\setlength{\tabcolsep}{4pt}
\begin{tabular}{llccccc}
\toprule
Method & Regime & PT/Marin & PT/MLF & PT/AI2 & FT/Marin & FT/HF \\
\midrule
Mean training shape & sweep seen & 0.182 & 0.231 & 0.195 & 0.009 & 0.113 \\
kNN on prefix shape & sweep seen & \textbf{0.105} & \textbf{0.172} & \textbf{0.193} & \textbf{0.004} & \textbf{0.110} \\
\midrule
Mean training shape & sweep held out & 0.183 & 0.237 & 0.197 & \textbf{0.214} & 0.110 \\
kNN on prefix shape & sweep held out & \textbf{0.165} & \textbf{0.205} & \textbf{0.195} & \textbf{0.214} & \textbf{0.108} \\
\midrule
Mean training shape & source held out & \textbf{0.188} & 0.247 & \textbf{0.208} & \textbf{0.081} & \textbf{0.054} \\
kNN on prefix shape & source held out & 0.354 & \textbf{0.216} & 0.411 & 0.100 & 0.097 \\
\midrule
NCPL (in-distribution only) & sweep seen & 0.192 & 0.152 & 0.234 & 0.005 & 0.082 \\
LC-PFN & zero-shot & 0.252 & 0.185 & 0.480 & 0.037 & 0.074 \\
\opus{} $\cdot$ \textsc{Full} & zero-shot & 0.150 & 0.213 & 0.216 & 0.022 & 0.119 \\
Last value & -- & 0.533 & 0.495 & 0.575 & 0.097 & 0.108 \\
\bottomrule
\end{tabular}
\end{table}

With the sweep seen, the nearest-neighbor regressor reaches $S=0.85$ and the mean training shape 0.73 (0.98 and 0.91 on the fine-tuning source whose 100 test runs come from one sweep), which is within-sweep interpolation: most of the pool shares a sweep family with a test run (82\% of the NCPL training and validation split; Appendix~\ref{app:data}).
Holding out the sweep family lowers them to 0.57 and 0.50, removing a coarser grouping of the Marin sweeps, a variant not in the table, to 0.55 and 0.50, and holding out the source gives 0.38 and 0.58, all below LC-PFN (0.62); retrained NCPL falls from 0.79 with the test run's own sweep seen to 0.28 with its sweep family held out and 0.46 with the source held out.
On Marin pretraining the better shape regressor reaches an MAE of 0.105 $[0.076,0.141]$ with the sweep seen, 0.165 $[0.112,0.227]$ with it held out, and 0.188 $[0.149,0.230]$ with the source held out, against 0.150 $[0.106,0.200]$ for the best zero-shot LLM.
Three limits qualify the table.
The strict pool contains no runs from the MLFoundations, AI2, or Hugging Face sources, so for those cells NCPL's sweep-held-out and source-held-out models were trained on the same rows and differ only by nondeterminism; with the two models' errors averaged in those three cells, the overall skill is 0.37 with the sweep held out and 0.39 with the source held out, so most of the gap between 0.28 and 0.46 is training nondeterminism and not the regime.
The Marin fine-tuning test runs come from one sweep that holds 69 of the 71 pool runs for that cell, so with the sweep held out both shape regressors are fitted on the remaining two runs (0.214, against 0.097 for the last value).
The source-held-out pool is 93--98\% Marin pretraining for the non-Marin cells, and the best method is chosen per column after the fact, which flatters the trained block.

\FloatBarrier

\section{Information conditions and setting--curve matching}
\label{app:sec4-more}
This appendix gives the analyses of the semantics referenced from \S\ref{sec:main}: the three information conditions, what in the semantics carries the gain, and the setting--curve matching task.

\subsection{Information conditions}
\label{app:conditions}
Appendix~\ref{app:factorial} adds the description and the code separately with the curve and the configuration fixed, and Appendix~\ref{app:decisions} tests controlled changes on newly trained runs.

\label{app:ladder}
\label{app:schedule}
Every API model was asked in three conditions: the curve alone (\textsc{Curve}), the curve with the numeric configuration (\textsc{Curve+Config}), and the curve with the configuration and the semantics (\textsc{Full}); per-model scores are in Table~\ref{tab:main} and paired contrasts in Tables~\ref{tab:significance} and~\ref{tab:significance-b}. Kimi K3 has no \textsc{Curve+Config} responses, so contrasts involving that condition are unavailable for it.
On the \textsc{Grid10} pretraining and fine-tuning runs, adding the configuration moves \opus{}, \sol{}, \luna{}, \gpro{}, and Qwen3-Next-80B by less than 0.02 (paired intervals include zero) and raises Claude Haiku 4.5 by $0.08$ $[0.03,0.14]$ and \gflash{} by $0.04$ $[0.00,0.08]$; adding the semantics on top helps \sol{} ($+0.03$ $[0.00,0.06]$), \luna{} ($+0.04$ $[0.00,0.09]$), and Claude Haiku 4.5 ($+0.06$ $[0.02,0.11]$), and costs \opus{} 0.06 $[0.02,0.09]$, \gflash{} 0.08 $[0.03,0.13]$, Muse-Glimmer-30B 0.09 $[0.03,0.16]$, and DeepSeek-V3.2 0.30 $[0.13,0.47]$.

The gain concentrates where the prefix misleads.
We derived schedule descriptors from each run's configuration and source facts and split runs by whether warmup covers at least half of the observed prefix. A second, LLM-derived labeling agreed with the descriptors on 30 sampled cases, as recorded in its saved summary; the transcript of that check was not kept, and the descriptors and the sample are reproducible. Framework-default schedule shapes are assumed.
On pretraining, several curve-only methods have higher error ratios in the warmup-dominated group than in the gradual-anneal group (LC-PFN 0.443 against 0.321, \sol{} 0.461 against 0.333, \gpro{} 0.606 against 0.331).
There, the paired ratio of \textsc{Full} to \textsc{Curve} error is 0.64 $[0.52,0.78]$ for \opus, 0.71 $[0.61,0.82]$ for \gpro, and 0.72 $[0.59,0.88]$ for \luna, against 1.32, 1.11, and 0.94 elsewhere; \sol{} gains in both groups (0.91 and 0.82).
No pretraining or fine-tuning run reports the shape of its schedule and no RL run reports any schedule field, so the effect of a late cooldown cannot be tested, and because the groups are nearly collinear with source, these comparisons are within stage and associational.

The semantics act as a cue to the kind of run.
On 100 fixed questions, replacing the full context by a different run's context from the same source cell gives placebo-to-full MAE ratios of 0.98 $[0.83,1.16]$ for \sol{} (99 valid pairs) and 1.03 $[0.90,1.16]$ for \luna{} (98), both intervals including one. Seventy-five questions meet the predeclared large-configuration-change rule (learning rate, model size, budget, batch size, or warmup), and on this subset the placebo interval also includes one. For \luna{}, the deletion-only description arm (task description, source facts, known unknowns, and interpretation rule) has an error ratio against \textsc{Curve} of 0.800 $[0.684,0.939]$, compared with 0.759 $[0.668,0.856]$ for \textsc{Full}, recovering about four fifths of the log gain.

\subsection{Setting--curve matching}
\label{app:matching-task}
\label{app:matching}
\label{app:knob}
Each matching question takes two runs from the same sweep, shows both curves and both settings with run identities hidden and the reported parameter values retained, and asks which setting produced which curve.
We score 340 pairs: the 150 pretraining and 70 fine-tuning main pairs of a 330-pair matching release in which no curve endpoint appears in two pairs of one set (Appendix~\ref{app:data}), and 120 RL matching pairs from the RL release. The RL matching pairs use 240 distinct run endpoints; 40 of these also appear among that release's 96 forecasting runs. Forecasting-run counts and matching-endpoint counts therefore describe different task rosters.
Each pair is asked in both orders and in two conditions (\textsc{Curves+Settings} and \textsc{Full}). The RL release additionally presents each order at two curve-sampling densities (the \textsc{Grid10} and \textsc{Grid20} request variants); its scorer averages the valid views, up to four, into one probability per pair and condition. The pretraining and fine-tuning release averages up to two valid order views. Thus a stage's $n$ in Table~\ref{tab:matching} counts pair--condition rows (chance 0.5), and uncertainty is aggregated by pair, never by view.
Within each stage's scored matching set, no run appears in two pairs. Intervals resample recorded lineage components on pretraining and fine-tuning (140 components for the 150 pretraining pairs, 20 for the 70 fine-tuning pairs) and source projects on RL; two pairs from one sweep that fall in different components are resampled independently, so the pretraining and fine-tuning intervals may be too narrow.
Table~\ref{tab:matching} gives the accuracy by model and stage; five models, among them the open-weight Kimi K3, reach 0.82--0.86 on the unweighted mean over the three stages.
Three models are at chance in every stage: the open-weight DeepSeek-V3.2 and Qwen3-Next-80B, and the closed Claude Haiku 4.5.
Matching by forecasting predicts each curve under each setting and keeps the closer assignment. From a 20\% prefix, \sol{} scores 0.49 $[0.43,0.55]$ on 260 valid pairs and \opus{} 0.47 $[0.41,0.54]$ on 206; both intervals include chance. The planned set has 266 pairs (the 220 pretraining and fine-tuning pairs and the 46 RL pairs whose curves cover the forecast windows) and 532 distinct run endpoints; 19 recorded lineage components link more than one pair. On pretraining and fine-tuning specifically, swapping the setting moves \sol's forecast by 14\% of the curve gap with a prefix (215 valid pairs), versus 106\% without one (217); these are ratios of summed displacements to summed gaps. Across all stages, the corresponding values are 5\% and 57\%. The direction accuracy without a prefix is 0.68 $[0.62,0.74]$ on 262 valid pairs, with a pair-level interval. This comparison also changes the scoring horizon: no-prefix predictions are scored over the whole curve, whereas prefix predictions are scored on future nodes.

A sensitivity analysis keeps the 256 pairs on which \sol{} is valid in both conditions and scores both on the future nodes. Direction accuracy is 0.50 $[0.43,0.56]$ with the prefix and 0.64 $[0.56,0.70]$ without it; the paired difference is 0.14 $[0.04,0.23]$. These 2{,}000-draw intervals resample 163 groups formed from recorded endpoint-lineage components for pretraining/fine-tuning and source projects for RL, whose endpoint lineage is missing. The comparison is exploratory.
The settings easiest to match directly (model size, initial checkpoint, RL budget: 96--100\%) are the ones a forecast from a prefix does not recover, and the learning rate is hard both ways; accuracy is lowest when a single kind of setting separates the two runs and highest when four do.
On speedrun pairs that differ in how an optimizer or schedule is implemented every model is at chance from the settings alone, and a diff of the training scripts lifts \sol{} to 0.88 and \gpro{} to 0.85; varying the diff budget over 3k, 6k, and 8k characters on the same 50 pairs has no detectable effect.

\begin{table}[t]
\centering\small
\caption{Setting--curve matching accuracy per pair, both information conditions pooled (chance 0.5), averaged over the three stages and per stage, and the rate at which both orders of a pretraining pair are answered correctly (chance 0.25). $n$ counts pair--condition rows: 150 pretraining, 70 fine-tuning, and 120 RL pairs, each in two conditions. The supervised row uses different protocols: its pretraining entry is trained on our pool and evaluated leave-project-out (150 pairs; 0.68 $[0.61,0.75]$), its RL entry is trained on release pairs from other sources and evaluated leave-source-out (0.59 $[0.50,0.66]$); no supervised variant exceeded chance on fine-tuning, where 52 of the 70 pairs differ in the initial checkpoint.}
\label{tab:matching}
\label{tab:matching-stage}
\setlength{\tabcolsep}{4pt}
\begin{tabular}{lccccc}
\toprule
Model & Avg. & PT ($n{=}300$) & FT ($n{=}140$) & RL ($n{=}240$) & Both orders, PT \\
\midrule
\opus & \textbf{0.859} & 0.775 & \textbf{0.893} & \textbf{0.91} & 0.614 \\
\gpro & 0.853 & \textbf{0.838} & 0.850 & 0.87 & 0.639 \\
\sol & 0.847 & 0.808 & 0.871 & 0.86 & \textbf{0.647} \\
\kimi & 0.834 & 0.798 & 0.821 & 0.88 & 0.525 \\
\gflash & 0.816 & 0.723 & 0.879 & 0.85 & 0.646 \\
\luna & 0.686 & 0.567 & 0.671 & 0.82 & 0.279 \\
DeepSeek-V3.2 & 0.515 & 0.495 & 0.525 & 0.53 & 0.201 \\
Claude Haiku 4.5 & 0.500 & 0.487 & 0.493 & 0.52 & 0.185 \\
Qwen3-Next-80B & 0.498 & 0.497 & 0.464 & 0.53 & 0.181 \\
\midrule
Ridge on config differences & -- & 0.68 & -- & 0.59 & -- \\
\bottomrule
\end{tabular}
\end{table}

Recast as four-way multiple choice (supplementary document, \S S4), every model falls toward chance in both directions (maximum 0.339, six of sixteen intervals above 0.25; \opus{} passes the schema on only 285 of 434 forward questions), which suggests that the models compare two candidates rather than evaluate one.

\FloatBarrier

\section{Observed share, input window, and failure monitoring}
\label{app:extremes-details}
\label{app:det-extreme}
This appendix gives the tables and protocols behind \S\ref{sec:extreme}: the observed share of the run (Figure~\ref{fig:extremes}a), the granularity of input and output (Figure~\ref{fig:extremes}b), and failure monitoring (Figure~\ref{fig:extremes}c).

\subsection{Observed share of the run}
Table~\ref{tab:prefix} scores the same 100 pretraining and fine-tuning questions (20 per cell) at observed shares from 0 to 50\% of the budget, relative to the last value at the standard 20\%.
Within each stage--source cell, numerator and denominator use the same valid questions. The 95\% intervals in Figure~\ref{fig:extremes}a resample lineage components within cells (2{,}000 draws; 85 components in the complete 100-question set).
Without any curve, \sol{} (1.32) is better than the mean training curve of the cell (1.53), \opus{} (1.57) is not, and Muse-Glimmer-30B is far worse (4.41), whereas the released configuration-only NCPL checkpoint with the Marin residual offset scores 0.54 on its 14 coverable MLFoundations pretraining questions (ten lineage components): without an observation, the scale of the loss must come from similar runs.
The LLMs miss the level but not the order: on a separate set of 100 questions (20 per cell) that asks for the whole curve with no prefix, the final values \sol{} predicts order the 20 runs of each cell with Spearman correlation 0.61--0.82 (from 0.61 $[0.22,0.86]$ on Hugging Face fine-tuning to 0.82 $[0.47,0.96]$ on MLFoundations pretraining; intervals resample recorded lineage components within each cell, 86 for the 100 questions).
From 10\% of the budget \opus{} and \sol{} lead every curve-only method and the retrained NCPL, which first beats the 20\% last value at a 15\% share; from 20\% to 50\% they improve by a further factor of 2.5, part of which comes from the shorter horizon that remains to be forecast.

\begin{table}[t]
\centering\small
\caption{Observed share of the run: the same 100 questions at observed shares from 0 to 50\% of the budget; MAE relative to the last value at the standard 20\% (geometric mean of paired-available-case MAE ratios over five cells; lower is better; Figure~\ref{fig:extremes}a plots $1-R$). NCPL is the retrained checkpoint, reading ten points resampled from the observed prefix; Muse-Glimmer-30B answers 88 of the 100 questions at 0\%.}
\label{tab:prefix}
\setlength{\tabcolsep}{5pt}
\begin{tabular}{lccccccc}
\toprule
Method & 0 & 2\% & 5\% & 10\% & 15\% & 20\% & 50\% \\
\midrule
Last value & -- & 5.99 & 2.80 & 1.77 & 1.30 & 1.00 & 0.37 \\
LC-PFN & -- & -- & 2.13 & 0.82 & 0.69 & 0.52 & 0.14 \\
\sol{} $\cdot$ \textsc{Full} & 1.32 & 1.15 & 0.90 & 0.46 & 0.47 & 0.39 & 0.14 \\
\opus{} $\cdot$ \textsc{Full} & 1.57 & 1.10 & 0.67 & 0.49 & 0.42 & 0.38 & 0.15 \\
Muse-Glimmer-30B $\cdot$ \textsc{Full} & 4.41 & 2.12 & 0.87 & 0.82 & 0.60 & 0.60 & 0.23 \\
NCPL (retrained) & -- & -- & 1.79 & 1.16 & 0.84 & 0.65 & 0.49 \\
\bottomrule
\end{tabular}
\end{table}

\subsection{Failure monitoring}
We index 358 runs whose metrics show a failure (KL blow-up, reward collapse, unrecovered loss jump, or non-finite values). Of the 171 whose failure point $x_c$ lies in $[0.14, 1]$ of the budget with data in every window, we draw 100, stratified by failure type and stage in a seeded order, and match each, at $x_c$, with two healthy runs of the same source and metric kind that have data at the same window positions, chosen in a seeded order without regard to their outcome.
Two protocols use~these~300~runs.

\emph{Lead time} (Figure~\ref{fig:extremes}c, Table~\ref{tab:lead}).
Each run is shown a window 10\% of the budget wide, holding everything logged in it, that ends a lead $g$ of 2, 5, or 10\% of the budget before $x_c$; a healthy run is shown the same absolute window as its failed partner, and a monitor returns the probability that the run will fail.
Groups whose failure comes too early for the window drop out, leaving 300, 291, and 258 runs (100, 97, and 86 failures) at the three leads.
AUROC is computed over runs.
The LLMs read the window, including the logged auxiliary series, with the run's context and return a failure probability; LC-PFN and NCPL score a run by the largest adverse change their median forecast predicts over the next 10\% of the budget, relative to the observed level; and the numeric detectors are trained leave-one-source-out.

\emph{Online monitoring} (Table~\ref{tab:monitor}).
Each run is visited at up to four checkpoints (the first and last admissible points of a 10, 20, \dots, 90\% grid and two equally spaced between them; 1{,}101 checkpoints over the 300 runs, a failed run only before its failure point), and each visit discloses everything logged up to the checkpoint and nothing after it. Every monitor is scored on these checkpoints.
A monitor may raise one alarm per run, and its threshold is fitted on held-out sources to allow one false alarm per five healthy runs.
In this online protocol, at that false-alarm rate and excluding a KL-divergence threshold, which reads the labeling criterion of 38 of the 100 failures, no monitor catches half of the failures: the best is boosted trees on curve statistics, which catch 0.41 $[0.32,0.51]$ of them with a median lead time of 0.60 of the budget, save 0.20 $[0.15,0.26]$ of the budget per failed run (the span between the alarm and the failure point, zero for a missed failure), raise a false alarm on 0.205 $[0.15,0.26]$ of the healthy runs, and forgo 0.17 $[0.13,0.22]$ of the budget per healthy run; the LLMs with the \textsc{Full} context catch 0.22--0.36.
These recall figures belong to the online protocol only; the lead-time AUROCs of Table~\ref{tab:lead} rank runs at a fixed lead without a threshold, and there the best LLM and the boosted trees are above chance at every lead (at $g=10\%$, \sol{} 0.67 $[0.62,0.74]$ and the trees 0.65 $[0.57,0.72]$).
Stacking an LLM's failure probability onto the numeric features changes AUROC by 0.013 $[-0.011,0.037]$ for \sol{} and 0.018 $[-0.008,0.047]$ for \opus{}.

\begin{table}[t]
\centering\small
\caption{Failure detection by lead time (Figure~\ref{fig:extremes}c): AUROC over runs of a monitor shown a 10\%-wide window ending a lead $g$ before the failure point (chance 0.5; brackets are 95\% intervals). Numeric detectors are trained leave-one-source-out; the trees are the $g \le 0.10$ part of the extended curve in the supplementary~document~(\S S5).}
\label{tab:lead}
\setlength{\tabcolsep}{5pt}
\begin{tabular}{lccc}
\toprule
Monitor & $g=2\%$ & $g=5\%$ & $g=10\%$ \\
\midrule
Runs (failures) & 300 (100) & 291 (97) & 258 (86) \\
\midrule
Logistic, primary metric & 0.65 & 0.62 & 0.58 \\
Boosted trees on curve statistics & 0.72 [0.66, 0.78] & 0.66 [0.60, 0.73] & 0.65 [0.57, 0.72] \\
LC-PFN & 0.45 [0.37, 0.53] & 0.46 [0.39, 0.54] & 0.48 [0.41, 0.56] \\
NCPL (retrained) & 0.43 [0.36, 0.50] & 0.45 [0.37, 0.53] & 0.47 [0.39, 0.55] \\
\opus{} & 0.73 [0.68, 0.78] & 0.64 [0.58, 0.69] & 0.61 [0.55, 0.68] \\
\sol{} & 0.73 [0.68, 0.78] & 0.67 [0.62, 0.73] & 0.67 [0.62, 0.74] \\
\luna{} & 0.74 [0.69, 0.79] & 0.68 [0.63, 0.73] & 0.65 [0.59, 0.71] \\
Muse-Glimmer-30B & 0.68 [0.62, 0.75] & 0.61 [0.55, 0.66] & 0.57 [0.50, 0.64] \\
\bottomrule
\end{tabular}
\end{table}

\begin{table}[t]
\centering\small
\caption{Online failure monitoring on 100 failed and 200 healthy runs, each visited at up to four checkpoints between 10\% and 90\% of the budget (1{,}101 checkpoints), at one false alarm per five healthy runs. Lead time, compute saved (mean over failed runs of the budget between alarm and failure point), and healthy stopped (mean over healthy runs of the budget after a false alarm) are fractions of a run's budget; brackets are 95\% cluster-bootstrap intervals. A KL-divergence threshold (0.51 $[0.42,0.61]$) is omitted because 38 of the failures are labeled by a KL criterion.}
\label{tab:monitor}
\setlength{\tabcolsep}{5pt}
\begin{tabular}{lcccc}
\toprule
Monitor & \makecell{Failures\\caught} & \makecell{Median\\lead time} & \makecell{Compute\\saved} & \makecell{Healthy\\stopped} \\
\midrule
Loss spike $>k\sigma$ & 0.17 [0.10, 0.25] & 0.08 & 0.04 & 0.12 \\
Gradient-norm rule & 0.30 [0.21, 0.39] & 0.17 & 0.11 & 0.16 \\
Logistic, next 10\% & 0.35 [0.26, 0.44] & 0.10 & 0.09 & 0.08 \\
Boosted trees on curve statistics, rest of run & \textbf{0.41} [0.32, 0.51] & 0.60 & 0.20 & 0.17 \\
\luna{} $\cdot$ \textsc{Full} & 0.22 [0.14, 0.30] & 0.06 & 0.04 & 0.16 \\
\opus{} $\cdot$ \textsc{Full} & 0.29 [0.21, 0.38] & 0.08 & 0.07 & 0.16 \\
\sol{} $\cdot$ \textsc{Full} & 0.36 [0.27, 0.46] & 0.09 & 0.07 & 0.15 \\
\bottomrule
\end{tabular}
\end{table}

Extended for the best numeric detector from $g=0$ to $g=0.30$ (supplementary document, \S S5), the lead-time protocol shows a ceiling: the pooled AUROC stays flat beyond $g=0.15$, KL blow-ups fall to chance beyond a 10\% lead (0.58, 0.43, 0.37 at $g=0.10$, 0.15, 0.20), and reward collapses decay from 0.81 to 0.53.
For an LLM monitor, recency matters more than coverage: for \sol{} and \luna{}, the last 10\% before a 2\% lead gives 0.73--0.74, the first and last 5\% of the run 0.69, and the first 10\% or the configuration alone 0.53--0.56.
Across 90 coded alarms, false alarms cite the configuration in 77\% of cases against 50\% for correct early alarms, which instead cite the shape of the curve and the gradient norm.

\subsection{Granularity of input and output}
\label{app:granularity}
We re-issued 95 questions of the frozen panel at input windows of 2, 1, 0.5, and 0.2\% of the budget and output grids of 10, 5, and 2\%, always scoring the common points $x\in\{0.3,\dots,1.0\}$ against unchanged targets; the 86 questions whose raw logs are dense enough for a 0.2\% window form the fixed set of Table~\ref{tab:granularity} and Figure~\ref{fig:extremes}b.
A finer input window leaves \sol{}'s per-question error unchanged (paired ratio $1.00$ $[0.84,1.20]$ between the 2\% and 0.2\% windows); the table's $R$ moves from 0.335 to 0.300 because it averages questions within each cell and then weights the five cells equally, whereas the paired ratio weights questions equally.
\opus{} stays between 0.29 and 0.35 at every window, whereas Muse-Glimmer-30B improves from 0.54 at the 2\% window to 0.34 at 0.2\%.
Among the curve methods only Chronos-2 gains (paired $0.72$ $[0.65,0.81]$), while TimesFM-2.5 ($1.73$ $[1.43,2.15]$) and the AIC ensemble ($1.45$ $[1.24,1.71]$) get worse; NCPL reads ten window means by construction, so its single entry is repeated across the windows of Figure~\ref{fig:extremes}b.
Question sets and averaging differ from Table~\ref{tab:prefix} (86 against 100 questions; per-question against per-cell ratios), so its entries at the 2\% window differ from the 20\% column there.
A finer output grid leaves LLM error at the common points unchanged (six paired contrasts between $0.88$ and $1.02$, every interval covering 1), but on a 40-point grid the schema-valid rate falls to $0.77$ for \sol{} and $0.47$ for \opus{}, every failure a truncation at the 4{,}096-token answer bound, while \luna{} stays at $0.99$.

\begin{table}[t]
\centering\small
\caption{Error ratio $R$ (lower is better) on the common prediction points as the input window shrinks (86 questions, the same in every column; Figure~\ref{fig:extremes}b plots $1-R$). Each method is scored on its valid answers (\opus{} 84--86, others 85--86). NCPL reads ten window means by construction and has one entry. Bold: lowest in the column.}
\label{tab:granularity}
\begin{tabular}{lcccc}
\toprule
& \multicolumn{4}{c}{input window} \\
\cmidrule(lr){2-5}
method & 2\% & 1\% & 0.5\% & 0.2\% \\
\midrule
last value      & 1.000 & 1.000 & 1.000 & 1.000 \\
AIC ensemble    & 0.595 & 0.852 & 0.951 & 0.892 \\
LC-PFN          & 0.352 & 0.411 & 0.436 & 0.409 \\
Chronos-2       & 1.223 & 1.020 & 1.077 & 0.900 \\
TimesFM-2.5     & 1.709 & 1.993 & 2.463 & 2.430 \\
\midrule
\sol{} (full)      & \textbf{0.335} & \textbf{0.249} & 0.296 & \textbf{0.300} \\
\sol{} (curve only)& 0.395 & 0.418 & 0.306 & 0.368 \\
\luna{} (full)     & 0.360 & 0.396 & 0.335 & 0.378 \\
\opus{} (full)     & 0.346 & 0.330 & \textbf{0.290} & 0.304 \\
Muse-Glimmer-30B (full) & 0.538 & 0.516 & 0.585 & 0.343 \\
NCPL (retrained)   & 0.604 & -- & -- & -- \\
\bottomrule
\end{tabular}
\end{table}

\FloatBarrier

\section{\reader{}: details}
\label{app:reader}
Appendix~\ref{app:temporal} tests the reader on later runs from new owners, and Appendix~\ref{app:clean} gives the reruns after removing the 22 related records described below.

\label{app:recipe}
This appendix gives the recipe, training data, training pools, untrained baselines, three-stage averages, and ablations of \reader{} (\S\ref{sec:training}).

\subsection{Recipe and training data}
\begin{itemize}\setlength{\itemsep}{0pt}
  \item \textbf{Backbone \& adapter.} Qwen3.5-2B; LoRA $r{=}64$ on the attention and MLP projections (q, k, v, o, gate, up, down), lr $2\times10^{-4}$.
  \item \textbf{Target.} Each query point as a \emph{ratio} to the last observed value, 4 decimals, as text.
  \item \textbf{Prompt.} The re-gridded curve prefix plus the numeric reported configuration; no description,
        no code, no retrieved runs.
  \item \textbf{Data.} 7{,}724 examples from 918 runs, filtered with the reference pool's own relation map (limits under a broader relation graph in Appendix~\ref{app:reader-pools});
        add three truncated views per example only if short observations are part of the target use.
  \item \textbf{Schedule.} 1{,}000 optimizer steps, effective batch 8, cosine, 3\% warmup, max length 8{,}192,
        loss on answer tokens only ($\approx$1.2 GPU-hours of training).
  \item \textbf{Decoding.} Greedy; three training seeds of the reference differ by up to 0.06 in skill $S$ on pretraining and 0.08 on fine-tuning runs of the test sets (\textsc{Grid10}).
\end{itemize}

The reference pool has 918 runs and 7{,}724 training examples: 578 runs from the strict pool of Appendix~\ref{app:data}, from which every test lineage component and test sweep family is removed (1{,}463 runs), and 340 public Hugging Face fine-tuning and RL runs admitted under the same rules; by stage it holds 218 pretraining, 52 synthetic-task, 280 fine-tuning or preference-tuning, and 368 RL runs.
Under the reference pool's own relation map, the pool contains no test run, no run from a non-Marin test project, no test sweep family, and no test lineage component; against the community set they share no repository and no lineage, but 107 of the 232 community runs share an author and 61 a model family with a training run.
The target representation decides whether training works: trained to emit raw metric values the reader is worse than LC-PFN, and writing each target as a ratio $y(x)/\bar y_{10}$ to the last observed value lowers the error on the Marin fine-tuning cell from 0.087 (raw) to 0.054 (difference) and 0.035 (ratio).
Quantiles written as text are too wide (coverage 0.89--0.93 at a nominal 0.80); a separate quantile head trained with pinball loss restores coverage to 0.79 at a 4--17\% cost~in~point~error.

\subsection{Training pools and backbones}
\label{app:reader-pools}
Table~\ref{tab:reader-macro} gives the reader under the recipe on four training pools and two backbones, reading the configuration alone (config) or the configuration and the semantics condensed into one paragraph (config + semantics), with three paired training seeds each.
The 2{,}641- and 3{,}002-run pools each add runs of one kind to the previous pool; the 8{,}727-run pool removes 22 runs of the 3{,}002-run pool before adding its own (below), so the pools are not nested, and the 9{,}112-run pool adds 385 runs to the 8{,}727-run pool.
The 2{,}641-run pool adds 1{,}723 public experiment-tracker runs, mostly pretraining (Marin ablations, optimizer benchmarks, and scaling sweeps), which are not isolated by project (53 of them, and 546 more in the 8{,}727-run pool, come from a Marin project that also supplies test runs of other sweeps).
The 3{,}002-run pool adds 361 Hugging Face Hub runs (356 RL, 5 fine-tuning), admitted under a family rule that treats a configuration default shared by unrelated owners as no relation; these are the only RL runs added to any pool.
The 8{,}727-run pool adds 5{,}747 public runs, almost all pretraining, and removes 22 Marin pretraining runs (180 training examples) that had entered with the 2{,}641-run pool and were carried into the 3{,}002-run pool.
None of the 22 is a test run; each links to a test run through a chain of sweep-family records collected for the 8{,}727-run pool, so the sweep-family isolation of \S\ref{sec:data} excludes it; the 918-, 8{,}727-, and 9{,}112-run pools contain none of them.
The readers trained on the 2{,}641- and 3{,}002-run pools, that is, every point of Figure~\ref{fig:reader} at those two pool sizes (both backbones, both inputs, all three stages), were therefore trained with the 22 linked records; the checks for the 918- and 8{,}727-run pools exempted the reference identities (broader relation graph below).
No added run has an audited description, so its semantics paragraph is rendered from its numeric configuration alone; the semantics variant therefore carries new textual evidence only for the 918 reference runs.

\paragraph{Broader relation graph.}
Under a broader graph of family and alias links, 462 of the 918 reference identities connect to an evaluation run. The check that admitted runs to the larger pools excluded additions with such a link but exempted the reference identities. Many of these links pass through broad family names, so a link does not imply a shared trajectory. The reported checkpoints and scores are unchanged.
Three caveats qualify Table~\ref{tab:reader-macro}: the 9B and 2B readers were trained on different clusters, so size is confounded with hardware and software; the rows of the 2{,}641- and 3{,}002-run pools include the 22 runs above; and community-run scores here are over all 232 runs with the \textsc{Full} prompts on both grids (464 views), not over the 211-run cohort of Figure~\ref{fig:main}.

The recipe was chosen on the reference pool after its frozen, RL, and community-run scores had been seen, so the reference-pool results and the test columns of the ablation tables (supplementary document, \S S1) are exploratory.
The three-seed runs of Table~\ref{tab:reader-macro} on the larger pools and with the 9B backbone were trained afterwards with the recipe unchanged, and every configuration of that matrix is reported. The only later choice made after test scores were seen is the 9{,}112-run pool, added after the community-run scores of the other pools had been seen. All of these runs reuse the recipe and so share its selection on test scores.

\subsection{Untrained backbones and three-stage averages}
\label{app:reader-macro}
The untrained backbones are scored in two ways (Table~\ref{tab:reader-macro}).
Under the benchmark prompts of \S\ref{sec:tasks}, the untrained Qwen3.5-2B is scored with the \textsc{Curve+Config} prompt, the input closest to the reader's, and reaches $S=0.58$ on pretraining and $0.18$ on fine-tuning (143 and 146 of 150 answers valid; it was not run on RL under that prompt); the untrained Qwen3.5-9B has no \textsc{Curve+Config} answers and is scored with the \textsc{Full} prompt, reaching $0.38$, $0.47$, and $0.03$ on pretraining, fine-tuning, and RL.
Under the reader's own prompt, both untrained backbones have negative skill on pretraining and RL, and their fine-tuning skill ranges from $-1.18$ to $0.03$; the trained-minus-untrained contrasts of Table~\ref{tab:significance-b}, which use this matched prompt, are positive for every backbone, input, seed, and stage.
The semantics raise the 9B three-stage average on every pool and lower the 2B average on the~three~larger~pools.

\begin{table}[t]
\centering\scriptsize\setlength{\tabcolsep}{3pt}
\caption{Reader by training pool, backbone, and input (skill score $S$, higher is better). PT and FT: the 150-run \textsc{Grid10} test sets; RL: the 96-run part of the RL set (both grids); Avg.: unweighted mean of PT, FT, and RL (bold: best among trained readers); PT+FT: the five pretraining and fine-tuning cells pooled; held-out sources and held-out families: validation splits; wild: the community set with \textsc{Full} prompts. Trained readers: mean of three training seeds. Untrained, benchmark: the benchmark prompt, \textsc{Curve+Config} for 2B (no RL answers) and \textsc{Full} for 9B. Untrained, reader: the reader's own prompt and ratio target without training, scored on valid answers (2B config / config + semantics: 131 / 44 pretraining and 149 / 111 fine-tuning answers of 150; 9B: 122 / 110 and 150 / 131; RL: 51--82 of 82 scoreable runs). LLM rows use the \textsc{Full} prompt. $^{\dagger}$This pool adds 385 Hugging Face runs of the same kind as the community set (no shared repository; 43 share an author, 19 a model family), so its community-run score is not a transfer test.}
\label{tab:reader-macro}
\label{tab:reader-scale}
\begin{tabular}{@{}llrrrrrrrr@{}}
\toprule
Pool & Model, input & PT & FT & RL & Avg. & PT+FT & \makecell{Held-out\\sources} & \makecell{Held-out\\families} & Wild \\
\midrule
918 (reference) & 2B, config & 0.688 & 0.607 & $-0.332$ & 0.321 & 0.659 & 0.680 & 0.237 & 0.267 \\
918 (reference) & 9B, config & 0.699 & 0.604 & $-0.085$ & 0.406 & 0.664 & 0.654 & 0.245 & 0.327 \\
918 (reference) & 9B, config + semantics & 0.701 & 0.603 & $-0.031$ & \textbf{0.424} & 0.665 & 0.622 & 0.244 & 0.289 \\
\midrule
2{,}641 & 2B, config & 0.678 & 0.495 & $-0.489$ & 0.228 & 0.615 & 0.757 & 0.223 & 0.292 \\
2{,}641 & 2B, config + semantics & 0.700 & 0.569 & $-0.770$ & 0.166 & 0.653 & 0.796 & 0.249 & 0.255 \\
2{,}641 & 9B, config & 0.708 & 0.527 & $-0.291$ & 0.315 & 0.648 & 0.722 & 0.233 & 0.266 \\
2{,}641 & 9B, config + semantics & 0.691 & 0.595 & $-0.186$ & 0.367 & 0.659 & 0.726 & 0.239 & 0.307 \\
\midrule
3{,}002 (+356 RL) & 2B, config & 0.686 & 0.592 & $-0.254$ & 0.341 & 0.652 & 0.778 & 0.222 & 0.298 \\
3{,}002 (+356 RL) & 2B, config + semantics & 0.699 & 0.616 & $-0.538$ & 0.259 & 0.668 & 0.784 & 0.219 & 0.310 \\
3{,}002 (+356 RL) & 9B, config & 0.679 & 0.590 & $-0.320$ & 0.316 & 0.646 & 0.713 & 0.223 & 0.282 \\
3{,}002 (+356 RL) & 9B, config + semantics & 0.711 & 0.639 & $-0.306$ & 0.348 & 0.685 & 0.729 & 0.222 & 0.286 \\
\midrule
8{,}727 & 2B, config & 0.689 & 0.602 & $-0.269$ & 0.341 & 0.659 & 0.810 & 0.271 & 0.292 \\
8{,}727 & 2B, config + semantics & 0.711 & 0.556 & $-0.413$ & 0.285 & 0.660 & 0.801 & 0.229 & 0.266 \\
8{,}727 & 9B, config & 0.711 & 0.675 & $-0.645$ & 0.247 & 0.697 & 0.790 & 0.217 & 0.229 \\
8{,}727 & 9B, config + semantics & 0.725 & 0.705 & $-0.322$ & 0.369 & 0.717 & 0.779 & 0.228 & 0.297 \\
\midrule
9{,}112 (+385 wild-class) & 2B, config & 0.714 & 0.599 & $-0.272$ & 0.347 & 0.673 & 0.809 & 0.290 & 0.288$^{\dagger}$ \\
\midrule
Untrained, benchmark & 2B, \textsc{Curve+Config} & 0.584 & 0.181 & -- & -- & -- & -- & -- & -- \\
Untrained, benchmark & 9B, \textsc{Full} & 0.375 & 0.470 & 0.032 & 0.293 & -- & -- & -- & -- \\
Untrained, reader & 2B, config & $-12.1$ & $-1.18$ & $-3.40$ & -- & -- & -- & -- & -- \\
Untrained, reader & 2B, config + semantics & $-5.74$ & $0.03$ & $-3.74$ & -- & -- & -- & -- & -- \\
Untrained, reader & 9B, config & $-19.5$ & $-1.15$ & $-3.11$ & -- & -- & -- & -- & -- \\
Untrained, reader & 9B, config + semantics & $-15.4$ & $-0.52$ & $-3.06$ & -- & -- & -- & -- & -- \\
\midrule
LLM & \opus{} & 0.697 & 0.559 & 0.385 & 0.547 & -- & -- & -- & -- \\
LLM & \sol{} & 0.659 & 0.653 & 0.312 & 0.542 & -- & -- & -- & -- \\
LLM & Muse-Glimmer-30B & 0.350 & 0.564 & 0.108 & 0.341 & -- & -- & -- & -- \\
\bottomrule
\end{tabular}
\end{table}

\subsection{Ablations}
\label{app:ablations}
The recipe was ablated on validation runs, scored by the geometric mean of the point and pinball-loss error ratios against the last value and read three ways (macro average over all cells, macro average over cells with at least 20 examples, and the per-example paired ratio), with a verdict only when two readings agree; the four tables and the full readings are in the supplementary document (\S S1).
The numeric configuration is the one large step on the input axis, and it improves the intervals rather than the median (point ratio $0.524\rightarrow0.523$, pinball ratio $1.69\rightarrow0.64$).
At 2B, schedule facts (combined ratio $0.576\rightarrow0.570$), the semantics paragraph added to the configuration (paired ratio $1.03$ $[0.95,1.10]$), and a placebo that gives each example another run's schedule facts ($1.00$ $[0.91,1.12]$) change nothing; run-level semantics trained on the relaxed pool is worse ($1.41$ $[1.19,1.71]$), and the RL cost of the semantics at 2B appears on the test sets (Table~\ref{tab:reader-macro}).
The semantics ratio at 9B ($0.91$ $[0.77,1.04]$, against $1.12$ $[1.02,1.23]$ at 0.8B) comes from RL examples, whose semantics paragraph adds only numeric fields; on pretraining and fine-tuning alone the combined ratio is 1.02, 1.03, and 1.02 at 0.8B, 2B, and 9B.
A Qwen3 ladder (1.7B, 8B, 14B; one seed per rung) shows no gain from scale; at 14B the reading for the semantics is mixed (a paired ratio of $0.89$ $[0.76,1.00]$, with the all-cell average pointing the other way), so the 14B threshold reported for context-aided forecasting~\citep{beyondnaive2025} is neither confirmed nor refuted.
Training for 3{,}000 steps lowers validation error without changing the frozen test sets; adding a short-observation mix does the same and worsens the community set, as does full fine-tuning; removing Marin from training is worse on validation, leaves the frozen sets unchanged, and helps on RL and community runs, while Marin alone is worse everywhere; a quarter of the training runs is worse ($1.24$ $[1.08,1.45]$) and half is not distinguishable from all ($1.09$ $[0.99,1.20]$); LoRA rank 256 does not help, and an inline rationale before the numbers and one retrieved related run at training and inference are worse.
Retrieving three related runs helps on the held-out-sweep split ($0.80$ $[0.68,0.90]$) and hurts on the held-out-source split ($1.29$ $[1.15,1.47]$); telemetry curves have no effect ($1.00$ $[0.95,1.05]$).
Frontier-model forecasts are not used as a reader input: in a predeclared validation check, the mean of three such forecasts has 0.48 $[0.36,0.64]$ of the reader's error on 41 held-out-source runs but 1.49--1.69 times its error on 74 held-out-family runs, so no fixed combination beats both.

\FloatBarrier

\subsection{The strongest reader we trained}
\label{app:reader-v2}
\textbf{Setting.} After the experiments of \S\ref{sec:training} we rebuilt the reader's data and evaluation to find the best model the public record supports.
Recipes are compared on a development panel of 350 runs (100 pretraining, 100 fine-tuning, 100 RL, and 50 preference-tuning runs) built with the protocol of the test sets and selected by a hash rule before any result was seen; every lineage of the 146 owner units that contribute a run to it (1{,}099 lineages) is removed from training, and none of its runs is connected to a test run in the relation graph used to isolate the training pools (Appendix~\ref{app:reader-pools}).
The training set holds 75{,}166 examples from 8{,}691 runs (1{,}172 pretraining, 4{,}167 fine-tuning, 2{,}001 preference-tuning, 1{,}349 RL, and 2 other runs).
The reader sees ten values that summarize the first 20\% of the curve (window means on the panel), the names of the metric and the axis, and two recorded configuration fields, and answers with three quantiles per query point as ratios to the last observed value; the interval targets are residual quantiles of a parametric fit, cross-fitted over lineages.
A recipe is a backbone and a number of training draws (examples sampled with replacement, eight per LoRA update).
Skill is $S$ on \textsc{Grid10} with an answer that does not parse scored as the last value, pooled over the pretraining, fine-tuning, and RL cells, with a 95\% bootstrap interval over lineages.

\begin{table}[t]
\centering\small
\caption{Skill $S$ on the development panel (350 runs). Draws: training examples sampled with replacement. The reference recipe is the 2B recipe of \S\ref{sec:training} with the new interval targets, trained on 1{,}661 runs; its three rows are three training seeds. Coverage: share of targets inside the 80\% interval.}
\label{tab:reader-v2}
\begin{tabular}{@{}lrrrrrlr@{}}
\toprule
Model & draws & PT & FT & RL & pref. & PT+FT+RL $[95\%]$ & coverage \\
\midrule
AIC ensemble & -- & 0.52 & 0.03 & $-0.38$ & 0.13 & 0.08 $[-0.06, 0.18]$ & 0.00 \\
LC-PFN & -- & 0.74 & 0.32 & $-0.09$ & 0.33 & 0.37 $[0.31, 0.42]$ & 0.53 \\
Chronos-2, fine-tuned & -- & 0.81 & 0.49 & 0.03 & 0.27 & 0.49 $[0.43, 0.54]$ & 0.73 \\
\midrule
Reference recipe, 2B & 8k & 0.60 & 0.27 & 0.05 & 0.43 & 0.31 $[0.25, 0.37]$ & 0.83 \\
Reference recipe, 2B & 8k & 0.53 & 0.25 & 0.12 & 0.39 & 0.30 $[0.24, 0.35]$ & 0.82 \\
Reference recipe, 2B & 8k & 0.58 & 0.26 & 0.12 & 0.36 & 0.32 $[0.27, 0.38]$ & 0.83 \\
\midrule
0.8B & 8k & 0.57 & 0.51 & 0.00 & 0.29 & 0.36 $[0.30, 0.42]$ & 0.80 \\
0.8B & 32k & 0.80 & 0.44 & 0.17 & 0.47 & 0.50 $[0.46, 0.54]$ & 0.81 \\
0.8B & 60k & 0.70 & 0.45 & 0.19 & 0.51 & 0.45 $[0.41, 0.50]$ & 0.80 \\
\ours 2B & 32k & 0.82 & 0.44 & 0.22 & 0.54 & 0.53 $[0.49, 0.56]$ & 0.80 \\
2B & 60k & 0.79 & 0.44 & 0.22 & 0.57 & 0.51 $[0.47, 0.54]$ & 0.82 \\
\bottomrule
\end{tabular}
\end{table}

\textbf{Readings.} Table~\ref{tab:reader-v2} gives the skill of each recipe and of the numeric forecasters on the panel.
The strongest reader is the 2B backbone trained for 32{,}000 draws, with $S=0.53$ $[0.49, 0.56]$: $0.82$ on pretraining, $0.44$ on fine-tuning, and $0.22$ on RL.
It is $0.21$ to $0.23$ above the reference recipe ($0.31$, $0.30$ and $0.32$ in three seeds, with $0.05$, $0.12$ and $0.12$ on RL) and $0.16$ above LC-PFN ($0.37$, with $-0.09$ on RL).
The gain comes from the runs and the length of training, not from the size of the model: at 0.8B, 32{,}000 draws give $0.50$ against $0.36$ for 8{,}000, 60{,}000 draws do not add to it ($0.45$ at 0.8B and $0.51$ at 2B), and the 2B backbone is within $0.06$ of the 0.8B backbone at equal draws ($0.53$ against $0.50$, and $0.51$ against $0.45$).
Three seeds of the reference recipe differ by $0.03$ on the pooled score and by up to $0.08$ on a single stage, so differences below $0.08$ between rows are not interpreted; the three strongest rows lie within this margin of one another.
A time-series model fine-tuned on curves from the same pool, Chronos-2 with 120M parameters, reaches $0.49$; the reader's margin over it lies in RL ($0.22$ against $0.03$) and preference tuning ($0.54$ against $0.27$), and its intervals are closer to nominal (coverage $0.80$ against $0.73$).
The rows other than the reference have one training seed, and the reader has not been evaluated on the test sets of \S\ref{sec:data}.

\section{Database of runs: details}
\label{app:det-retrieval}
Appendix~\ref{app:matched} adds baselines that receive the same retrieved neighbors.

This appendix gives the protocol and the full results behind \S\ref{sec:retrieval}.

The database holds 1{,}726 training runs with their configuration, code, and lineage.
For each of 125 pretraining and fine-tuning questions we embed the configuration text with Gemini \texttt{gemini-embedding-2}, retrieve the $k$ nearest runs outside the question's sweep family, and add their complete curves to the \textsc{Full} prompt of \sol{}; the score is the paired change in error against the same model's answer without retrieval.
Smaller databases are subsamples of 1, 5, and 25\% of the runs.
Table~\ref{tab:retrieval} (top left) gives the gain by database size, by $k$, and by quartile of the nearest neighbor's cosine similarity; ratios for further retrievers and for \luna{} are in Table~\ref{tab:significance-c}.

\paragraph{Questions whose own series is in the database.}
Leave-family-out retrieval removes the question's sweep family, but a project can group several sweep families into one coarser experiment series (for example, one study launched as several sweeps).
For 24 of the 123 valid questions, all Marin pretraining, at least one of the three runs retrieved from the full database belongs to the question's series though not to its sweep family.
Figure~\ref{fig:retrieval}a draws these 24 questions hollow at every database size (the group is fixed by the full-database retrieval; in the 1, 5, and 25\% subsamples, 6, 14, and 21 of them retrieve such a run) and the other 99 filled.
With the full database their error ratio is 0.58 $[0.34,0.96]$ (a 42\% reduction) against 0.61 $[0.47,0.79]$ for the other 99; with the smallest database it is 1.15 $[0.84,1.61]$ against 0.85 $[0.72,0.99]$, so the gain does not come from runs of the~question's~own~series.

\begin{table}[t]
\centering\small
\caption{A database of 1{,}726 past runs, queried on the same 125 questions with \sol. Top left: all valid questions pooled (Figure~\ref{fig:retrieval}a separates the two question groups); paired change in MAE against the same model without retrieval (negative is better) and the geometric mean of the ratio of the two errors, by database size, by $k$, and, for the full database, by quartile of the nearest neighbor's similarity. Top right: top-1 accuracy of the sweep recommender; the best label-free prior picks the value most common in the database. Bottom: MAE per cell (in each source's own units, 25 planned questions per cell; each method on its own valid answers) against the better of two regressors trained on the same runs (kNN on prefix shape and mean training shape) under the regimes of Appendix~\ref{app:regimes-main}; absolute levels are not comparable with Table~\ref{tab:regimes}. $^{\ast}$One question: the other 24 MLFoundations questions have no run of their own sweep family in the database.}
\label{tab:retrieval}
\setlength{\tabcolsep}{3pt}
\begin{tabular}{lcc@{\hskip 1.5em}lc}
\toprule
\sol{} arm & $\Delta$MAE & ratio & \multicolumn{2}{l}{\emph{Recommender: 70 questions, 7 settings}} \\
\midrule
Library 17 runs, $k{=}3$ & $-0.039$ & 0.90 & \sol{} + retrieved sweeps & \textbf{0.61} \\
\quad 86 runs & $-0.027$ & 0.91 & \sol, configurations only & 0.46 \\
\quad 431 runs & $-0.040$ & 0.71 & Best label-free prior & 0.37 \\
\quad 1{,}726 runs & $-0.063$ & \textbf{0.61} & Random & 0.39 \\
\makecell[l]{$k{=}1$\\\quad / $k{=}5$} & \makecell{$-0.063$\\/ $-0.052$} & \makecell{0.75\\/ 0.63} & & \\
\cmidrule(r){1-3}
Nearest neighbor, $Q_1$ & $+0.003$ & 0.85 & & \\
\quad $Q_2$ & $-0.098$ & 0.60 & & \\
\quad $Q_3$ & $-0.056$ & 0.47 & & \\
\quad $Q_4$ & $-0.102$ & 0.57 & & \\
\bottomrule
\end{tabular}

\vspace{0.6em}
\begin{tabular}{lccccc}
\toprule
Method (isolation) & FT/HF & FT/Marin & PT/AI2 & PT/Marin & PT/MLF \\
\midrule
Best regressor, leave-family-out & 0.117 & 0.154 & \textbf{0.210} & 0.270 & 0.231 \\
\sol, no retrieval & 0.057 & 0.031 & 0.410 & 0.262 & 0.237 \\
\sol{} + retrieval, leave-family-out & \textbf{0.049} & \textbf{0.027} & 0.240 & \textbf{0.186} & \textbf{0.182} \\
\midrule
Best regressor, in-distribution & 0.120 & 0.0037 & 0.209 & 0.209 & 0.196 \\
\sol{} + retrieval, siblings allowed & \textbf{0.040} & \textbf{0.0021} & \textbf{0.150} & \textbf{0.108} & 0.039$^{\ast}$ \\
\bottomrule
\end{tabular}
\end{table}

The gain grows with the database because the nearest neighbor becomes closer: from the smallest to the full database the mean cosine similarity of the best neighbor rises from 0.870 to 0.944, and the share of neighbors from the question's own project from 0.20 to 0.93.
Retrieval leaves a heavy tail: 11--16\% of questions become more than twice as bad, and neither retrieving only above a similarity threshold (chosen on half of the questions, no effect on the other half) nor listing each neighbor's configuration differences from the question ($-0.023$ $[-0.067,+0.005]$ over the plain neighbors)~removes~it.

In Table~\ref{tab:retrieval} (bottom), the regressor means use all 25 questions per cell, whereas the two leave-family-out \sol{} rows have 25/24/25/25/24 valid questions in column order and the siblings-allowed row 25/24/24/25/1. Restricting both regressors to each retrieval arm's valid questions gives leave-family-out MAE 0.117/0.155/0.210/0.270/0.234 and sweep-seen MAE 0.120/0.0036/0.215/0.209/0.213 for the better regressor; \sol{} remains lower in four of five leave-family-out cells and in all four sufficiently populated sweep-seen cells. ``Siblings allowed'' means that the question's own sweep family may be retrieved.
The recommender questions are built only from training-pool sweeps and ask which of the values actually tried for one of seven settings ends lowest; retrieved sweeps lower the normalized regret from 0.31 to 0.19.
The recommender is perfect on the 15 depth and model-scale questions, at 51\% on the 45 learning-rate questions, and correct on one of the four that differ only in the random seed (random-choice expectation 46\%); the remaining six questions cover batch size (one), budget (two), expert count (one), and warmup (two). The pool holds no controlled sweep of schedule shape, weight decay, data mixture, or any RL setting.

\FloatBarrier

\section{Coding agent: details}
\label{app:det-agent}
\label{app:analysis-details}
This appendix gives the protocol of the agent study of \S\ref{sec:agent}, which compares three arms (\emph{single call}, \emph{agent}, and \emph{agent + search}), and a pilot that compared agents with single calls given the curves~of~related~runs.

\subsection{Protocol of the agent study}
Each question becomes an offline folder with the prefix, the configuration, the description, and the code excerpts of the \textsc{Full} prompt.
In the \emph{agent} arm the model can list, read, and write files in that folder and run Python in a network-isolated sandbox that mounts only the folder, for up to twenty turns; the \emph{agent + search} arm adds literature search and public-code search, and the tool layer, not the prompt, removes every result from the run's own project (no leaked run identifier or target value was found among the returned items).
The \emph{single call} is the frozen \textsc{Full} answer of \S\ref{sec:main}.
The question set is the first 100 pretraining and first 100 fine-tuning questions in a fixed task-identifier order, plus 50 RL questions, each run by \opus{} and \sol{} in both agent arms (1{,}000 planned attempts). The question set was cut from a planned 447 questions after the runs had begun, by task-identifier order and independently of answer quality and completion; the cut was not planned in advance.
Six attempts were not run after a provider outage and are not imputed; answers recorded for them in a later run are excluded from the scores.
Provider errors, not the models, account for most missing answers: \sol{} returned a valid forecast on 81--87 of the 100 pretraining or fine-tuning questions per stage in the two agent arms (48--50 of the 50 RL questions) and \opus{} on at least 96 of 100.
On average an agent attempt took 5.2--5.3 turns without search and 7.2--8.4 with search, and cost \$0.17 (\opus) and \$0.08 (\sol) per question without search and \$0.58 and \$0.13 with it.
Figure~\ref{fig:agent} averages agent cost over completed attempts. Its single-call reference for pretraining and fine-tuning is the average cost of the \textsc{Grid10} requests of the main comparison, whereas the agent questions use \textsc{Grid20}, so those cost multipliers are approximate. The RL reference uses the \textsc{Grid20} cost records of the same questions.
Paired contrasts of each arm against the single call, per stage, are in Table~\ref{tab:significance-b}.

\subsection{Pilot: related runs against agent behavior}
\label{app:agent-table}
Before the agent study, a pilot on 50 pretraining and fine-tuning questions (ten per source) compared a plain single call with calls and agents given the complete curves of two to five runs from the same project but a different sweep (Table~\ref{tab:agent}). Arms~2a--4 receive these curves; arm~2b also receives training code, matching the agent's available files.
The related curves reduce the error of \opus{} by 24\% (paired difference $-0.037$ $[-0.075,-0.005]$, 49 questions) and of \sol{} by 44\% ($-0.086$ $[-0.166,-0.031]$, 50 questions), whereas the offline agent changes the error of the single call that holds the same curves by $+0.002$ $[-0.037,+0.066]$, and search changes the error by $-0.005$ $[-0.018,+0.009]$ (\opus) and $+0.001$ $[-0.006,+0.013]$ (\sol).
Without the related curves (arm~4$'$) the agent with search is no better than a plain single call.
The lower half of Table~\ref{tab:agent} summarizes the logs: agents fitted a parametric law in most attempts, no agent revised a written forecast, and with the related curves the per-question errors of the agents correlate with those of the single call that holds the same curves (0.62--0.97).

\begin{table}[t]
\centering\scriptsize\setlength{\tabcolsep}{4pt}
\caption{The agent pilot on 50 pretraining and fine-tuning questions. Top: MAE (46--50 answered per condition); arms 2a--4 see the same related runs; arm~3 is an agent in an external coding-agent harness, and arms~4 and~4$'$ are agents + search in our harness. Bottom: what the agents did, from the harness logs of the 250 pilot attempts (50 questions $\times$ 5 arm--model pairs). ``Read'' counts attempts that opened a related curve or a source file beyond the manifest; ``fitted'' counts attempts whose executed code called an optimizer or regression on a power-law, exponential, or logarithmic form, or ran the provided fitting script. $^a$Arm~3 ran in a different harness whose shell calls often read several files at once, so its call counts are not comparable. $^b$Not metered; estimated from token counts at list price. The single call with related curves (arm 2a) costs \$0.05 per question.}
\label{tab:agent}
\label{tab:agent-behavior}
\begin{tabular}{lcc}
\toprule
Condition & \opus & \sol \\
\midrule
1 \; Single call, \textsc{Full} prompt & 0.152 & 0.192 \\
2a \; + complete curves of 2--5 related runs & 0.116 & \textbf{0.107} \\
2b \; + related runs + training code & \textbf{0.101} & 0.108 \\
3 \; Agent, same files (external harness) & 0.118 & -- \\
4 \; Agent + search (literature and code) & 0.111 & 0.108 \\
4$'$ \; Agent + search, no related runs & 0.139 & 0.216 \\
\bottomrule
\end{tabular}

\vspace{0.8em}
\begin{tabular}{lccccc}
\toprule
 & Arm 3 & \multicolumn{2}{c}{Arm 4} & \multicolumn{2}{c}{Arm 4$'$} \\
\cmidrule(lr){2-2}\cmidrule(lr){3-4}\cmidrule(lr){5-6}
 & \opus{} & \opus{} & \sol{} & \opus{} & \sol{} \\
\midrule
Model turns, mean [range]            & 9.0 [7--12] & 8.9 [6--14] & 8.6 [6--13] & 11.4 [6--16] & 10.1 [7--15] \\
Tool calls per question$^a$          & 10.3 & 15.7 & 26.1 & 17.0 & 20.9 \\
Python executions per question       & 3.4 & 3.1 & 2.6 & 3.2 & 3.4 \\
Read the related curves              & 50/50 & 48/50 & 50/50 & -- & -- \\
Read the training code               & 16/50 & 34/50 & 50/50 & 40/50 & 50/50 \\
Fitted a parametric law              & 47/50 & 33/50 & 46/50 & 47/50 & 50/50 \\
Searches per question (paper / code) & -- & 1.2 / 1.7 & 1.3 / 2.4 & 1.4 / 4.6 & 1.7 / 3.5 \\
Code searches answered by the index  & -- & 52/87 & 50/86 & 13/231 & 35/171 \\
Retrieved items shown to the model   & -- & 686 & 1{,}115 & 398 & 399 \\
Items dropped by the project filter  & -- & 68 & 98 & 0 & 0 \\
Changed a written forecast           & 0/50 & 0/50 & 0/50 & 0/50 & 0/50 \\
Output tokens per question           & 6.1k & 6.5k & 3.8k & 8.2k & 3.7k \\
Cost per question (\$)               & $\approx$0.55$^b$ & 0.70 & 0.23 & 0.98 & 0.24 \\
Median wall clock (s)                & 80 & 121 & 95 & 135 & 60 \\
Correlation of per-question error with arm 2a & 0.62 & 0.96 & 0.97 & 0.69 & 0.51 \\
\bottomrule
\end{tabular}
\end{table}

\FloatBarrier

\section{Prompts and a full example}
\label{app:prompts}
This appendix gives worked examples for the tasks of \S\ref{sec:tasks} (one complete question with its \textsc{Full} input and the answer, three forecasts per cell, matching rationales, and one agent transcript) and the prompts of the prediction, matching, and agent tasks.

\subsection{Full examples}
\label{app:examples}
{\scriptsize
\paragraph{One complete question and answer.}
\label{app:example-full}
The pretraining run of Figure~\ref{fig:overview} (case \texttt{92793c43c27fab0a87980503}, Marin, \textsc{Grid10}) as \opus{} saw it in the \textsc{Full} condition, and the answer it returned.
The request is the prediction prompt of Appendix~\ref{app:prompt-text}, a line \texttt{INPUT:}, and the input below as one line of compact JSON with sorted keys (3{,}530 bytes); only the layout is changed for display (the code excerpt, one JSON string in the request, is printed after the object), and every value is byte-exact.
\begin{verbatim}
<PREDICTION_PROMPT>
INPUT:
{"axis": "reported log index divided by configured update budget; not completed-update count",
 "case": "92793c43c27fab0a87980503",
 "context": {
   "code_context": [{"binding": "project_level_example_not_run_revision",
      "code": "<CODE, below>", "language": "python", "line_end": 337, "line_start": 322,
      "source_sha256": "ae88a8d1e95b7bcd75710bb2913fbcadf429e17257a8ef3fec23f815c49a0024",
      "source_url": "https://github.com/marin-community/marin"}],
   "interpretation_rule": "Reported per-run configuration takes precedence over source example
      defaults. Project-level code is explanatory context, not proof of execution. Missing
      information is unknown, not zero.",
   "known_unknowns": ["Readable training data mixture and tokenizer identity are not linked
      for this run.", "This source example is not bound to a recorded run revision."],
   "reported_configuration": {"model.hidden_dim": 512, "model.max_seq_len": 8192,
      "model.num_heads": 8, "model.num_layers": 1,
      "optimizer.learning_rate": 0.0028575255750067892, "optimizer.warmup": 0.01,
      "optimizer.weight_decay": 0.1, "planned_step_budget": 1058, "trainer.log_every": 1,
      "trainer.trainer.num_train_steps": 1058, "trainer.trainer.seed": 0,
      "trainer.trainer.train_batch_size": 64, "trainer.z_loss_weight": 0.0001},
   "source_facts": ["Do not assume that train/loss is pure cross-entropy: project source can
      include log-sum-exp and router auxiliary penalties."],
   "task_description": "Language-model pretraining candidate from the Marin MoE experiment
      collection. Forecast the reported training objective loss using the early trajectory and
      the reported architecture, optimizer and training-budget configuration."},
 "grid": 10,
 "metric": "reported train/loss; Grug objective/version pending",
 "prefix": [{"value": 11.209885215759277, "x": 0.02}, {"value": 10.238005956013998, "x": 0.04},
            {"value": 9.437644867669968, "x": 0.06},  {"value": 8.730281375703358, "x": 0.08},
            {"value": 8.179786023639497, "x": 0.1},   {"value": 7.7530267125084285, "x": 0.12},
            {"value": 7.474150050770152, "x": 0.14},  {"value": 7.189137186322894, "x": 0.16},
            {"value": 6.961845102764311, "x": 0.18},  {"value": 6.787371090480259, "x": 0.2}],
 "query_x": [0.3, 0.4, 0.5, 0.6, 0.7, 0.8, 0.9, 1.0],
 "task_semantics": "Predict or assign the recorded metric; source-specific objective terms may
    differ. Missing context is unknown."}

<CODE>
def loss_fn(params):
    compute_params = mp.cast_to_compute(params)
    return compute_params.next_token_loss(
        batch.tokens,
        batch.loss_weight,
        mask=batch.attn_mask,
        reduction="mean",
        logsumexp_weight=z_loss,
        return_router_metrics=True,
    )

(loss, summarized_metrics), grads = jax.value_and_grad(loss_fn, has_aux=True)(qb_params)
metrics = {"train/loss": loss, **summarized_metrics}
updates, opt_state = optimizer.update(grads, state.opt_state, qb_params)
params = optax.apply_updates(qb_params, updates)
\end{verbatim}
The answer of \opus{} (999 output tokens; the parsed JSON object, with non-ASCII arrows and signs written in ASCII):
\begin{verbatim}
{"predictions": [{"x": 0.3, "p10": 5.85, "p50": 6.2,  "p90": 6.55},
                 {"x": 0.4, "p10": 5.5,  "p50": 5.9,  "p90": 6.3},
                 {"x": 0.5, "p10": 5.28, "p50": 5.7,  "p90": 6.12},
                 {"x": 0.6, "p10": 5.1,  "p50": 5.54, "p90": 5.98},
                 {"x": 0.7, "p10": 4.95, "p50": 5.4,  "p90": 5.85},
                 {"x": 0.8, "p10": 4.8,  "p50": 5.26, "p90": 5.75},
                 {"x": 0.9, "p10": 4.65, "p50": 5.12, "p90": 5.62},
                 {"x": 1.0, "p10": 4.48, "p50": 4.98, "p90": 5.5}],
 "rationale": {
   "evidence": [
     "Prefix falls 11.21->6.79 over x=0.02->0.20 with rapidly shrinking increments (1.39 over
      last doubling).",
     "Very small model (num_layers=1, hidden_dim=512, 8 heads) limits attainable loss despite
      8192 seq len and batch 64.",
     "z_loss_weight=1e-4 plus possible router aux terms add a small positive offset to
      reported train/loss."],
   "assumptions": [
     "Loss continues an approximately power-law/log decay with ~0.5-0.6 nats per doubling of x,
      flattening late.",
     "Cosine-style LR decay near end of the 1058-step budget yields a modest extra drop after
      x~0.8.",
     "No divergence, restarts, or mixture change in the remaining run."],
   "uncertainty": "Moderate: shape of the tail is extrapolated from only 20% of the run; final
      value uncertain by roughly +/-0.5 nats, with asymmetric downside if late LR decay is
      strong."}}
\end{verbatim}
The recorded values at the eight query points are 6.20, 5.89, 5.70, 5.50, 5.42, 5.28, 5.22, and 5.14.
The median forecast has MAE 0.045 against 1.24 for repeating the last value ($R = 0.036$, $S = 0.96$), and all eight targets fall inside the 80\% intervals; the same model given the curve alone (\textsc{Curve}) has MAE 0.58 ($R = 0.47$) on this run.
}

\subsection{Prompts}
\label{app:prompt-text}
{\scriptsize

\paragraph{Prediction prompt.} Shared by the three prediction conditions, which differ only in the \texttt{INPUT} object that follows it.

\begin{verbatim}
Forecast the recorded metric at every query_x from only the supplied prefix and
available context. No browsing, experiment identification, or remembered final
results. Return one JSON object: {"predictions":[{"x":number,"p10":number,
"p50":number,"p90":number}],"rationale":{"evidence":[short observations citing input
fields],"assumptions":[short assumptions],"uncertainty":short string}}. Cover
exactly query_x in order; use finite ordered quantiles in the original metric units.
Rationale must be concise, at most three evidence items and three assumptions.
Missing information is unknown. Do not provide a chain of thought.
\end{verbatim}

\paragraph{Matching prompt.} Shared by the two matching conditions.

\begin{verbatim}
Assign the two visible recorded-metric curves to the reported settings A and B using
only the supplied input. Curve order is arbitrary. This is assignment, not causal
identification; unobserved settings may differ. Return one JSON object:
{"probability_A_curve1":number,"rationale":{"evidence":[short observations citing
input fields],"assumptions":[short assumptions],"uncertainty":short string}}.
Probability must lie in [0,1]; the alternative assignment is its complement.
Rationale must be concise, at most three evidence items and three assumptions. No
browsing, experiment identification, or remembered final results. Do not provide a
chain of thought.
\end{verbatim}
}

\FloatBarrier

\section{From forecasts to decisions: pilot studies}
\label{app:decisions}

This appendix gives the definitions behind \S\ref{sec:decisions}, the full historical replay, the protocol and complete readings of the controlled sweep, the direct pairwise task, the same pairwise question on historical sweeps, a simulation that maps skill to saving, and a second controlled sweep with described changes of both signs.

\subsection{Policy and metrics}
A group is a set of $n\ge 4$ comparable runs of one sweep, family or project under fixed grouping rules (the eligible groups of the seven primary pools: PT/Marin 6, PT/DCLM 4, PT/AI2 1, FT/HF 3, FT/Marin 10, RL release 6, RL supplement 3).
At the cutoff $t=0.2$ every forecaster sees the same prefix and context and predicts each run's final value; the policy keeps the $m$ runs with the best predicted final value and trains them to completion ($m=1$, $\lceil n/4\rceil$, $\lceil n/2\rceil$).
With equal-cost runs the budget is $(0.2n+0.8m)/n$ of training everything, 0.42--0.47 on average in the six displayed pools at $m=\lceil n/4\rceil$, and is set by the policy, not by the forecaster.
Regret is the final value of the best kept run minus the group's best final value, in the metric's units or normalized by the group's range; missed-good is the share of groups whose kept set contains no near-best run; false-stop is the share of near-best runs not kept; near-best means within 5\% of the group's range (2\% and 10\% as sensitivity).
Ties in a forecaster's ranking are scored by expectation over orders.
The pairwise wrong-order rate counts, over all pairs of a group whose final values differ by more than 5\% of the range, the pairs that the forecasts at $t$ order wrongly.
Persistence (repeating the last observed value) and a random policy are always included.

\begin{table}[t]
\centering\scriptsize\setlength{\tabcolsep}{2.5pt}
\caption{Summary of the two pilot studies. Left: normalized regret of the best kept run, $\lceil n/4\rceil$ kept at 20\%, by pool (groups/runs). Right: controlled 72-run sweep: regret (nats) keeping one run, and accuracy of the direct ``which of two runs ends lower'' question on the 33 cooldown and 36 code post-cutoff pairs. C: curve + configuration; +D: plus a description of the code change. $^{\ddagger}$Secondary: more than 5\% of the arm's answers replaced by persistence.}
\label{tab:decisions}
\begin{minipage}[t]{0.47\linewidth}\vspace{0pt}\centering
\resizebox{\linewidth}{!}{%
\begin{tabular}{@{}lcccc@{}}
\toprule
Pool (groups/runs) & random & persist. & \sol{} & \opus{} \\
\midrule
PT Marin (6/63) & 0.149 & 0.000 & 0.000 & 0.022 \\
PT DCLM (4/22) & 0.327 & 0.065 & 0.000 & 0.000 \\
FT HF (3/36) & 0.172 & 0.000 & 0.037 & 0.000 \\
FT Marin (10/100) & 0.074 & 0.000 & 0.000 & 0.026 \\
RL release (6/50) & 0.173 & 0.000 & 0.001 & 0.000$^{\ddagger}$ \\
RL supplement (3/37) & 0.197 & 0.274 & 0.274 & 0.000$^{\ddagger}$ \\
\bottomrule
\end{tabular}}
\end{minipage}\hfill
\begin{minipage}[t]{0.52\linewidth}\vspace{0pt}\centering
\resizebox{\linewidth}{!}{%
\begin{tabular}{@{}llccc@{}}
\toprule
Method & input & regret & cooldown & code \\
\midrule
persistence & curve & 0.060 & 0.455 & 0.458 \\
exclude described & +D & 0.015 & 0.455 & 1.000 \\
\sol{} & C / C+D & 0.082 / 0.027 & 0.242 / 0.455 & 0.431 / 1.000 \\
\opus{} & C / C+D & 0.050 / 0.035$^{\ddagger}$ & 0.773 / 0.727 & 0.444 / 1.000 \\
\bottomrule
\end{tabular}}
\end{minipage}
\end{table}

\subsection{Historical replay}
Table~\ref{tab:decisions} gives the normalized regret at $m=\lceil n/4\rceil$.
Across the main LLM forecasters and the groups of the seven pools, 26 forecaster--group comparisons beat persistence, 366 tie and 141 lose; no forecaster is significantly better than persistence in any pool (pooled sign tests; \sol{} full context 1/30/2, \opus{} full context 2/24/7).
Persistence already keeps an optimal run in PT/Marin, FT/HF, FT/Marin and RL release.
At $m=\lceil n/4\rceil$ on PT/Marin, false-stop rates are 0.08 for persistence (equal to the oracle, since more runs are near-best than can be kept), 0.14 for \sol{} and LC-PFN and 0.31 for \opus{}; missed-good rates are 0, 0, 0 and 0.17.
Under successive halving (keep $\lceil n/2\rceil$ at 20\%, $\lceil n/4\rceil$ at 35\%, one at 50\%) persistence's regret is at most 0.0068 nats on pretraining and fine-tuning and 0.0058 reward units on the RL release.
At earlier cutoffs that leave headroom the LLMs do not recover it: on PT/Marin at 10\% the regret is 0.10 nats for persistence, 0.22 for \sol{} and 0.009 for LC-PFN.
Across the LLM forecasters within a pool, regret correlates with the pairwise wrong-order rate (Spearman 0.46--0.87 in all six displayed pools) and more weakly with the skill score, where higher skill goes with lower regret in five pools (Spearman $-0.29$ to $-0.57$) and with higher regret in the RL supplement ($+0.70$); these are descriptive associations among dependent arms.
Selecting by a best-run probability derived from \sol's quantiles (2{,}000 split-normal draws) finds the best run in 15 of 33 groups against 23 for persistence, and promoting by the optimistic bound (p10 for loss, p90 for reward) is worse than persistence (5 wins, 12 ties, 16 losses, $p=0.027$), although that probability is close to its realized rate (0.42 against 0.45) and \sol's 80\% intervals cover 0.80 of final values (308 runs).

\textbf{Budget at equal quality.} Table~\ref{tab:budget-quality} reads the replay the other way round: for each forecaster, the smallest tested promotion count $m\in\{1,\lceil n/4\rceil,\lceil n/2\rceil\}$ at which mean normalized regret is at most 0.01 (about 1\% of a group's range). With equal-cost runs the budget of keeping $m$ runs is $0.2+0.8\,m/n$, averaged over the groups of a pool. On the six Marin pretraining sweeps (sizes 16, 14, 14, 8, 6, 5) \sol{} with full context meets the target keeping one run per sweep, mean budget 0.293, where persistence needs $\lceil n/4\rceil$ (0.441); on the four DCLM sweeps (9, 5, 4, 4) \sol{} and \luna{} meet it at $m=1$ (0.362) where persistence needs $\lceil n/2\rceil$ (0.631): 33.5\% and 42.6\% less total training budget, or 61\% and 62\% less after the cutoff, as retrospective point estimates (\sol's Marin regret at $m=1$ is 0.010 [0.000, 0.023]). What matters is the order at the selection boundary, which neither a higher skill score (\opus{}, $S=0.80$ on PT/Marin, has regret 0.193 at $m=1$ and 0.012 at $\lceil n/2\rceil$) nor a low wrong-order rate (on DCLM, \sol{} with the curve only has rate 0.086 and regret 0.178 at $m=1$) guarantees. In FT/Marin and the RL release persistence already meets the target at $m=1$, so the prefix suffices there.

\begin{table}[t]
\centering\scriptsize\setlength{\tabcolsep}{3pt}
\caption{Smallest tested promotion count $m$ at which mean normalized regret is at most 0.01, by pool and forecaster; in parentheses the regret when keeping one run. $S$: trajectory skill in the pool; wo: pairwise wrong-order rate; $^{\ddagger}$: arm with more than 5\% of answers replaced by persistence (secondary). Budgets for $m=1$ / $\lceil n/4\rceil$ / $\lceil n/2\rceil$: Marin 0.293 / 0.441 / 0.613; DCLM 0.362 / 0.447 / 0.631.}
\label{tab:budget-quality}
\resizebox{\linewidth}{!}{%
\begin{tabular}{@{}llllll@{}}
\toprule
Pool (groups/runs) & persistence & LC-PFN & \sol{} full ($S$, wo) & \opus{} full ($S$, wo) & \luna{} full ($S$, wo) \\
\midrule
PT/Marin (6/63) & $\lceil n/4\rceil$ (0.044) & $\lceil n/4\rceil$ (0.033) & \textbf{1} (0.010; 0.75, 0.16) & $>\lceil n/2\rceil$ (0.193; 0.80, 0.16) & $\lceil n/2\rceil$ (0.053; 0.72, 0.13) \\
PT/DCLM (4/22) & $\lceil n/2\rceil$ (0.065) & $\lceil n/4\rceil$ (0.034) & \textbf{1} (0.000; 0.71, 0.10) & \textbf{1} (0.006; 0.58, 0.08) & \textbf{1} (0.000; 0.71, 0.07) \\
FT/HF (3/36) & $\lceil n/4\rceil$ (0.030) & $\lceil n/4\rceil$ (0.030) & $\lceil n/2\rceil$ (0.068; 0.60, 0.24) & $\lceil n/4\rceil$ (0.084; $-0.11$, 0.19) & $\lceil n/2\rceil$ (0.079; 0.44, 0.18) \\
FT/Marin (10/100) & \textbf{1} (0.000) & 1 (0.000) & $\lceil n/4\rceil$ (0.032; 0.72, 0.06) & $\lceil n/2\rceil$ (0.070; 0.79, 0.18) & $\lceil n/4\rceil$ (0.104; 0.61, 0.12) \\
RL release (6/50) & \textbf{1} (0.002) & 1 (0.009) & $\lceil n/4\rceil$ (0.021; 0.17, 0.06) & 1$^{\ddagger}$ (0.001; 0.31, 0.07) & 1 (0.002; 0.27, 0.06) \\
RL supplement (3/37) & $\lceil n/2\rceil$ (0.302) & $\lceil n/2\rceil$ (0.341) & $\lceil n/2\rceil$ (0.369; 0.47, 0.28) & $\lceil n/4\rceil^{\ddagger}$ (0.323; 0.55, 0.18) & $>\lceil n/2\rceil$ (0.247; 0.24, 0.18) \\
\bottomrule
\end{tabular}}
\end{table}

\subsection{Controlled sweep}
\textbf{Design.} 72 runs of a 124M-parameter modded-nanogpt model (12 layers, width 768) on the FineWeb 10B sample, 500 optimizer steps of 131{,}072 tokens, Muon for the hidden matrices and Adam for embeddings, head and scalars; validation cross-entropy on a fixed 2{,}097{,}152-token set logged every 20 steps.
Grid: Muon learning rate $\times\{1,2,4\}$ $\times$ cooldown fraction $\{0.2,0.6\}$ $\times$ training code $\{$unchanged, embeddings frozen from step 150$\}$, in six seed blocks (the 12 runs of a block share the configured initialization and data-order seed).
A 12-run pilot fixed the grid and the analysis rules before any test run finished: the code change costs 0.097 nats at the end of the pilot runs ($16\times$ the seed standard deviation of 0.006) and 0.002 nats at the cutoff.
Runs were trained on a shared cluster with mixed GPU types; same-seed reruns on different nodes differ by 0.01--0.02 nats mid-training.
Forecasters: persistence, an AIC ensemble of parametric fits, LC-PFN, \sol{} and \opus{} under five inputs (curve; C = curve + configuration; C+D = plus a description stating the run's code change; C+K = plus the executed training script; C+D+K), with a placebo for the code change (a described but behavior-neutral edit: the embedding group's weight decay set to zero from step 150, its default value), and a transfer rule fitted on the 12 labeled pilot runs.
Rules fixed in advance: a forecaster counts as improving the decision only if it beats persistence in all six blocks; missing answers are replaced by persistence and arms with more than 5\% missing are flagged; the code change is scored on its predicted effect size against the measured one and against the placebo.

\textbf{Readings (20\% cutoff; Table~\ref{tab:mock-full}).} Skill $S$ is 0.86--0.90 for every LLM condition, yet no forecast-based promotion beats persistence in all six blocks, and when three runs are kept, every LLM condition has regret at least as high as persistence (0.004).
Keeping one run, \sol{} with C+D has mean regret 0.027 nats (per block 0.023, 0.024, 0.021, 0.036, 0.026, 0.034) against 0.060 for persistence (0.097, 0.024, 0.121, 0.020, 0.000, 0.096); the rule that excludes the runs described as freezing the embeddings and otherwise follows persistence has 0.015.
Direction accuracy on the 24 single-change pairs per block: learning-rate pairs 0.89--0.96 for the LLMs and 1.00 for persistence; cooldown pairs 0.29--0.44 for \sol{} (it predicts the longer cooldown to hurt, $+0.09$ nats, where the measured mean effect is $-0.011$) and mostly the right sign but 3--7$\times$ overstated for \opus{}; the code change (measured mean $+0.116$ nats, positive in every pair) is predicted as $+0.001$ from the configuration, $-0.059$ from the executed script and $+0.098$ from the description by \sol{} (pairwise effect MAE 0.136), with the placebo shift $-0.035$ $[-0.087, +0.017]$; \opus{} with C+D predicts $+0.042$ on answered pairs (6\% fallback, so secondary).
A defer rule fixed in advance (continue all runs to 50\% if the predicted gap between the 3rd and 4th run is below 0.006 nats) fires in every block for persistence and removes all regret at 0.625 of the full compute instead of 0.40; for the eligible LLM conditions it fires in 0--2 of 6 blocks.
API cost of the sweep's forecasts: \$111.

\begin{table}[t]
\centering\scriptsize\setlength{\tabcolsep}{3pt}
\caption{Controlled sweep, 20\% cutoff: skill $S$, mean regret in nats when one or three runs are kept (W/T/L against persistence over the six blocks), pairwise wrong-order rate, and fraction of answers replaced by persistence. Arms with more than 5\% replacement ($^{\ddagger}$) are secondary and excluded from the claims.}
\label{tab:mock-full}
\resizebox{\linewidth}{!}{%
\begin{tabular}{@{}llrrrrrrr@{}}
\toprule
Method & input & $S$ & regret $m{=}1$ & W/T/L & regret $m{=}3$ & W/T/L & wrong-order & replaced \\
\midrule
persistence & curve & 0 & 0.060 & -- & 0.004 & -- & 0.253 & 0 \\
AIC ensemble & curve & 0.846 & 0.050 & 3/1/2 & 0.004 & 1/4/1 & 0.225 & 0 \\
LC-PFN & curve & 0.477 & 0.044 & 1/4/1 & 0.004 & 1/4/1 & 0.236 & 0 \\
\sol{} & curve & 0.866 & 0.096 & 2/1/3 & 0.014 & 0/3/3 & 0.275 & 0 \\
\sol{} & C & 0.875 & 0.082 & 1/2/3 & 0.025 & 0/1/5 & 0.276 & 0 \\
\sol{} & C+D & 0.861 & 0.027 & 3/1/2 & 0.015 & 1/1/4 & 0.214 & 0 \\
\sol{} & C+K & 0.895 & 0.057 & 2/2/2 & 0.027 & 0/2/4 & 0.288 & 0.01 \\
\sol{} & C+D+K & 0.881 & 0.065 & 3/1/2 & 0.011 & 0/4/2 & 0.235 & 0 \\
\opus{} & curve & 0.873 & 0.092 & 1/2/3 & 0.024 & 0/4/2 & 0.261 & 0.01 \\
\opus{} & C & 0.887 & 0.050 & 3/0/3 & 0.008 & 1/3/2 & 0.261 & 0.03 \\
\opus{} & C+D$^{\ddagger}$ & 0.884 & 0.035 & 3/2/1 & 0.008 & 1/3/2 & 0.231 & 0.06 \\
\opus{} & C+K$^{\ddagger}$ & 0.882 & 0.037 & 4/1/1 & 0.005 & 0/5/1 & 0.299 & 0.11 \\
\opus{} & C+D+K$^{\ddagger}$ & 0.879 & 0.066 & 2/3/1 & 0.039 & 0/2/4 & 0.360 & 0.19 \\
pilot transfer & C & 0.957 & 0.051 & 2/2/2 & 0.000 & 1/5/0 & 0.199 & 0 \\
pilot transfer + code & C+K & 0.973 & 0.004 & 4/1/1 & 0.000 & 1/5/0 & 0.042 & 0 \\
exclude described change, then persistence & C+D & -- & 0.015 & -- & 0.000 & -- & -- & 0 \\
\bottomrule
\end{tabular}}
\end{table}

\subsection{Direct pairwise comparison}
\textbf{Design.} For the same 72 runs, the 24 single-change pairs of each block (12 learning-rate, 6 cooldown, 6 code pairs; 144 in all) are posed as one question: given both runs' prefixes to 20\% (and, by condition, both configurations, both descriptions, or both executed scripts), return the probability that run A ends with the lower validation loss.
Every pair is asked in both presentation orders and the two probabilities are averaged; the models are told that the two runs share the configured seed, architecture, data, step budget and validation set.
Accuracy is scored on pairs whose final losses differ by more than 0.006 nats (60 learning-rate, 33 cooldown, 36 code pairs), ties count 0.5, and the primary test, fixed in advance, is \sol{} C+D against persistence on the equal-weight average of the cooldown and code block accuracies (six-block sign test).
The comparison with ordering derived from per-run forecasts uses the same model and condition, the median forecast at $x=1$, and only pairs for which both per-run forecasts parsed strictly (common-valid).
API cost \$49.

\begin{table}[t]
\centering\scriptsize\setlength{\tabcolsep}{3pt}
\caption{Direct pairwise comparison at 20\%: accuracy by pair type, the non-learning-rate block summary with its block-bootstrap interval, the same summary from forecast-derived ordering on common-valid pairs (in parentheses the number of pairs with two strictly valid forecasts before the effect-size filter; the direct columns use all pairs), order agreement between the two presentations, first-position bias, and blocks won/lost against persistence. C+K was asked only for code pairs.}
\label{tab:pairwise}
\resizebox{\linewidth}{!}{%
\begin{tabular}{@{}llrrrrrrrr@{}}
\toprule
Model & input & LR & cooldown & code & non-LR [CI] & forecast-derived (common-valid) & order agree & position bias & W/L vs persistence \\
\midrule
persistence & curve & 1.000 & 0.455 & 0.458 & 0.451 [0.35, 0.55] & -- & -- & -- & -- \\
LC-PFN & curve & 1.000 & 0.485 & 0.500 & 0.489 [0.36, 0.61] & -- & -- & -- & 2/2 \\
\sol{} & curve & 1.000 & 0.455 & 0.444 & 0.444 [0.33, 0.55] & 0.442 (144) & 0.97 & $-0.01$ & 1/2 \\
\sol{} & C & 1.000 & 0.242 & 0.431 & 0.337 [0.24, 0.44] & 0.411 (144) & 0.90 & $0.00$ & 0/5 \\
\sol{} & C+D & 1.000 & 0.455 & 1.000 & 0.722 [0.66, 0.81] & 0.543 (144) & 0.83 & $+0.01$ & 6/0 ($p=0.031$) \\
\sol{} & C+K & -- & -- & 1.000 & -- & 0.414 (code only) & 0.72 & $+0.13$ & -- \\
\opus{} & curve & 1.000 & 0.455 & 0.444 & 0.444 [0.35, 0.54] & 0.500 (62) & 0.93 & $-0.01$ & 1/2 \\
\opus{} & C & 1.000 & 0.773 & 0.444 & 0.610 [0.49, 0.72] & 0.639 (59) & 0.90 & $-0.01$ & 5/1 \\
\opus{} & C+D & 1.000 & 0.727 & 1.000 & 0.867 [0.84, 0.89] & 0.621 (78) & 0.95 & $-0.01$ & 6/0 ($p=0.031$) \\
\opus{} & C+K & -- & -- & 1.000 & -- & 0.571 (code only) & 1.00 & $+0.03$ & -- \\
\bottomrule
\end{tabular}}
\end{table}

\textbf{Readings (Table~\ref{tab:pairwise}).} Learning-rate pairs, visible in the prefix, are ordered correctly by every direct arm and by persistence.
With C+D both models beat persistence in all six blocks on the non-learning-rate summary ($p=0.031$); \sol's gain is confined to the code pairs, which the rule that excludes the described change also solves (Appendix~\ref{app:mock2} repeats the test with changes of both signs), whereas \opus{} orders 73--77\% of the cooldown pairs correctly with or without the description.
On scored common-valid non-LR pairs, the direct C+D summary exceeds the forecast-derived one (\sol{} 0.722 against 0.543 on 69 pairs; \opus{} 0.935 against 0.621 on 39 pairs); before the effect-size filter both forecasts are strictly valid for 144/144 \sol{} pairs and 78/144 \opus{} pairs. With C alone the forecast-derived order is ahead (\sol{} C direct 0.337 against 0.411 derived). On the cooldown pairs alone, \opus{} asked directly is right on 0.923 against 0.423 for the order derived from its own forecasts with C (13 common-valid pairs) and on 0.824 against 0.471 with C+D (17 pairs).
The results are stable when the threshold on the difference in final loss is 0, 0.003, or 0.012 nats instead of 0.006; absolute first-position bias is below 0.014 for the curve, C and C+D arms.

\subsection{Direct comparison on historical sweeps}
\label{app:decision-panel}
\textbf{Design.} The pairwise question is repeated on the 24 pretraining and fine-tuning groups of the historical replay (225 runs: PT/Marin 6 groups / 63 runs, PT/DCLM 4 / 22, PT/AI2 1 / 4, FT/HF 3 / 36, FT/Marin 10 / 100) at cutoffs of 6\%, 10\% and 20\% of the budget.
At each cutoff the leader of a group is the run with the lowest last observed loss, and every other run of the group is paired with it: 201 leader--challenger pairs per cutoff, 603 in all.
Pairs are not independent runs, and the unit of inference is the group.
Each pair is asked in both presentation orders, with the two prefixes alone (curve) or with the two prefixes and both contexts (full), and the two probabilities are averaged.
The selection leaves the leader only if some challenger is given a probability above 0.5 of ending lower, and then takes the challenger with the highest probability; a missing or invalid answer counts as 0.5.
The grouping, the selection rule, the statistic and the test were fixed before any answer was read: per group, the normalized regret when one run is kept under the direct comparison with full context minus that of keeping the leader, averaged over the three cutoffs, with an exact two-sided sign test over the 24 groups and Holm correction over the two models.
Keeping the leader already selects the final best run in 51 of the 72 group--cutoff cells; 9 of the 24 groups leave headroom at some cutoff (21 cells), and in 39 of the 603 pairs the challenger ends lower than the leader by more than 5\% of the group's range (overtakes).
The forecast-derived ordering uses \sol's per-run forecasts with full context at the same cutoffs.
The RL groups were not run, and \opus's per-run forecasts were not run at 6\% and 10\%.
API cost \$90.

\begin{table}[t]
\centering\scriptsize\setlength{\tabcolsep}{3pt}
\caption{Direct comparison on 24 historical groups (225 runs): mean normalized regret when one run is kept, by cutoff (wins/ties/losses against keeping the leader over the 24 groups); switches in the 21 group--cutoff cells with headroom (to a better run / to a worse run) and in the 51 cells where the leader is already the final best (wrong switches); overtakes recalled among the 39 that exceed tolerance. Curve: prefixes only; full:~prefixes~and~contexts.}
\label{tab:decision-panel}
\resizebox{\linewidth}{!}{%
\begin{tabular}{@{}llrrrrrrr@{}}
\toprule
 & & \multicolumn{3}{c}{normalized regret, one run kept} & \multicolumn{2}{c}{headroom (21 cells)} & no headroom (51) & overtakes \\
\cmidrule(lr){3-5}\cmidrule(lr){6-7}
Method & input & 6\% & 10\% & 20\% & to better & to worse & wrong switches & recalled (39) \\
\midrule
keep the leader & curve & 0.074 & 0.048 & 0.028 & -- & -- & -- & -- \\
LC-PFN & curve & 0.049 (5/17/2) & 0.029 (3/20/1) & 0.020 (2/18/4) & 10 & 2 & 5 & -- \\
\sol{}, direct & curve & 0.081 (1/22/1) & 0.033 (2/22/0) & 0.018 (1/22/1) & 4 & 2 & 0 & 5 \\
\opus{}, direct & curve & 0.072 (1/23/0) & 0.033 (2/22/0) & 0.018 (1/22/1) & 4 & 1 & 0 & 4 \\
\sol{}, direct & full & 0.048 (3/19/2) & 0.038 (2/21/1) & 0.013 (2/21/1) & 7 & 0 & 4 & 10 \\
\opus{}, direct & full & 0.042 (4/18/2) & 0.025 (3/20/1) & 0.015 (3/20/1) & 10 & 0 & 4 & 16 \\
\sol{}, forecast-derived & full & 0.145 (4/8/12) & 0.107 (3/13/8) & 0.027 (3/17/4) & 10 & 4 & 20 & -- \\
\bottomrule
\end{tabular}}
\end{table}

\textbf{Readings (Table~\ref{tab:decision-panel}).} Averaged over the three cutoffs, the normalized regret is 0.050 for keeping the leader, 0.033 for the direct comparison with full context by \sol{} and 0.027 by \opus{}: mean difference $-0.017$ $[-0.038, +0.001]$ and $-0.023$ $[-0.048, 0.000]$, wins/ties/losses over groups 5/17/2 and 7/15/2, sign test $p=0.45$ and $p=0.18$ ($0.45$ and $0.36$ after Holm correction).
The difference is not significant for either model.
The comparison recalls 10 (\sol) and 16 (\opus) of the 39 overtakes; in the 21 cells with headroom it moves to a better run in 7 and 10 and to a worse run in none, and it switches wrongly in 4 of the 51 cells where the leader is already the final best.
The two losses of each model are the same two groups, in which the leader is the final best at every cutoff.
The ordering derived from \sol's own forecasts moves to a better run in 10 of the 21 cells and to a worse run in 4, and switches wrongly in 20 of the 51 (14 of the 30 FT/Marin cells, where no direct arm switches); its regret is above that of keeping the leader at 6\% and 10\%.
LC-PFN is close to the direct comparison by \opus{} (10 to a better run, 2 to a worse run, 5 wrong switches).
With the prefixes alone the direct comparison recalls 5 and 4 of the 39 overtakes, switches in none of the 51 cells without headroom and moves to a worse run in 2 and 1 of the 21 with headroom; the group-level difference between full context and curve only is not significant (\sol{} $-0.011$ $[-0.027, +0.003]$, 6/16/2; \opus{} $-0.014$ $[-0.035, +0.007]$, 5/17/2).
The comparison does not help measurably more at the earlier cutoffs (difference between its margin over the leader at 6\% and at 20\%: $-0.011$ $[-0.054, +0.034]$ for \sol{} and $-0.018$ $[-0.063, +0.024]$ for \opus).
The two presentation orders agree on 97--98\% of pairs, and at most 1.8\% of an arm's answers were missing or invalid.

\subsection{From skill to saving: a simulation}
\label{app:skill-saving}
\textbf{Design.} To map a skill level to a saving, we replace the forecaster in the historical replay by synthetic forecasters of controlled quality, on the same groups and with the same policy.
A \emph{shrink} forecaster removes a fraction $F$ of the last value's error on every run, $\hat y_i = p_i + F\,(y_i - p_i)$, so that its final-value skill is exactly $F$ and its errors keep the structure of the last value's.
A \emph{noise} forecaster predicts $\hat y_i = y_i + e_i$ with $e_i = s\,(\sqrt{\rho}\,c_g + \sqrt{1-\rho}\,\epsilon_i)$, where $c_g$ is drawn once per group (an error shared by the whole sweep), $\epsilon_i$ once per run, and $s$ sets the skill; $\rho\in\{0, 0.5, 0.9\}$, 500 draws per setting.
For each forecaster we find the lowest skill from which one kept run per sweep reaches mean normalized regret $\le 0.01$, and the lowest skill from which it needs no more budget than the last value.
No~model~is~called.

\begin{table}[t]
\centering\small
\caption{Skill needed to save compute, by pool. \emph{Last value needs}: the smallest tested promotion count at which the last value reaches mean normalized regret $\le 0.01$, and its budget. \emph{One run from $F$}: the lowest final-value skill at which a synthetic forecaster reaches the target keeping one run (shrink), or keeping one run / needing no more budget than the last value (noise, by shared-error fraction $\rho$). ``Never'': not reached at any tested skill.}
\label{tab:skill-saving}
\resizebox{\linewidth}{!}{%
\begin{tabular}{@{}lllccccc@{}}
\toprule
 & & & \multicolumn{2}{c}{shrink} & \multicolumn{3}{c}{noise: one run / no loss from $F$} \\
\cmidrule(lr){4-5}\cmidrule(lr){6-8}
Pool (groups/runs) & last value needs & budget, one run & one run from $F$ & saving & $\rho=0$ & $\rho=0.5$ & $\rho=0.9$ \\
\midrule
PT/Marin (6/63) & $\lceil n/4\rceil$, 0.441 & 0.293 & 0.6 & 0.148 & 0.95 / 0.80 & 0.95 / 0.70 & 0.85 / 0.29 \\
PT/DCLM (4/22) & $\lceil n/2\rceil$, 0.631 & 0.362 & 0.3 & 0.269 & 0.95 / 0.80 & 0.95 / 0.70 & 0.85 / 0.29 \\
PT/AI2 (1/4) & $\lceil n/2\rceil$, 0.600 & 0.400 & 0.8 & 0.200 & never & never & never \\
FT/HF (3/36) & $\lceil n/4\rceil$, 0.418 & 0.273 & 0.5 & 0.145 & 0.95 / 0.80 & 0.90 / 0.71 & 0.80 / 0.29 \\
FT/Marin (10/100) & 1, 0.280 & 0.280 & -- & 0 & never & never & 0.95 / 0.95 \\
RL release (6/50) & 1, 0.311 & 0.311 & -- & 0 & 0.50 / 0.50 & 0.21 / 0.21 & $\le 0$ / $\le 0$ \\
RL supplement (3/37) & $\lceil n/2\rceil$, 0.627 & 0.308 & 0.6 & 0.319 & 0.85 / 0.40 & 0.81 / 0.13 & 0.46 / $\le 0$ \\
\bottomrule
\end{tabular}}
\end{table}

\textbf{Readings (Table~\ref{tab:skill-saving}).} A skill level does not fix a saving; the sweep and the structure of the error do.
With errors that keep the structure of the last value's, a skill of 0.6 is enough to keep one run on the Marin pretraining sweeps (saving 0.148 of the total budget, a third of what the last value needs) and 0.3 on the DCLM sweeps.
With errors that are independent across the runs of a sweep, one kept run needs a skill of about 0.95, and below about 0.8 such a forecaster needs more budget than the last value: most of the last value's error is shared within a sweep, which costs skill but does not change the order of the runs.
In the Marin fine-tuning sweeps and the RL release the last value already keeps the eventual best run, so no skill level saves anything.
The real forecasters mostly behave like independent-error forecasters: at their own final-value skill, 118 of 137 forecaster--pool cells lie on the $\rho=0$ curve when one run is kept.
Across the LLM forecasters, the pooled rank correlation between the skill score $S$ and lower regret is 0.23 $[0.08, 0.49]$ at $\lceil n/4\rceil$ and 0.31 $[0.05, 0.35]$ keeping one run (bootstrap over groups; 0.14 $[0.04, 0.36]$ and 0.21 $[0.03, 0.29]$ when the last value and the curve fitters are included).
A benchmark of decisions therefore has to report the order of the runs next to the skill score.

\subsection{A second controlled sweep: described changes of both signs}
\label{app:mock2}
\textbf{Design.} In the controlled sweep the one described code change always raises the final loss, so a rule that predicts the worse outcome for any described change also orders the code pairs correctly.
A second sweep of 72 runs of the same model on the same data removes this confound: Muon learning rate $\times\{1,2,4\}$ $\times$ four arms, in six seed blocks.
The arms are \emph{unchanged}; \emph{freeze} (embeddings frozen from step 150, as before); \emph{frozen base} (embeddings frozen from step 50, before the cutoff at step 100); and \emph{unfreeze} (frozen from step 50 and released at step 150).
The two primary pair types share the curve up to the cutoff and differ only after it: freeze against unchanged, where the described change hurts, and unfreeze against frozen base, where it helps (18 pairs each; all 36 differ by more than twice the seed standard deviation of the pilot, $\sigma=0.0087$ nats).
The candidate changes, the selection rule and the primary test were fixed before the test runs. A first pilot of eight candidate changes produced none that lowers the final loss by the pre-set margin, as expected for a tuned configuration, so the helping change was constructed as the removal of a handicap shared by both runs of a pair, in an amendment written before the second pilot.
Questions, conditions and scoring are those of the direct pairwise task above: curve only, C, and C+D, and in addition C+D with a placebo sentence on the unchanged run, so that both runs of a pair carry a described change.
The comparators are persistence and the two single-direction rules (a described change hurts; a described change helps), each of which is right on one pair type and wrong on the other.

\begin{table}[t]
\centering\small
\caption{Second controlled sweep, direct pairwise comparison at 20\%: accuracy by pair type (18 pairs each; the primary set is the first two columns pooled, 36 pairs; ties count 0.5). In freeze against unfreeze the run that is behind at the cutoff ends ahead in all 18 pairs. C+D$^{p}$ adds a placebo sentence to the unchanged run; it and the two single-direction rules are not defined for freeze against unfreeze, where both runs carry a described change. W/L counts the six seed blocks won and lost on the primary set against persistence.}
\label{tab:mock2}
\resizebox{\linewidth}{!}{%
\begin{tabular}{@{}llrrrrr@{}}
\toprule
Method & input & \makecell[r]{freeze vs.\\unchanged} & \makecell[r]{unfreeze vs.\\frozen base} & primary & \makecell[r]{freeze vs.\\unfreeze} & \makecell[r]{W/L vs.\\persistence} \\
\midrule
persistence & curve & 0.528 & 0.556 & 0.542 & 0.000 & -- \\
rule: a described change hurts & description & 1.000 & 0.000 & 0.500 & -- & -- \\
rule: a described change helps & description & 0.000 & 1.000 & 0.500 & -- & -- \\
\midrule
\sol{} & curve & 0.528 & 0.500 & 0.514 & 0.000 & -- \\
\sol{} & C & 0.500 & 0.444 & 0.472 & 0.000 & -- \\
\sol{} & C+D & 1.000 & 1.000 & 1.000 & 1.000 & 6/0 ($p=0.031$) \\
\sol{} & C+D$^{p}$ & 1.000 & 0.944 & 0.972 & -- & -- \\
\opus{} & curve & 0.583 & 0.500 & 0.542 & 0.000 & -- \\
\opus{} & C & 0.556 & 0.556 & 0.556 & 0.000 & -- \\
\opus{} & C+D & 1.000 & 0.944 & 0.972 & 1.000 & -- \\
\opus{} & C+D$^{p}$ & 1.000 & 1.000 & 1.000 & -- & -- \\
\bottomrule
\end{tabular}}
\end{table}

\textbf{Readings.} Table~\ref{tab:mock2} gives the accuracy of every arm. Freezing raises the final loss by 0.097, 0.098 and 0.156 nats at the three learning rates, and unfreezing lowers it by 0.152, 0.155 and 0.214 nats; the two runs of a primary pair differ by at most $1.5\sigma$ at the cutoff.
With C+D, \sol{} orders 36 of 36 primary pairs correctly and \opus{} 35 of 36, against 0.50 for either single-direction rule and 0.54 for persistence; the test fixed in advance (\sol{} C+D above both comparators in all six blocks) is met ($p=0.031$ for each).
Without the descriptions both models are at chance (curve only: \sol{} 0.51, \opus{} 0.54; C: 0.47 and 0.56), and with the placebo sentence they remain at 35 and 36 of 36, so the answers depend on what the description says.
On the 18 pairs of freeze against unfreeze, the run that is behind at the cutoff ends ahead: both models choose the wrong run in all 18 pairs from the curve or the configuration, and the right run in all 18 with the descriptions.
The test covers one mechanism in two directions, with large effects, on one model: the models read the direction of a described change, and ranking small or unfamiliar changes is untested.

\section{Additional analyses}
\label{app:additional}

\subsection{Later runs from new owners}
\label{app:temporal}
\textbf{Construction.} From a later Weights~\&~Biases survey we took every lineage whose first run was created after 26 September 2026, when the test sets, the reader, and the prompts were already fixed, and whose owner appears in none of our test sets and not in the reader's 918-run training pool, capped at two lineages per family and 50 per owner unit; a run enters if it has at least 20 logged points, a value in each of the ten prefix windows and an observed end point, and if its prefix is not constant.
The panel has 129 runs (110 pretraining, 19 RL) from 6 projects, 5 owner units and 96 lineage families; about half come from two course projects, and 85\% are smooth pretraining losses, so it tests later runs from new owners rather than harder data.
Requests use the community-panel format verbatim (two grids, curve-only and full-context conditions); the reader adapters are those of the 2B reader of \S\ref{sec:training} trained on the curve and configuration of the 918-run pool (three seeds, not re-selected), which answer both requests, so the curve-only request omits the configuration they were trained with and the full request adds a context they were not trained with, run through the unchanged generation path, and \sol{} and \opus{} are the model snapshots of the main comparison, whose training data cannot contain runs created after that date.
Intervals are family-clustered bootstraps.

\begin{table}[t]
\centering\footnotesize\setlength{\tabcolsep}{4pt}
\caption{Top: skill $S$ on the 129 later runs (258 views) with family-clustered 95\% intervals. Bottom: paired error ratios against three baselines (geometric mean; $<1$ favors the row; for the reader, the range over its three seeds). The community column repeats the scores on all 232 community runs (Appendix~\ref{app:wild-details}) for reference.}
\label{tab:temporal}
\resizebox{\ifdim\width>\linewidth\linewidth\else\width\fi}{!}{%
\begin{tabular}{@{}llcccc@{}}
\toprule
Method & input & all & PT (110) & RL (19) & community (232) \\
\midrule
\opus{} & full & 0.742 [0.693, 0.782] & 0.783 & 0.283 & 0.31 \\
\sol{} & full & 0.736 [0.690, 0.772] & 0.770 & 0.414 & 0.31 \\
\sol{} & curve & 0.755 [0.711, 0.792] & 0.794 & 0.332 & -- \\
\opus{} & curve & 0.727 [0.677, 0.767] & 0.765 & 0.361 & -- \\
\reader{} 2B (3 seeds) & full & 0.724 (0.719 / 0.730 / 0.723) & 0.788 & $-0.27$ & 0.27 \\
\reader{} 2B (3 seeds) & curve & 0.743 & 0.799 & $-0.06$ & -- \\
ExtraTrees (train split) & curve & 0.764 [0.708, 0.807] & 0.831 & $-0.66$ & $-0.24$ \\
LC-PFN & curve & 0.644 [0.582, 0.695] & 0.719 & $-0.41$ & 0.09 \\
AIC ensemble & curve & 0.589 [0.499, 0.659] & 0.643 & 0.08 & 0.05 \\
hindsight smooth fit (not a forecaster) & full curve & 0.902 & 0.910 & 0.834 & -- \\
\bottomrule
\end{tabular}}

\vspace{4pt}
\begin{tabular}{@{}llccc@{}}
\toprule
Paired error ratio & input & vs LC-PFN & vs AIC ensemble & vs ExtraTrees \\
\midrule
\opus{} & full & 0.73 [0.62, 0.86] & 0.63 [0.54, 0.73] & 1.10 [0.94, 1.29] \\
\sol{} & full & 0.74 [0.64, 0.86] & 0.64 [0.55, 0.74] & 1.12 [0.92, 1.34] \\
\reader{} 2B (3 seeds) & full & 0.76--0.79 & 0.66--0.68 & 1.14--1.19 \\
\reader{} 2B (3 seeds) & curve & 0.68--0.78 & 0.59--0.68 & 1.03--1.18 \\
\bottomrule
\end{tabular}
\end{table}

\textbf{Readings.} The frozen \sol{}, \opus{} and reader beat LC-PFN and the AIC ensemble on later runs from new owners, with all intervals below one; the full context changes the error over the curve alone by 1.07 [0.96, 1.20] (\sol{}) and 0.95 [0.86, 1.03] (\opus{}), neither significant; a curve-prefix ExtraTrees trained on the benchmark's training split matches the LLMs and beats the reader; on the 19 RL runs the LLMs have the highest skill (0.28--0.41, point estimates), the AIC ensemble is at 0.08 and the other forecasters in Table~\ref{tab:temporal} are below zero. API cost \$39.

\subsection{Matched baselines for retrieval}
\label{app:matched}
\textbf{Same-neighbor curve transfer.} For each of the 125 retrieval questions we recovered the three neighbor curves shown to \sol{} from the dispatched requests and transferred each neighbor's post-cutoff change onto the target's last observed value (difference or ratio, combined by mean, median or similarity weighting; the variant was chosen by leave-one-out on the training library before any test score was read, giving difference--mean).
On the 123 questions valid for both \sol{} arms it reaches $S=0.597$ [0.508, 0.670] against 0.799 [0.757, 0.833] for \sol{} with the same neighbors and 0.670 without (paired error ratio 2.00 [1.61, 2.48]); on pretraining it scores 0.722 against \sol's 0.666 without and 0.850 with retrieval, on fine-tuning 0.297 against 0.677 and 0.686.
An ExtraTrees on prefix and configuration trained on the library with the question's sweep family held out reaches 0.81 on pretraining and 0.10 on fine-tuning.

\textbf{Configuration-conditioned FT-PFN with the same neighbors.} The FT-PFN baseline of the main comparison saw one context curve and mostly neutral configuration slots.
We ran the frozen author checkpoint with four curves per question (the target's prefix and the three neighbors' complete curves) and a ten-slot configuration vector per source with explicit field paths (learning rate, weight decay, global batch, sequence length, width, depth, heads, and three source-specific fields), fixed monotone transforms and a within-question map to $[0.1, 0.9]$ (missing fields recorded and set to 0); losses are normalized per question from the four curves ($c$ = median, $s=\max(\mathrm{IQR}/1.349, 10^{-3}\max(|c|,1))$, $y=\sigma((c-\ell)/s)$) and inverted with the same constants; curves carry explicit run identities, since the stock tokenizer merges curves with equal configuration vectors (it would have done so in 107 of 125 questions).
A second arm gives the model the target prefix only, with the same encoding and constants.
Results on the 123 common-valid questions (Table~\ref{tab:ftpfn}): FT-PFN with the neighbors reaches $S=0.820$ [0.794, 0.843] against 0.799 for \sol{} with retrieval; the paired error ratio is 0.90 [0.70, 1.17] overall, 1.48 [1.17, 1.87] on pretraining and 0.43 [0.32, 0.62] on fine-tuning, and 0.48 [0.40, 0.57] against the prefix-only arm.
The neighbors are identical; \sol{} also reads the description and code.

\begin{table}[t]
\centering\small\setlength{\tabcolsep}{4pt}
\caption{Retrieval baselines on the 123 common-valid questions: skill $S$ overall and by stage, and paired error ratios against \sol{} with retrieval (geometric mean, lineage-component~bootstrap).}
\label{tab:ftpfn}
\resizebox{\linewidth}{!}{%
\begin{tabular}{@{}lcccccc@{}}
\toprule
Method & $S$ all & PT & FT & ratio vs \sol{}+retrieval: all & PT & FT \\
\midrule
FT-PFN, three neighbors + configuration & 0.820 & 0.780 & 0.867 & 0.90 [0.70, 1.17] & 1.48 [1.17, 1.87] & 0.43 [0.32, 0.62] \\
FT-PFN, target prefix only & 0.623 & 0.625 & 0.620 & 1.89 [1.52, 2.39] & 2.53 [1.88, 3.46] & 1.21 [0.94, 1.59] \\
same-neighbor curve transfer & 0.597 & 0.722 & 0.297 & 2.00 [1.61, 2.48] & -- & -- \\
\sol{}, no retrieval & 0.670 & 0.666 & 0.677 & -- & -- & -- \\
\sol{}, three neighbors & 0.799 & 0.850 & 0.686 & 1 & 1 & 1 \\
\bottomrule
\end{tabular}}
\end{table}

\subsection{Description and code with curve and configuration fixed}
\label{app:factorial}
On 200 \textsc{Grid20} test cases (100 pretraining, 100 fine-tuning, chosen by hash within stage), we derived three conditions from the full-context request by keeping the curve-only prompt and the numeric configuration (the existing C condition) and adding the task description with its source facts (C+D), the code excerpt (C+K) or both (C+D+K); the interpretation rule and known-unknown fields of the full prompt are not added, so the full condition differs from C+D+K by those fields and by the string-valued configuration entries.
\sol{} answered all 600 new requests; \opus{} 328 strictly, 261 through the paper's lenient parser, 11 not at all. API cost \$37.

\begin{table}[t]
\centering\small\setlength{\tabcolsep}{4pt}
\caption{Real-log factorial: skill $S$ per condition and paired error ratios for adding the description (D) or the code (K) with curve and numeric configuration fixed ($<1$ = lower error with the component; lineage-component~bootstrap).}
\label{tab:factorial}
\resizebox{\linewidth}{!}{%
\begin{tabular}{@{}lccccccccc@{}}
\toprule
Model & C & C+D & C+K & C+D+K & D: C+D/C & D: C+D+K/C+K & K: C+K/C & K: C+D+K/C+D & C+D+K/C \\
\midrule
\sol{} & 0.664 & 0.666 & 0.678 & 0.644 & 0.96 [0.89, 1.05] & 1.02 [0.93, 1.14] & 1.00 [0.91, 1.10] & 1.07 [0.97, 1.17] & 1.03 [0.94, 1.12] \\
\opus{} & 0.703 & 0.675 & 0.672 & 0.690 & 1.01 [0.91, 1.12] & 0.99 [0.90, 1.09] & 1.00 [0.89, 1.12] & 0.94 [0.87, 1.03] & 0.97 [0.89, 1.07] \\
\bottomrule
\end{tabular}}
\end{table}

Table~\ref{tab:factorial} gives the skill in each condition and the paired ratios.
Neither component gives a consistent gain for either model, overall or within a stage (the most extreme stage ratio is 0.87 [0.76, 1.01], \opus{} C+D+K/C+D on pretraining).
Real descriptions are largely templates (6 distinct texts for 150 pretraining runs), and code excerpts are project-level examples rather than the executed revision.
In the controlled sweep of Appendix~\ref{app:decisions}, where the description states the run's own code change, the description does help.

\subsection{Data composition and missingness}
\label{app:census}
Every run in \bench{} trains a language model; every target is a training loss (464 runs) or a reward (218).
Pretraining: 150 runs in 144 lineage components; next-token language models, dense and MoE (dense up to 412M parameters, MoE up to width 6{,}144 with 48 layers); one project (Marin) holds 67\%; run-bound training-data names are available for 16 runs.
Fine-tuning: 150 runs in 60 components; SFT of ${\approx}$21M--1.4B (estimated non-embedding) and ${\approx}$3.3B--35B (from configuration) checkpoints; Marin holds 67\%; no run binds its dataset name.
RL: 150 runs (133 scoreable) in 24 projects, the largest with 20\%; GRPO 61, PPO 45, REINFORCE 9, online DPO 5, other 8, unrecorded 22; model size identifiable for 98 (0.12B--30B); task domain known for 94 (math 30, reasoning-gym 10, code 9, agentic 6, games 3, RLHF with a learned reward model 30, other 6).
Community: 232 runs by 184 owners, the largest owner with 2.2\% and the largest project 0.4\%; SFT/preference 140 and GRPO/RLOO 92; model size identifiable for 194 (0.135B--70B); dataset named for 95.

Table~\ref{tab:data-card} classifies the 682 runs along the axes other than stage: source and its share, model family and scale, algorithm and data or task, target and budget; every row is language-model training, and the \emph{Ex.} lines give one run of each set as the forecaster sees it.
Appendix~\ref{app:axis} reports forecasting error along these axes.

\textbf{Recorded compute.} A training runtime is recorded for 271 of the 682 runs (the 100 Marin pretraining runs, 30 RL runs, and 141 community runs) and sums to 3{,}445 wall-clock hours. The 100 Marin pretraining runs also record their devices and used 4{,}076 GPU-hours and 3{,}213 TPU-chip-hours. The other 411 runs record no runtime, so these figures are lower bounds for the benchmark.

\newcommand{\exrunw}[1]{%
\arrayrulecolor{black!30}\midrule
\multicolumn{6}{@{}p{\linewidth}@{}}{\textcolor{gray}{\emph{Ex.}: #1}}}
\begin{table}[t]
\centering\scriptsize
\setlength{\tabcolsep}{3pt}
\caption{What the four test sets contain. \emph{Runs} counts runs (each forecast at two query grids); \emph{units} counts the independent units that the 95\% intervals resample: lineage components for PT and FT, runs for RL (from 24 projects), and authors for the community set. Model scale is reported where the configuration or the~model~name~states~it.}
\label{tab:data-card}
\begin{tabular}{@{}l|r|>{\raggedright\arraybackslash}p{0.15\linewidth}|>{\raggedright\arraybackslash}p{0.18\linewidth}|>{\raggedright\arraybackslash}p{0.25\linewidth}|>{\raggedright\arraybackslash}p{0.16\linewidth}@{}}
\toprule
\textbf{Set} & \textbf{Runs (units)} & \textbf{Sources (largest share)} & \textbf{Models and scale} & \textbf{Algorithm; data or task} & \textbf{Target (axis); budget} \\
\midrule
Pretraining & 150 (144) & Marin MoE, DCLM, OLMoE (67\%) & MoE and dense decoders, width 288--6{,}144, 1--48 layers & next-token prediction; C4, RefinedWeb, RedPajama where named (16 runs) & loss (steps, tokens); 0.05--130B tokens \\
\exrunw{a Marin MoE run with 128 experts, width 768, one of 91 learning rates in the set, 20\% of its planned steps observed.} \\
\arrayrulecolor{black!100}\midrule
Fine-tuning & 150 (60) & Marin, Open-R1 (67\%) & Gemstone 21M--1.4B; Qwen2-shaped 3B--35B & SFT; T\"ulu~3 mixture (Marin), reasoning SFT (Open-R1) & loss (steps, epochs); 3{,}834 steps or 3--15 epochs \\
\exrunw{T\"ulu~3 SFT of one of 10 Gemstone models from one of 10 pretraining checkpoints; the checkpoint is what varies.} \\
\arrayrulecolor{black!100}\midrule
RL & 150 (150) & 24 projects, $\sim$12 institutions (20\%) & GPT-2, Qwen2.5/3, Llama, OLMo, Gemma; 0.12--30B & GRPO 61, PPO 45, REINFORCE 9, online DPO 5, other 8, unrecorded 22; math, code, agentic, games, RLHF & reward 126, loss 24 (steps, epochs); 48--15{,}492 steps \\
\exrunw{GRPO on math prompts from a 1.5B checkpoint, 86 of 430 steps observed; \emph{will the mean reward keep rising?}} \\
\arrayrulecolor{black!100}\midrule
In the wild & 232 (184) & Hugging Face Hub, 184 authors (2\%) & Qwen, Mistral, Llama, Gemma; 0.1--70B; 82 LoRA, 24 vision-language & SFT/DPO/ORPO/KTO 140, GRPO 91, RLOO 1; dataset named for 95 runs & loss 140, reward 92 (steps); 67--149{,}068 steps \\
\exrunw{a community LoRA run whose trainer state logs 62 points, twelve of them in the prefix; the repository is the only description.} \\
\arrayrulecolor{black!100}\bottomrule
\end{tabular}
\end{table}

\subsection{Community-run diagnostics}
\label{app:community-diag}
Table~\ref{tab:community-diag} collects the evidence on why skill drops on community runs (\S\ref{sec:main}). Algorithm, author type and logging density are confounded (75 of the 92 community RL runs are by individuals), and data quality and project familiarity are not identified separately.

\begin{table}[t]
\centering\small\setlength{\tabcolsep}{4pt}
\caption{Candidate causes of the community-run drop and the evidence for each.}
\label{tab:community-diag}
\resizebox{\linewidth}{!}{%
\begin{tabular}{@{}lp{0.62\linewidth}@{}}
\toprule
Candidate cause & Evidence \\
\midrule
Logging density and noise & Thinning curated curves to community density and adding community-level noise raises the hindsight-best-fit error floor from 0.090 to 0.214, against 0.225 on community runs: 95\% of the gap in the floor on a log scale (a diagnostic of the floor, not of LLM error). \\
Algorithm & On community RL runs the LLMs have essentially no skill over persistence (error ratio 0.96--1.04), against 0.55--0.64 on the 140 SFT/preference runs; context helps only on SFT/preference runs (curve-only/full 1.31 for \sol{}, 1.22 for \luna{}; 0.96 and 1.03 on RL). \\
Documentation quality & Stripping or enriching the metadata of the same run changes error by 0.96--1.02. \\
Author type & After normalizing each run by its own hindsight-fit error, the individual/institution ratios are 0.87--1.15 with all intervals including 1. \\
Shift for trained predictors & Regressors trained on the benchmark's training split transfer negatively (ExtraTrees error ratio 1.24 [1.05, 1.48]). \\
Later runs from new owners & On the 129 later runs (Appendix~\ref{app:temporal}) the frozen LLMs keep their lead over LC-PFN (error ratio 0.73--0.74); the panel is 85\% smooth pretraining loss and is not a matched test of the community drop. \\
\bottomrule
\end{tabular}}
\end{table}

\subsection{Forecasting error by data axis}
\label{app:axis}
We regrouped the per-view error ratios already produced by the frozen scorers along axes other than training stage (metric, planned budget, owner type, model scale, and the change from the last observed value to the final target as a share of the curve's range, a post-hoc axis computed from the truth), without re-scoring any request; the table is in the supplementary document (\S S6).
Curve shape is the only axis that explains at least 5\% of the between-run variance of the run-mean log ratio in every set ($\eta^2$ for \sol{} 0.05--0.10 over the four sets, where scale reaches 0.05 only on fine-tuning), and on RL and community runs the lowest tertile, curves that barely move after the cutoff, is the hardest (\sol{}/\opus{} 0.86/0.85 on RL and 0.96/1.06 on community, against 0.43--0.64 in the other tertiles); on community reward curves that stay flat all three forecasters are worse than persistence (\sol{} 1.91 [1.33, 2.82], \opus{} 1.74, LC-PFN 1.79).
Model scale explains little and is not monotone (on community runs \sol{} reaches 0.74, 0.73 and 0.54 at 1--4B, 4--10B and $\ge$10B), community difficulty is concentrated in reward curves (\sol{} 0.55 on loss against 1.00 on reward) while the RL set shows no such gap (0.65 on reward against 0.69 on loss), and whether the training dataset is named in the context makes no difference within any set ($\eta^2\le 0.04$).

\subsection{Cleaned reruns of the reader}
\label{app:clean}
The 2{,}641- and 3{,}002-run pools of \S\ref{sec:training} contain 22 records related to test runs (Appendix~\ref{app:reader}).
We removed them (180 training examples), which gives 2{,}619 and 2{,}980 runs, and retrained the four 2B cells with the recipe, the inputs, and the three seeds unchanged (Table~\ref{tab:reader-clean}).
Skill on the frozen pretraining and fine-tuning test sets changes by at most 0.022, and the gain on held-out sources over the 918-run reference remains (0.779--0.806 against 0.680).
The 9B cells of these two pools~were~not~retrained.

\begin{table}[t]
\centering\scriptsize\setlength{\tabcolsep}{3pt}
\caption{Cleaned reruns of the 2B cells on the two middle pools after removing the 22 related records (2{,}641 $\to$ 2{,}619 and 3{,}002 $\to$ 2{,}980 runs; recipe, inputs and three seeds unchanged). Columns as in Table~\ref{tab:reader-macro}; $\Delta$ is the paired change against the original cell with its 95\% interval over three seeds and test runs. The RL test set contains none of the 22 records, and the reruns do not separate the removal from other training variation.}
\label{tab:reader-clean}
\resizebox{\linewidth}{!}{%
\begin{tabular}{@{}llrrrrrrrrll@{}}
\toprule
Pool & Model, input & PT & FT & RL & Avg. & PT+FT & \makecell{Held-out\\sources} & \makecell{Held-out\\families} & Wild & $\Delta$ frozen & $\Delta$ RL \\
\midrule
2{,}619 & 2B, config & 0.677 & 0.521 & $-0.682$ & 0.172 & 0.623 & 0.786 & 0.237 & 0.270 & $+0.009$ $[-0.001, +0.019]$ & $-0.19$ $[-0.35, -0.08]$ \\
2{,}619 & 2B, config + semantics & 0.673 & 0.560 & $-0.648$ & 0.195 & 0.632 & 0.794 & 0.215 & 0.275 & $-0.022$ $[-0.036, -0.007]$ & $+0.12$ $[-0.04, +0.31]$ \\
2{,}980 & 2B, config & 0.682 & 0.663 & $-0.611$ & 0.245 & 0.675 & 0.779 & 0.245 & 0.273 & $+0.021$ $[+0.011, +0.031]$ & $-0.36$ $[-0.57, -0.21]$ \\
2{,}980 & 2B, config + semantics & 0.696 & 0.605 & $-0.695$ & 0.202 & 0.663 & 0.806 & 0.258 & 0.263 & $-0.007$ $[-0.018, +0.003]$ & $-0.16$ $[-0.30, -0.03]$ \\
\bottomrule
\end{tabular}}
\end{table}

\subsection{Why full context lowers one model's fine-tuning skill}
\label{app:context-hurts}
On the fine-tuning test set, full context changes \opus's skill by $-0.19$ $[-0.29, -0.10]$ and \sol's by $+0.06$ $[0.02, 0.11]$ against the curve alone (Appendix~\ref{app:significance}).
We decomposed both contrasts from the existing answers; no new answer was generated.
Three quarters of \opus's change comes from the 50 Open-R1 runs, which train for several epochs: there its skill falls from 0.565 to 0.007 ($\Delta S=-0.56$ $[-0.90, -0.28]$, 47 runs), against $-0.05$ $[-0.11, -0.01]$ on the Marin runs (96 runs).
Once the prompt states the number of epochs, \opus{} predicts a median of 2.71 times the loss decline that actually occurs between the cutoff and the end of the run (0.67 times from the curve alone); its final forecast is below the truth on 43 of 47 runs, and 43 of its 47 explanations mention memorization or overfitting (2 of 47 with the curve alone).
The extra decline is predicted in intervals with and without an epoch boundary, so it is an overall expectation and not a misplaced epoch step.
On the 62 runs where both answers are strictly valid, the contrast is $-0.22$ $[-0.31, -0.12]$.
\sol{} reads the same field and adjusts by a smaller amount: with the curve alone it predicts 0.51 times the real decline on Open-R1, with full context 0.66 times, and its gain is there ($+0.13$ $[0.01, 0.22]$, 50 runs; $+0.02$ on Marin).
On one 10-epoch Open-R1 run whose loss is 0.625 at the cutoff and 0.526 at the end, \opus{} predicts 0.535 from the curve and 0.210 with full context, and \sol{} 0.589 and 0.529.
The same context thus moves both models in the same direction, and the difference is one of amount.
These are observed associations; removing the epoch field from the prompt would be the causal test.

\FloatBarrier
\section{Research-progress curves of research agents}
\label{app:rsi}

\subsection{Setting}
An autonomous research agent repeatedly edits a solution, scores it against a judge, and keeps the best.
Its progress curve is the best score reached so far over the hours, dollars, generations, or iterations of its budget: a staircase with long flat stretches and rare jumps, driven by the agent's choices rather than by an optimizer.
We collect such curves from 13 public sources of agent runs and read them under the protocol of \S\ref{sec:tasks}.
Table~\ref{tab:rsi-sources} lists the nine sources that enter the totals: the OpenRSI Index~\citep{openrsi2026index}; the trajectories of LLM-guided evolutionary search released by \citet{zhang2026optimizer} (Evo-LLM); Autoresearch Bench~\citep{autoresearchbench2026}; EdgeBench~\citep{zhu2026edgebench}; the independent agents of a public set of 12-hour coding-agent runs (Solo-vs-forum);\footnote{\url{https://huggingface.co/datasets/junlinw/autoresearch-solo-vs-forum}} EvoTrace~\citep{pelleriti2026evotrace}; the traces of the Gome agent on MLE-bench competitions~\citep{zhang2026gome,chan2025mlebench}; a public set of OpenEvolve runs~\citep{sharma2025openevolve} on KernelBench problems~\citep{ouyang2025kernelbench};\footnote{\url{https://huggingface.co/datasets/willychan21/casestudy_openevolve_results}} and AlgoTune~\citep{press2025algotune}.
Four more are scored but kept out of the totals: the trajectories released with MLGym-Bench~\citep{nathani2025mlgym}, where test scores were returned to the agent; the logs of MLAgentBench~\citep{huang2024mlagentbench}, public before the forecasting models were trained; and result logs committed to forks of autoresearch~\citep{karpathy2026autoresearch} and agent trajectories on ALE-Bench~\citep{imajuku2025alebench}, each with a single test task.
The independent unit is the task: within each source, tasks are split by a hash of their name into a development part and a test part (30\% of the tasks), runs, views, and requests are nested in tasks, and every interval is a 95\% bootstrap over tasks.
Every reading of an LLM below is on the test tasks.

\begin{table}[t]
\centering\small\setlength{\tabcolsep}{5pt}
\caption{Sources of research-progress curves that enter the totals. Tasks are the independent experiments. Axis: the budget against which a prefix is cut; for Gome the axis is the index of the loop within the finished trace. Fleet: the source also enters the task-level reading of Table~\ref{tab:rsi-fleet}. EdgeBench publishes only the running best of each run.}
\label{tab:rsi-sources}
\begin{tabular}{@{}llrrrc@{}}
\toprule
Source & What a run is & Tasks & Test tasks & Runs & Fleet \\
\midrule
OpenRSI Index & self-improvement run on an ML task; hours & 11 & 3 & 22 & \\
Evo-LLM trajectories & LLM-guided evolutionary search; generations & 8 & 2 & 238 & \checkmark \\
Autoresearch Bench & research loop of 4--12 hours; hours & 20 & 6 & 495 & \checkmark \\
EdgeBench & agent, 12 hours; hours & 134 & 40 & 1{,}912 & \checkmark \\
Solo-vs-forum & independent coding agents, 12 hours; hours & 9 & 3 & 1{,}362 & \checkmark \\
EvoTrace & evolutionary search, 100 iterations; iterations & 18 & 5 & 103 & \checkmark \\
Gome traces & ML-engineering agent; loop index & 40 & 12 & 120 & \\
OpenEvolve case study & kernel search; iterations & 18 & 5 & 18 & \\
AlgoTune & agent with a \$1 budget; dollars & 153 & 46 & 2{,}014 & \checkmark \\
\bottomrule
\end{tabular}
\end{table}

\textbf{Protocol.} A view is one run cut at 5, 10, 15, or 20\% of its budget, with at least three scored evaluations before the cut and one after it.
The forecaster sees the evaluations up to the cut with the running best, and forecasts three quantiles of the running best at the later tenths of the budget; the last value is the running best at the cut.
The conditions are the curve alone, the curve with the configuration of the run (agent model, scaffold, budget, hardware, and a short task description), and the full context, which adds the task statement and the evaluation notes; task names are removed everywhere.
Skill is $S=1-R$ as in \S\ref{sec:tasks}, with one cell per source and the total taken over sources.
A view whose running best does not move after the cut has a last-value error of zero and no error ratio, so it does not enter $R$; we report the answers on these views separately.
Answers that do not parse are dropped, as in the main comparison, and so are answers whose forecast leaves the range of the prefix by more than five times its width.
\sol{}, \opus{}, and \haiku{} are evaluated, with LC-PFN as the curve fitter (constrained, like every baseline on a running best, not to fall below~the~last~value).

\subsection{One run at a time}
\begin{table}[t]
\centering\scriptsize\setlength{\tabcolsep}{3pt}
\caption{Skill $S$ on single runs of the test tasks (cutoffs 5--20\% pooled), with 95\% intervals over tasks. Tasks: test tasks with a scored view. The first total is over the eight sources with per-evaluation scores, the~second~adds~EdgeBench.}
\label{tab:rsi-run}
\resizebox{\linewidth}{!}{%
\begin{tabular}{@{}lrlllll@{}}
\toprule
Source & tasks & \sol{}, curve & \sol{}, full & \opus{}, curve & \opus{}, full & LC-PFN \\
\midrule
OpenRSI Index & 3 & $-0.33$ $[-0.84, 0.44]$ & $-1.17$ $[-1.90, 0.27]$ & $-0.42$ $[-1.33, 0.32]$ & $-0.64$ $[-1.06, 0.45]$ & $-0.21$ \\
Evo-LLM trajectories & 2 & $-0.65$ $[-1.36, -0.14]$ & $-0.66$ $[-1.76, 0.02]$ & $-0.54$ $[-1.33, -0.03]$ & $-0.34$ $[-0.76, -0.04]$ & $0.00$ \\
Autoresearch Bench & 5 & $0.32$ $[-0.06, 0.48]$ & $0.19$ $[-0.35, 0.44]$ & $0.29$ $[0.08, 0.36]$ & $0.26$ $[-0.10, 0.44]$ & $0.05$ \\
Solo-vs-forum & 3 & $0.25$ $[0.20, 0.54]$ & $0.27$ $[0.23, 0.32]$ & $0.30$ $[0.28, 0.31]$ & $0.27$ $[0.23, 0.36]$ & $0.06$ \\
EvoTrace & 5 & $-0.02$ $[-0.51, 0.34]$ & $-0.15$ $[-0.88, 0.06]$ & $-0.16$ $[-0.78, 0.32]$ & $-0.31$ $[-3.08, 0.26]$ & $0.09$ \\
Gome traces & 11 & $0.44$ $[0.26, 0.57]$ & $0.42$ $[0.29, 0.52]$ & $0.43$ $[0.35, 0.49]$ & $0.41$ $[0.27, 0.50]$ & $0.16$ \\
OpenEvolve case study & 2 & $-0.88$ $[-8.3, 0.62]$ & $-3.87$ $[-16.7, -0.34]$ & $-0.07$ $[-1.27, 0.50]$ & $-0.17$ $[-3.2, 0.68]$ & $-1.04$ \\
AlgoTune & 16 & $-0.89$ $[-2.04, -0.23]$ & $-1.29$ $[-2.83, -0.46]$ & $-0.93$ $[-2.52, -0.19]$ & $-0.70$ $[-1.60, -0.14]$ & $-0.06$ \\
\midrule
Total, eight sources & 47 & $-0.11$ $[-0.45, 0.17]$ & $-0.41$ $[-0.83, -0.09]$ & $-0.05$ $[-0.26, 0.14]$ & $-0.08$ $[-0.41, 0.16]$ & $-0.08$ $[-0.21, 0.06]$ \\
\midrule
EdgeBench & 31 & $-0.06$ $[-0.45, 0.21]$ & $-0.67$ $[-1.60, -0.11]$ & $0.00$ $[-0.30, 0.20]$ & $-0.18$ $[-0.87, 0.23]$ & $0.02$ \\
Total, nine sources & 78 & $-0.10$ $[-0.41, 0.15]$ & $-0.44$ $[-0.82, -0.14]$ & $-0.05$ $[-0.23, 0.13]$ & $-0.09$ $[-0.39, 0.13]$ & $-0.06$ $[-0.19, 0.06]$ \\
\bottomrule
\end{tabular}}
\end{table}

\textbf{No forecaster improves on the last value across sources.}
Table~\ref{tab:rsi-run} gives the skill by source.
Over the eight sources with per-evaluation scores, the totals are $-0.11$ $[-0.45, 0.17]$ for \sol{}, $-0.05$ $[-0.26, 0.14]$ for \opus{}, and $-0.16$ $[-0.51, 0.13]$ for \haiku{} with the curve alone, and $-0.08$ $[-0.21, 0.06]$ for LC-PFN; adding EdgeBench leaves them unchanged.
Two sources are positive for every model and condition, Gome ($0.41$ to $0.44$ for \sol{} and \opus{}) and Solo-vs-forum ($0.25$ to $0.30$), and two, Evo-LLM and AlgoTune, are negative for both models with the curve, with intervals that exclude zero; without Gome, whose axis is known only after the run has finished, the totals of \sol{} and \opus{} fall to $-0.22$ and $-0.15$.
Forecasting the raw score in a 2\% window instead of the running best gives the same picture ($0.04$ $[-0.11, 0.23]$ and $0.03$ $[-0.08, 0.14]$ for \sol{} and \opus{} with the curve).

\textbf{Context does not help.}
On the views that a model answers under two conditions, the full context raises the error of \sol{} by 27\% $[18, 41]$ over the curve alone and leaves \opus{} and \haiku{} unchanged within their intervals ($1.03$ $[0.91, 1.22]$ and $1.19$ $[0.96, 1.52]$).
The configuration alone raises the error of \sol{} by 21\% $[11, 51]$, and replacing it by the configuration of a run on another task changes neither model ($1.01$ $[0.80, 1.14]$ and $1.26$ $[0.91, 1.65]$ against the true configuration), so its content carries no measurable information for the forecast.

\textbf{On flat stretches the models forecast gains that do not occur.}
Over all 13 sources, 104 to 109 views per LLM condition (116 for LC-PFN) have a running best that never moves after the cut.
On these views the final-node forecast claims an improvement of more than 1\% of the prefix scale in 81\% of the answers of \sol{}, 85\% of \opus{}, and 68\% of \haiku{} with the curve alone, against 25\% for LC-PFN.
These views do not enter $S$, so the totals above describe the views on which the agent still improves.

\subsection{A fleet of runs on one task}
\begin{table}[t]
\centering\small
\caption{Reading every run of a task at 20\% of the budget: skill $S$ on the future best and the future median of the task's runs (the last value is the current best or median), and Spearman $\rho$ between the forecast and the realized remaining gain at the end of the budget, pooled over six sources (100 test tasks, 110 fleets; \opus{} 98 tasks, 107 fleets). 95\% intervals over tasks. Lower block: \sol{} and \opus{} on the median target under other inputs; with $k$ runs shown, the last value is the median of the shown runs and the target is that of the whole fleet.}
\label{tab:rsi-fleet}
\resizebox{\linewidth}{!}{%
\begin{tabular}{@{}lllll@{}}
\toprule
Forecaster & $S$, best & $S$, median & $\rho$, best & $\rho$, median \\
\midrule
\sol{} & $-0.07$ $[-0.74, 0.33]$ & $0.33$ $[0.07, 0.55]$ & $0.49$ $[0.30, 0.65]$ & $0.72$ $[0.55, 0.85]$ \\
\opus{} & $0.03$ $[-0.39, 0.33]$ & $0.34$ $[0.08, 0.52]$ & $0.48$ $[0.26, 0.66]$ & $0.69$ $[0.52, 0.84]$ \\
Order statistics & $0.04$ $[-0.16, 0.21]$ & $0.33$ $[0.09, 0.50]$ & $0.21$ $[0.01, 0.41]$ & $0.37$ $[0.17, 0.55]$ \\
Record rate & $0.04$ $[-0.27, 0.27]$ & $0.17$ $[-0.04, 0.35]$ & $0.02$ $[-0.20, 0.23]$ & $0.29$ $[0.11, 0.48]$ \\
LC-PFN, per run & $0.04$ $[-0.11, 0.17]$ & $0.11$ $[0.05, 0.18]$ & $0.12$ $[-0.09, 0.34]$ & $0.23$ $[0.04, 0.40]$ \\
ExtraTrees & $-0.51$ $[-1.65, 0.15]$ & $-0.11$ $[-0.82, 0.32]$ & $0.30$ $[0.08, 0.47]$ & $0.54$ $[0.36, 0.71]$ \\
\midrule
 & \multicolumn{2}{c}{$S$, median} & \multicolumn{2}{c}{$\rho$, median} \\
\cmidrule(lr){2-3}\cmidrule(lr){4-5}
Input & \sol{} & \opus{} & \sol{} & \opus{} \\
\midrule
$k=2$ runs & $0.15$ $[0.02, 0.26]$ & $0.18$ $[-0.05, 0.37]$ & $0.42$ $[0.25, 0.57]$ & $0.43$ $[0.25, 0.58]$ \\
$k=4$ runs & $0.37$ $[0.25, 0.46]$ & $0.39$ $[0.27, 0.51]$ & $0.50$ $[0.33, 0.64]$ & $0.56$ $[0.42, 0.67]$ \\
$k=8$ runs & $0.14$ $[-0.10, 0.32]$ & $0.24$ $[0.06, 0.37]$ & $0.63$ $[0.47, 0.76]$ & $0.68$ $[0.53, 0.80]$ \\
Statement of another task & $0.40$ $[0.18, 0.57]$ & $0.37$ $[0.10, 0.56]$ & $0.65$ & $0.68$ \\
No curves & $-2.87$ $[-4.85, -1.55]$ & $-3.34$ $[-5.09, -2.09]$ & $-0.15$ & $-0.16$ \\
\bottomrule
\end{tabular}}
\end{table}

\textbf{Design.} A research-agent benchmark runs many agents on each task, and the decision it faces is where to spend the next runs.
For this reading the unit shown to the forecaster is a fleet: all usable runs of one task under one setting (the task itself, except for Autoresearch Bench, where a task has several boards, and Solo-vs-forum, where a fleet holds the runs of one agent model), each as its windowed curve up to the cut, with the task statement, the evaluation notes, and the roster of agent models.
The targets are the best and the median running best of the fleet at the later nodes, and tasks are ranked by the remaining gain that the forecast implies at the end of the budget.
Six sources have at least three runs per task on a budgeted axis; their test parts hold 100 tasks with a fleet (one AlgoTune and one EvoTrace test task of Table~\ref{tab:rsi-sources} have none), 110 fleets, and 2{,}037 runs, 85 of the tasks from EdgeBench and AlgoTune.
The baselines are an extreme-value fit to the pooled scores with per-run order statistics, the expected rate of record-breaking evaluations, LC-PFN applied to each run, and ExtraTrees trained on the development tasks of the other sources.

\textbf{A fleet forecasts the typical outcome of a task and ranks tasks by their headroom.}
Table~\ref{tab:rsi-fleet} gives the readings.
For the median of the fleet, \sol{} and \opus{} reach $S=0.33$ $[0.07, 0.55]$ and $0.34$ $[0.08, 0.52]$, level with the order-statistic baseline ($0.33$ $[0.09, 0.50]$); for the best of the fleet no forecaster improves on the current best ($-0.07$ and $0.03$).
The forecasts order the tasks by the gain still to come: $\rho=0.72$ $[0.55, 0.85]$ and $0.69$ $[0.52, 0.84]$ for the median, where the highest baseline, ExtraTrees, reaches $0.54$ $[0.36, 0.71]$ and the order statistics $0.37$.
The ordering sharpens with the number of runs shown, from $0.42$ with two runs to $0.50$ with four, $0.63$ with eight, and $0.72$ with all of them for \sol{}.
The signal is in the curves: without them, from the description of the task alone, the error on the same fleets is $4.4$ to $4.9$ times as large and the ordering is lost ($\rho=-0.15$), and the statement of another task in place of the true one changes the error by a factor of $0.90$ to $1.03$, with every interval including one.

\subsection{Matching and failure monitoring}
The two other tasks of the benchmark do not carry over.
Asked which of two settings produced which of two curves, on 146 pairs from 82 tasks, \sol{} and \opus{} are right on 62\% and 68\% of the pairs when the agent models are named and on 56\% and 55\% when they are anonymized; the difference comes entirely from the pairs that differ in the agent model, that is, from a prior about which model is stronger.
Monitors of agent-side failure and of stagnation are at chance for both LLMs (AUROC $0.49$ to $0.57$ on 211 and 199 runs, every interval including $0.5$); a logistic regression on the same windows reaches $0.62$ $[0.55, 0.68]$ for failures.

\subsection{Scope}
The readings use three models, the 20\% cutoff or earlier, and test tasks that two sources dominate at the task level.
The ordering of tasks is not tested against the baselines with a paired statistic, and at a 10\% cutoff ExtraTrees reaches the same value as \opus{} ($0.65$ against $0.66$).
Several sources state no license; the requests derived from them are not part of the release.

\end{document}